\documentclass[british]{article}
 \usepackage{jmlr2e}

\usepackage[utf8]{inputenc}

\usepackage{graphicx}
\usepackage{epstopdf}
\usepackage[round]{natbib}
\usepackage{xr-hyper}
\usepackage[colorlinks=true]{hyperref}
\usepackage[dvipsnames]{xcolor}
\hypersetup{
    allcolors = MidnightBlue,
}

\usepackage{amsmath}
\usepackage[utf8]{inputenc} %
\usepackage[T1]{fontenc}    %
\RequirePackage{doi}
\usepackage{url}            %
\usepackage{booktabs}       %
\usepackage{amssymb}
\usepackage{amsthm}
\usepackage{amsfonts}       %
\usepackage{nicefrac}       %
\usepackage{microtype}
\usepackage[justification=centering]{subfig}
\usepackage{enumitem}
\usepackage{wrapfig}
\usepackage{mathtools}
\usepackage{pgffor}
\usepackage{grffile}
\usepackage{caption}
\usepackage{tikz}
\usepackage{subfiles}
\usepackage{varioref}
\usepackage{etoolbox}
\usepackage{fixltx2e}
\usepackage{tabularx}
\usepackage{adjustbox}

\newcolumntype{C}{>{\centering\arraybackslash}X}
\usepackage{multirow}

\usetikzlibrary{patterns}
\usetikzlibrary{bayesnet}
\usetikzlibrary{arrows,decorations.markings}
\tikzstyle{disent_latent} = [circle,pattern=north east lines, pattern color=black!20,draw=black,inner sep=1pt,
minimum size=20pt, font=\fontsize{10}{10}\selectfont, node distance=1]

\tikzstyle{obs_det} = [latent, diamond, fill=gray!25]

\DeclareMathOperator{\expect}{\mathbb{E}}

\DeclareMathOperator{\ELBO}{\mathcal{L}}
\DeclareMathOperator{\KL}{\mathrm{KL}}

\newcommand{\vecto}[1]{\boldsymbol{\mathbf{#1}}}
\renewcommand{\v}{\vecto}
\newtheorem{theorem}{Theorem}
\newtheorem{defn}{Definition}
\newtheorem{proposition}{Proposition}

\DeclarePairedDelimiterX\MeijerM[3]{\lparen}{\rparen}%
{\begin{smallmatrix}#1 \\ #2\end{smallmatrix}\delimsize\vert\,#3}

\title{I Don't Need $\mathbf{u}$: Identifiable Non-Linear ICA Without Side Information}

\author{\name Matthew Willetts  \email mwilletts@turing.ac.uk \\
       \addr University College London {\&} The Alan Turing Institute
       \AND
       \name Brooks Paige  \email bpaige@turing.ac.uk \\
       \addr University College London {\&} The Alan Turing Institute
}

\begin{document}

\maketitle

\begin{abstract}
In this paper, we investigate the algorithmic stability of unsupervised representation learning with deep generative models, as a function of repeated re-training on the same input data. Algorithms for learning low dimensional linear representations---for example principal components analysis (PCA), or linear independent components analysis (ICA)---come with guarantees that they will always reveal the same latent representations (perhaps up to an arbitrary rotation or permutation). Unfortunately, for non-linear representation learning, such as in a variational auto-encoder (VAE) model trained by stochastic gradient descent, we have no such guarantees. Recent work on identifiability in non-linear ICA have introduced a family of deep generative models that have identifiable latent representations, achieved by conditioning on side information (e.g. informative labels). We empirically evaluate the stability of these models under repeated re-estimation of parameters, and compare them to both standard VAEs and deep generative models which learn to cluster in their latent space. Surprisingly, we discover side information is not necessary for algorithmic stability: using standard quantitative measures of identifiability, we find deep generative models with latent clusterings are empirically identifiable to the same degree as models which rely on auxiliary labels. 
We relate these results to the possibility of identifiable non-linear ICA.

\end{abstract}
\section{Introduction}
\label{intro}

When evaluating the performance of a deep generative model (DGM) with latent variables, there are two primary concerns.
The first is the quality of the fit to the data: i.e., to what extent does the learned model approximate the distribution of the underlying data.
This includes checking whether synthetic generations from the trained model look like plausible ``real'' data.
Evaluating the fit of the model to the data typically involves integrating out the latent variables.

The second slightly-more-elusive concern is whether we can ascribe any particular meaning or utility to the latent variables themselves.
Linear methods such as probabilistic principal components analysis~\citep{Tipping1999}, and nonlinear methods such as variational autoencoders~\citep{Rezende2014,Kingma2013}, associate each data point with a posterior distribution over a low-dimensional latent variable.
These posteriors (or summaries, such as the posterior mean) are often hoped to be useful for downstream tasks, as unsupervised representations; they also are often hoped to correspond to underlying ground-truth or statistically independent factors of variation.
Much recent work has focused on the question of learning variational autoencoders with ``disentangled'' representations \citep{Higgins2017,Kim2018,Chen2018,Esmaeili2018}, with mixed results \citep{Locatello2019, Rolinek2019}.

We argue that the latter concern is closely connected with a notion of \textit{algorithmic stability}, where we are interested in the replicability of the particular learned representation under re-estimation of model parameters using a randomized algorithm.
As an example, principal components analysis can be estimated using a power iteration method, which despite a random initialization has guarantees to recover a particular representation.
However, when considering autoencoder-style models, instead we must perform stochastic gradient descent on a non-convex objective.
To take seriously any claims that variational autoencoders learn conceptually meaningful representations, we suggest it is necessary to also understand whether such representations are reliably recovered by existing learning algorithms, or instead may just be found by the ``luck'' of selecting a particular random seed.

To understand the stability of learned representations under stochastic optimization, we turn to recent work on the {\em identifiability} of the latent representations of a model.
 In the classic definition of identifiability for parameters in a probabilistic model, this means that two different settings of the model parameters define two different probability distributions.
In highly flexible models, 
multiple settings of parameters could define the same probability distribution (so the model's parameters are not identifiable) while having \emph{identifiable latent representations}~\citep{tcl, Hyvarinen2019, Khemakhem2019,iflows,gin,Khemakhem2020,Roeder2020}. 
A standard definition of identifiability of latent representations used in deep generative models is that different models' learnt representations are only an affine transform, or (more strictly) a permutation, away from each other~\citep{tcl,Khemakhem2019, Roeder2020, Khemakhem2020}.
It is often these forms of identifiability, whether learnt representations are an affine transform/permutation away from each other, that a) are tested empirically when discussing the identifiability of a given model, including in experiments within primarily theoretical treatments, and b) are of importance to practitioners.

In the last few years there has been a resurgence in identifiability results in machine learning models
within certain problem-settings \citep{tcl, Hyvarinen2019, Khemakhem2019,iflows,gin,Khemakhem2020,Roeder2020}.
Many of these recent results have been in the space of non-linear ICA, where data has been made by a non-linear mixing of statistically-independent latent sources.
Classically the aim is to `unmix' the data, recovering the true sources in the process.
For two decades now it has been understood that, for anything other than a lucky subset of problems, it is not theoretically possible to learn a definitive non-linear unmixing when the model being applied by the analyst is a universal function approximator~\citep{Hyvarinen1999}.
These recent results provide theory and methods for finding identifiable non-linear unmixing, using auxiliary side information to sufficiently structure the model's representations.
Informally, the side information is used to `break the symmetry' in the space of representations the model could learn, resulting in a stable and repeatable training procedure.

In this paper we investigate
how various purely-unsupervised deep generative models compare to deep generative models with side information at consistently learning the same latent representations, when tested under the same conditions as in the literature on identifiable models~\citep{tcl, Hyvarinen2019, Khemakhem2019,iflows,gin,Khemakhem2020,Roeder2020}.
From the ICA perspective, we are interested in whether the various sources of inductive bias, such as the neural architecture of the model, its probabilistic form and the method of inference used~\citep{Shu2018} sufficiently constrain the training of these purely-unsupervised models to give them highly-repeatable learnt representations when retrained with different seeds.
We train both vanilla VAEs and VaDEs~\citep{Jiang2017} (a VAE that clusters via a learnt mixture model in its latent space) and benchmark them against iVAEs~\citep{Khemakhem2019} using methods from the identifiability literature. %
Note that VaDE is in effect an iVAE where the clustering is learnt rather than given.
Theory suggests we should see stable, consistent representations learned in the iVAE setting, but not necessarily for VAEs or for VaDE.

Our main finding is that under standard empirical metrics of the consistency and identifiability of the representations, we observe that VaDE gives performance which matches, or surpasses, provably-identifiable baselines, even without
side-information present.
VAEs themselves perform similarly (and thus surprisingly) well under the same testing conditions.
We see no statistically significant difference between (unsupervised) VaDE and (supervised) iVAE, while both improve on a standard VAE to a degree which is statistically significant, but likely not operationally significant.
This is also, we note, the first large-scale evaluation of the identifiability of iVAEs, VAEs and VaDEs on image data.
Our results raises questions as to both the theoretical requirements for identifiability---which these unsupervised models do not fulfill---and perhaps also as to the current methods of empirical analysis in the identifiability literature.

\section{Background}
\label{sec:back}
\begin{figure*}[t]
    \centering
\begin{tabularx}{\textwidth}{CCC}
\centering
\includegraphics[height=3.2cm]{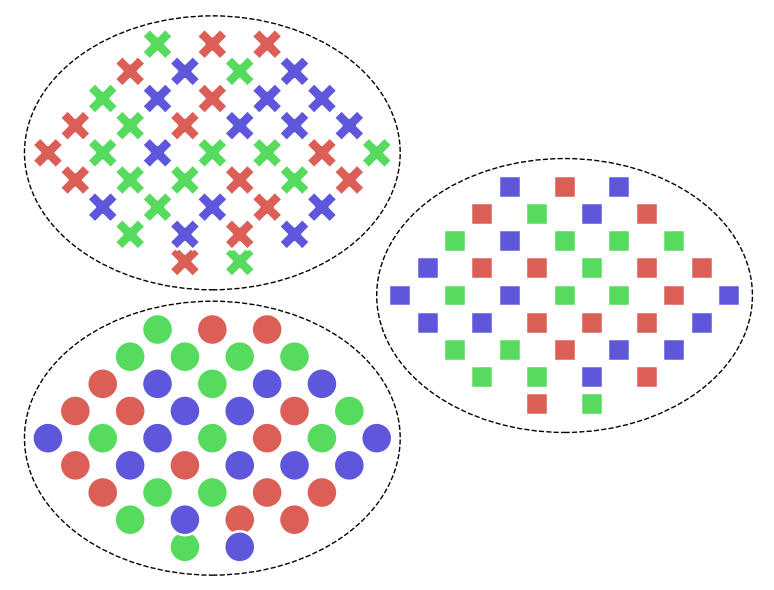}&
\includegraphics[height=3.2cm]{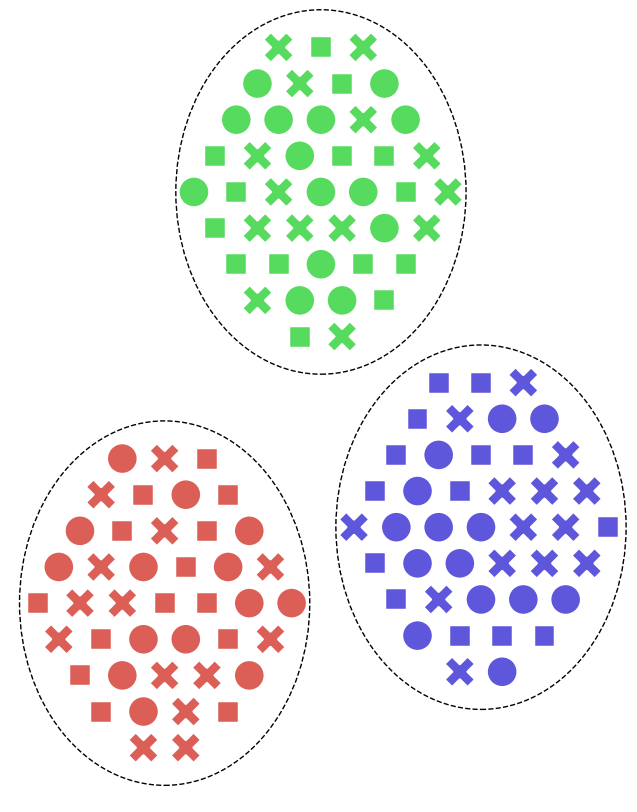}&
\includegraphics[height=3.2cm]{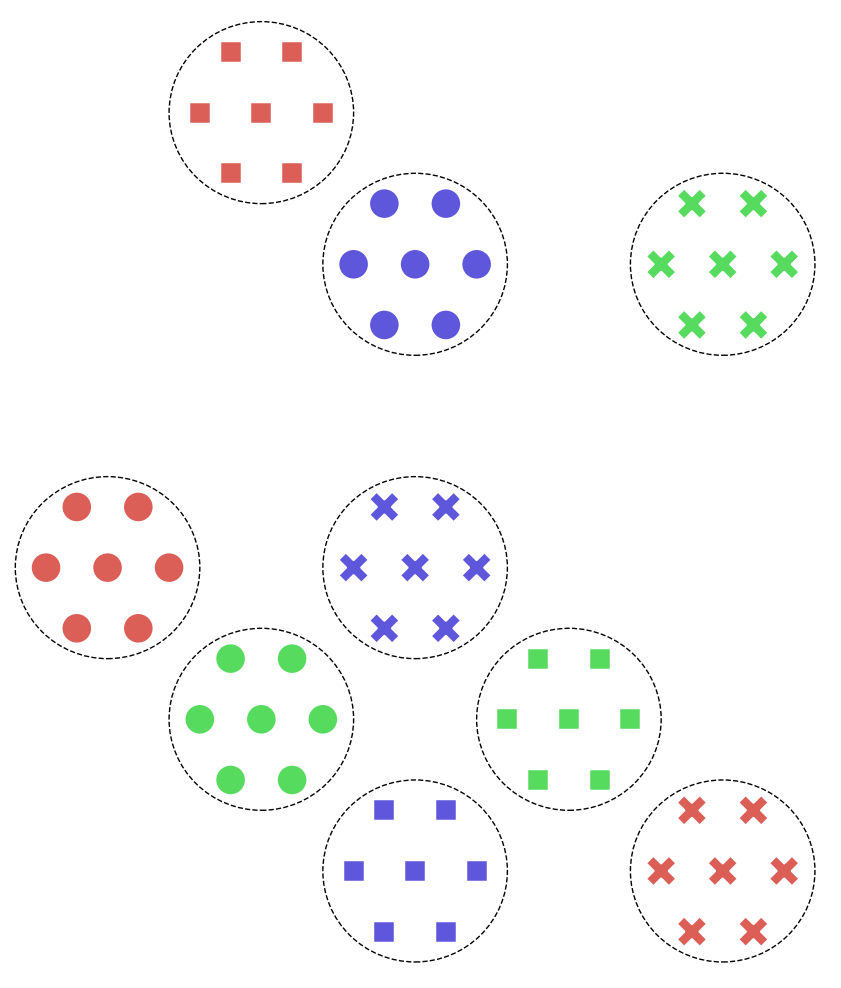}\\
$\v{u}_1$, shape & $\v{u}_2$, colour & $\v{u}_3=\v{u}_1\times\v{u}_2$, shape and colour \\

\end{tabularx}
\caption{\emph{Different ways to partition a set of objects.} Methods theoretically guaranteed that produce identifiable representations must be given a particular way, $\v u$, to partition the data being trained on, say $\v{u}_1$, the shape of an object. 
If a clustering model, due to its inductive biases, always learns the same $\v u$-clustering, in this illustrative example it could be shape, colour or both together, it could then obtain identifiable representations.
}    \label{fig:clustering}
\end{figure*}
There have been two main paradigms in identifiability results in non-linear ICA.
The first is concerned with learning a discriminative model, a classifier, under an appropriate classification task for the data.
An example of a task would be identifying what large-scale segment of a temporal sequence a particular observation $\v x$ is from~\citep{tcl}, or predicting the datapoint's class-label~\citep{Hyvarinen2019}.

The dependent variable in our classification task is denoted $\v u$.
Using the density ratio trick~\cite[\S14.2.4]{elements}, one can build a noise contrastive estimation (NCE) loss~\citep{Gutmann2012}.
The classifier's predictive probability can be linked to related probabilities in a matching generative model.

If the model is well-trained, the final feature layer of the classifier corresponds to 
the underlying generating sources, assuming that the chosen and measured $\v u$ variable connects to the true generative process for the data.
Thus it can then be shown that two classifiers that give the same predictive probabilities produce identifiable unmixed representations~\citep{tcl, Hyvarinen2019}.
This setting induces identifiability, up to the choice of classification task; different tasks will lead to different learnt representations.

The second, more recent, strand is based around learning a generative model that embodies both the non-linear procedure that generated the data and a (perhaps approximate) method of inverting it~\citep{Khemakhem2019}.
For a latent variable model for data $\v x \in \mathcal{X}$ with latents $\v z \in \mathcal{Z}$, the definition of identifiability is:
\begin{defn}[Identifiability up to equivalence class,~\cite{Khemakhem2019}]\ \\
Let $\sim$ be an equivalence relation on $\Theta$.
A model defined by $p_\theta(\v x, \v z) = p_\theta(\v x| \v z)p_\theta(\v z)$ with $\theta\in\Theta$ is said to be identifiable up to $\sim$ if
$p_\theta(\v x) = p_{\theta'}(\v x) \rightarrow \theta \sim \theta'$.
The elements of the quotient space $\Theta/\sim$ are called the identifiability classes.
\label{defn:ident}
\end{defn}

In this approach, identifiability comes from \textit{conditional priors} -- the latents $\v z\in\mathcal{Z}=\mathbb{R}^{d_z}$ in the generative model are themselves conditioned on the same kinds of auxiliary information $\v u \in \mathcal{U}$ as used in the constrastive method.
The priors must have the form
\begin{equation}
    p_{\theta}(\v z | \v u) = \prod_{i=1}^{d_z} \frac{Q_i(z_i)}{Z_i(\v u)}\exp\left[\sum_{j=1}^k T_{i,j}(z_i)\lambda_{i,j}(\v u)\right],
    \label{eq:cond_prior}
\end{equation}
where $T_{i,j}$ are the sufficient statistics of the distribution, $\lambda_{i,j}$ its natural parameters with normaliser $Z_i$ and base measure $Q_i$ (commonly 1).
$k$ is the maximum order of statistics of these distributions.
(For us $k=2$ throughout as, following previous work, we choose $p_{\theta}(\v z | \v u)$ to be a product of independent Gaussians.)
The generative model must form a Bayes net $\v u \rightarrow \v z \rightarrow \v x$, and $p_\theta(\v x| \v z) = p_{\v \epsilon}(\v x - f_\theta(\v z))$ for some noise $\v \epsilon$.
The equivalence relation for identifiability in this class of latent variable models is:
\begin{proposition}[Equivalence $\sim$,~\cite{Khemakhem2019}]\ \\
$(f_\theta, \v T, \lambda)\sim(\Tilde{f}_\theta, \Tilde{\v T}, \Tilde{\lambda})$ are of the same equivalence class if and only if there exist $\v A$ and $\v c$ such that $\forall \v x \in \mathcal{X}$,
$\v{T}(f^{-1}_\theta(\v x)) = \v{A}\Tilde{\v{T}}(\Tilde{f}^{-1}_\theta(\v x)) + \v c$.
\end{proposition}

\begin{defn}[Weak and Strong $\sim$,~\cite{Khemakhem2019}]\ \\
If $\v{A}$ is invertible, we denote the equivalence relation $\sim_A$.
This is \textup{weak} identifiability.\\
If $\v{A}$ is a block permutation matrix (block-wise wrt the $k$ orders of sufficient statistics), we denote the equivalence relation $\sim_P$.
This is \textup{strong} identifiability.
\label{defn:weakstrong}
\end{defn}

Analogously to the contrastive case, the choice of $\v u$-task is the choice of which unmixing you get---it is intrinsic in all these approaches that they provide identifiability up to the choice of $\v u$-task.
See Figure~\ref{fig:clustering} for an example of different $\v u$-tasks for an imagined dataset.
That is, different $\v u$-tasks cannot be expected to provide the same learnt, mutually identifiable representations as each other-- after all, the representations are \textit{for} a particular purpose, and that purpose is determined by the choice of $\v u$-task.
In order for the generative model to be identifiable, the $\v u$-task has to follow certain mathematical constraints.

\begin{theorem}[Weak Identifiability Theorem,~\cite{Khemakhem2019}]\ \\
Suppose that: (i) The set $\{\v x \in \mathcal{X}|\Psi_\epsilon(\v x) = 0\}$ is of measure zero, where $\Psi_\epsilon(\v x)$ is the characteristic function of the noise density $p_{\v \epsilon}(\cdot)$;
(ii) the mixing function $f_\theta$ is injective;
(iii) The sufficient statistics $T_{i,j}$ in Eq~\eqref{eq:cond_prior} are differentiable almost everywhere and $T_{i,j}'= 0$ almost everywhere for $i \in \{1,\dots, d_z\}$ and $j \in \{1,\dots, k\}$;
(iv) There exist $(k d_z+1)$ distinct values of $\v u$, $\v{u}_0,\dots, \v{u}_{(k\times d_z)+1}$, such that the following $k d_z\times k d_z$ matrix is invertible:
\begin{equation}
    \v L = \left(\v{\lambda}(\v{u}_1)-\v{\lambda}(\v{u}_0),\dots,\v{\lambda}(\v{u}_{k d_z+1})-\v{\lambda}(\v{u}_0) \right).
    \label{eq:l_matrix}
\end{equation}
then the parameters $(f_\theta,\v{T},\v{\lambda})$ are $\sim_A$ identifiable.
\label{thm:weak_ident}
\end{theorem}

\begin{theorem}[Strong Identifiability Theorem ($K\geq2$),~\cite{Khemakhem2019}]\ \\
Suppose that in addition to the assumptions in Theorem~\ref{thm:weak_ident} above, further: (i) $k\geq2$, (ii) The sufficient statistics $\v T$ are twice differentiable, (iii) The mixing function $f_\theta$ has all second order cross-derivatives.
then the parameters $(f_\theta,\v{T},\v{\lambda})$ are $\sim_P$ identifiable.
\label{thm:strong_ident}
\end{theorem}

Thus in this approach the theory of non-linear ICA is linked to the theory of deep generative models.
When this theory is applied to VAEs one gets Identifiable-VAEs (iVAEs)~\citep{Khemakhem2019}.
For iVAEs the ELBO, for a single datapoint $\v x$ with auxiliary $\v u$, is
\begin{equation}
    p(\v x | \v u) \geq \ELBO^{\mathrm{iVAE}}(\v x | \v u ;\theta,\phi)=\smashoperator{\expect_{q_\phi(\v z| \v x, \v u)}}\log \frac{p_\theta(\v x | \v z)p_\theta(\v z | \v u)}{q_\phi(\v z | \v x, \v u)}.
    \label{eq:ivae_elbo}
\end{equation}

For this identifiability theory to hold for iVAEs: the generative model must be an infinitely-flexible function-approximator; the true posterior has to be in the family of $q$ posteriors; we must have infinite data; and the global maximum value of $\ELBO^{\mathrm{iVAE}}$ must be found during optimisation~\citep{Khemakhem2019}.
The conditional priors in $\mathcal{Z}$ can be implemented as Gaussians, $p_\theta(\v z|\v u) = \mathcal{N}(\v z|\v{\mu}_\theta(\v u),\v{\Sigma}_\theta(\v u))$ where $\v{\mu}_\theta(\cdot)$ and $\v{\Sigma}_\theta(\cdot)$ are themselves neural networks or look-up tables (and $\v{\Sigma}_\theta(\cdot)$ must be diagonal to fulfill Eq~\eqref{eq:cond_prior}).

\section{Non-Linear ICA via Clustering?}
\label{sec:ident_clustering}
Here we motivate why models that learn to cluster the data, rather than being given that assignment \emph{a priori}, might be able to compete with iVAEs in learning highly-similar representations when repeatedly retrained.

In other words, we are flipping the perspective from having $\v u$ given, to instead jointly learning $\v u$ along with $\v z$.
We will consider each parameterisation of a deep generative model with structure $\v u \rightarrow \v z \rightarrow \v x$ as implicitly inducing a particular $\v u$-task, towards which it is identifiable.
Recall that $\v u$ is used in the iVAE context to index over different priors in the latent space $\v z$---there is a set of distributions $\{ p_\theta(\v z | \v u)\}$ for $\v u \in \mathcal{U}$.
In the iVAE setting our data is composed of $\{\v x, \v u\}$ pairs.
In prior work when training on standard datasets the class index is used as $\v u$.

Instead of having these $\v u$ labels given to us as data, our key idea is to obtain our $\v u$ values from learning a clustering of our data.
We will learn our clustering jointly with learning the (identifiable) representation $\v z$, in a fully probabilistic and unified manner.

For iVAEs, having access to true values of $\v u$ is equivalent to having a posterior for $\v u$ that is one-hot, deterministic.
If we relax this, aiming to jointly learn both the $\v u$-task and $\v u$ posteriors as we learn our $\v z$ posteriors and the rest of the generative model.
Thus we have changed the problem to a form of clustering---$\v u$ is the learnt clustering index in the latent space.
The structure of the learnt $\v u$ will emerge, in part, as a natural consequence of the biases of the model.
For example, a model with single-layer linear mappings will naturally learn very different representations to a model with multiple convolutional layers trained on the same dataset.

Our claim is that the same DGM clustering architecture will learn highly-similar $\v u$ representations when retrained with different initialisations and thus lead to consistently-learnt $\v z$ representations when the model is retrained on the same dataset.
What representations the model learns, what $\v u$-task, is a result of the interactions between the model's particular inductive biases, including inductive biases arising from the chosen method of inference, and the dataset being trained on.
In our experimental results we find, for various datasets, that the learnt representations of our clustering approach, where $\v u$ is learnt, are identifiable.

Critically, we do not need our model to naturally cluster into any particular pre-specified $\v u$-tasks---we are leveraging the theoretical insight in~\cite{Khemakhem2019} that there is a whole world of $\v u$-tasks that provide sufficient separation to obtain identifiability.
In particular we are \textit{not} aiming to match the ground-truth class labels, for example, that are given with standard datasets. 
Instead, we observe that deep generative models with latent mixtures naturally learn clusterings that correspond to appropriate, sufficient $\v u$-tasks.
(For an intuitive example of this, see Fig~\ref{fig:clustering}.)
Here we bring together these ideas to demonstrate the empirical identifiability of latent representations in deep clustering models, which implicitly perform non-linear ICA.

We wish to underline that we in no way dispute %
previous results on the impossibility of generic, unique non-linear ICA.
Rather we are saying that, for imperfect models their particular inductive biases induce a natural emergent clustering in their latent space from which we can obtain identifiable representations.
We highlight that theoretical results~\citep{tcl,Hyvarinen2019,Khemakhem2019} assume universal function approximators.
Further, our chosen method of inference, stochastic amortised variational inference~\citep{Kingma2013, Rezende2014}, induces its own biases into optimisation~\citep{Shu2018}.

The ELBO for an iVAE, Eq~\eqref{eq:ivae_elbo}, can be re-written, so that instead of having our log evidence conditioned on $\v u$ we have it as a latent variable, albeit one with a one-hot posterior,
\begin{equation}
   \smashoperator{\expect_{q_\phi(\v u | \v x)}}\,\,\,\,\,\left[\expect_{q_\phi(\v z| \v x, \v u)}\log p_\theta(\v x | \v z) - \KL(q_\phi(\v z | \v x, \v u)||p_\theta(\v z | \v u)\right].
\end{equation}%
where $q_\phi(\v u |\v x)=\delta(\v u - \omega(\v x))$ is the deterministic distribution that maps perfectly from each $\v x$ to its associated $\v u$, returned by $\omega(\v x)$.
In the standard iVAE setting, $q_\phi(\v u |\v x)$ is implemented using a lookup-table---we are told the true values of $\v u$ for each datapoint $\v x$.

Now $\v u$ is simply an additional latent variable, a discrete latent over cluster components in $\mathcal{Z}$.
This means we will perform inference over it, having placed a prior over it.
This means our generative model is $p_\theta(\v x | \v z)p_\theta(\v z| \v u)p_\theta(\v u)$.
For inference we use the approach first proposed in VaDE\footnote{Recently \cite{mfcvae} has corrected some mistakes in the original VaDE paper, but interestingly these mistakes only become apparent when using more than one Monte Carlo sample $\v z\sim q_\phi(\v z |\v x)$ when estimating $\ELBO$. We use one such sample, as is common practice regardless. Our description of VaDE reflects the corrections made in~\cite{mfcvae}.}~\citep{Jiang2017}.

In this approach one makes two design choices.
First, choose the factorisation $q_\phi(\v z, \v u | \v x)=q_\phi(\v z|\v x)q_\phi(\v u | \v x)$.
Writing the generative model as $p_\theta(\v x | \v z)p_\theta(\v z)p_\theta(\v u | \v z)$, we can write the ELBO as
\begin{align}
\label{eq:expand_elbo_single_vade}
\ELBO^{\mathrm{VaDE}}(\v x;\theta,\phi) =\smashoperator{\expect_{q_\phi(\v{z} | \v{x})}}\,\,&\log p_\theta(\v x | \v z) - \KL\left[q_\phi(\v z | \v x)||p_\theta(\v z)\right]\nonumber \\
&- \smashoperator{\expect_{q_\phi(\v z | \v x)}}\,\,\KL\left[q_\phi(\v u|\v x)||p_\theta(\v u|\v z)\right].
\end{align}
The second choice is not parameterising $q_\phi(\v u | \v x)$ with an additional recognition network. Instead one constructs the Bayes-optimal posterior: the $q_\phi(\v u | \v x)$ that minimises $\expect_{q_\phi(\v z | \v x)}\KL(q_\phi(\v u | \v x)||p(\v u | \v z))$.
This is the case when $q_\phi(\v u|\v x)\propto\exp\left(\expect_{q_\phi(\v z | \v x)}\log p_\theta(\v u|\v z)\right)$ \citep{mfcvae}.

When estimating $\ELBO$ using a single sample $\v z\sim q_\phi(\v z |\v x)$ and constructing $q_\phi(\v u|\v x)$ using that sample, the Monte Carlo (MC) estimate of $\expect_{q_\phi(\v z | \v x)}\KL\left[q_\phi(\v u|\v x)||p_\theta(\v u|\v z)\right]$ is zero~\citep{mfcvae}.
This further simplifies optimisation, giving us our single-MC-sample estimator for $\ELBO^{\mathrm{VaDE}}$,
\begin{equation}
\label{eq:mc_vade}
\ELBO^{\mathrm{VaDE}}_{\mathrm{MC}}(\v x;\theta,\phi) =\smashoperator{\expect_{q_\phi(\v{z} | \v{x})}}\,\,\log p_\theta(\v x | \v z) - \KL\left[q_\phi(\v z | \v x)||p_\theta(\v z)\right].
\end{equation}
Overall this approach has numerous benefits.
Firstly, with it we do not have to perform expensive exact marginalisation over the discrete latent.
Secondly, we do not have to use the Gumbel-Softmax trick to get reparameterised samples from a relaxed discrete distribution, which would introduce bias.
As a result, this approach admits a simple, lightweight, stochastic estimator over minibatches of data.
Finally, in this work we are particularly interested in the intrinsic identifiability of the model we are training.
Each additional module introduced requires its own design choices that in turn will affect the model's overall inductive bias.
With this approach, we end up training a vanilla VAE but for the fact that we have a learnt Gaussian Mixture Model as the prior in $\mathcal{Z}$.
We study the effect of different design choices for $p$ and $q$ in the following sections in detail.
Further, see Appendix~\ref{app:rade} for experiments with a randomly-sampled and then fixed amortised posterior for $\v u$.

\subsection{With No Clustering?}
In the above discussion we are focusing on getting an unsupervised VAE that most-closely resembles an iVAE---VaDE is an iVAE but with $\v u$ learnt.
But given that we are most interested in the empirical nature of the learnt representations of these models (that is, if we retrain a model multiple times are the learnt representations of each run an affine transformation or a permutation away from each other with high fidelity), the performance of vanilla VAEs is also potentially of interest.
Here there is no synthetic $\v u$ task at all.

Of course previous work in `disentangling' has shown that VAEs and various unsupervised extensions~\citep{Higgins2017,Kim2018,Chen2018,Esmaeili2018} only consistently learn meaningful, axis-aligned, latent representations with lucky design choices \citep{Locatello2019, Rolinek2019}.
Disentangling is a highly-related problem to identifiability as understood in the non-linear ICA literature, but not exactly the same.
We are not aware of any empirical study of the learnt representations of VAEs using the methods of analysis used in studying the identifiability of deep generative models.

\section{Experiments}
We are most interested in the question, applicable to training on image datasets: \emph{do reruns of the same VaDE/VAE model result empirically in identifiable latent representations, benchmarked against iVAEs?}
\footnote{We also study the case of training on synthetic data, where the true generative factors are known, in Appendix~\ref{app:synth}.}

Further, we are interested in both the `strong' formulation of identifiability, where $\sim$ in Definition~\ref{defn:ident} is up to permutations only ($\sim_P$ in Definition~\ref{defn:weakstrong}) and also the `weak' formulation where an affine operation maps between two sets of representations ($\sim_A$ in Definition~\ref{defn:weakstrong}).

For the former, strong, case, we measure the mean correlation coefficient (MCC) between two sets of values, learnt representations from two different runs.
We compute the maximum linear correlation, up to a permutation, of components.
We use the Hungarian algorithm~\citep{hungarian} to obtain the optimal permutation.
Large MCC indicates strong correlation between in inputs
(see, for example, \cite{Khemakhem2020},~\S A.2, for more discussion).

In the latter, weak, case, we are interested in the identifiability of affine-aligned representations.
This means we calculate the MCC between two sets of representations having learnt this affine alignment between them.
We follow~\cite{Khemakhem2020} and use Canonical Correlation Analysis (CCA) to learn this mapping.

\begin{figure*}[p]
    \centering
\begin{tabularx}{\textwidth}{cCCC}
\centering
& \hspace{2em} $d_z=50$  & \hspace{2.4em} $d_z=90$  & \hspace{2.3em} $d_z=200$  \\
\raisebox{2.4\height }{\rotatebox[origin=c]{90}{MNIST}}&
\includegraphics[height=3.0cm]{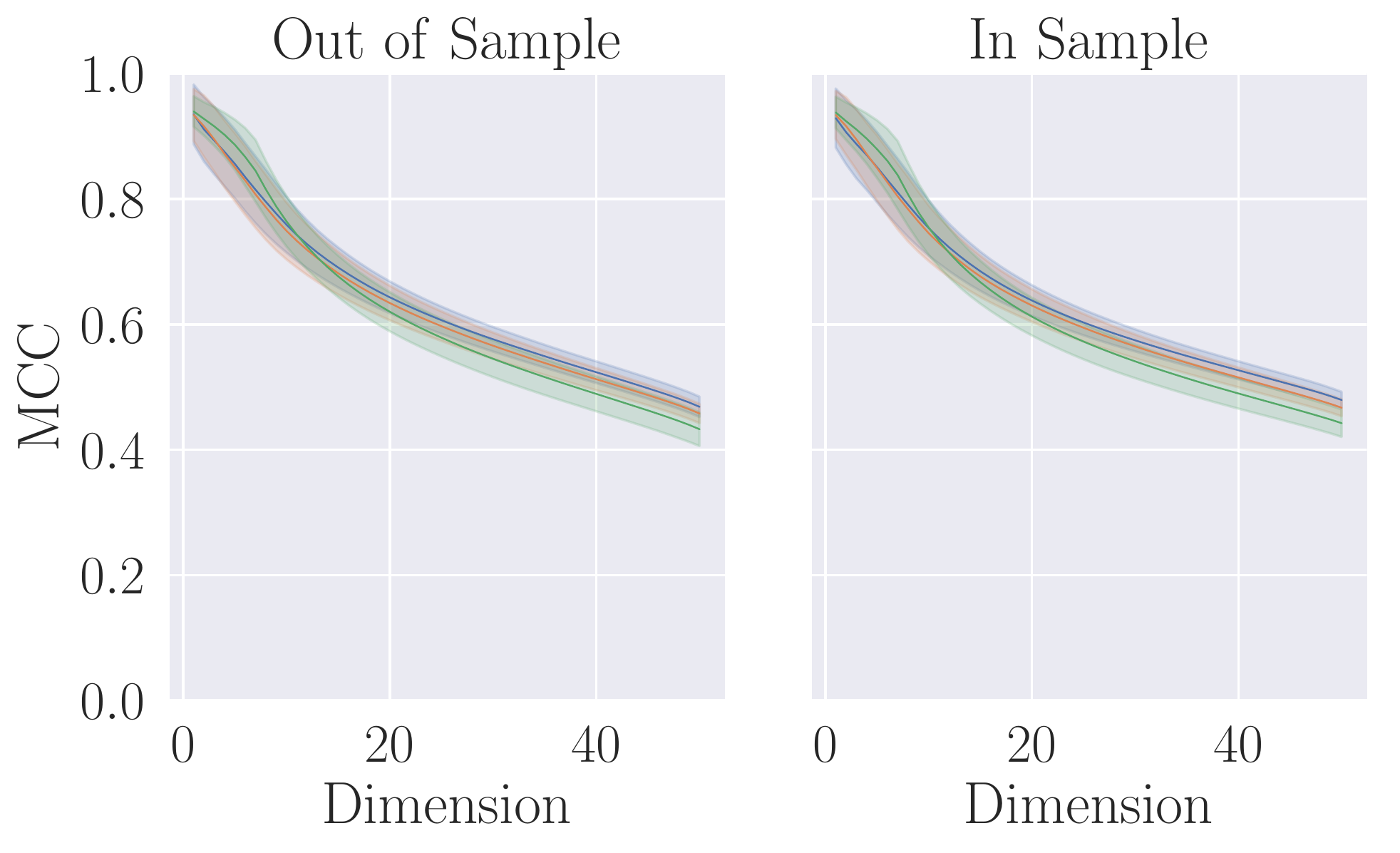}&
\includegraphics[height=3.0cm]{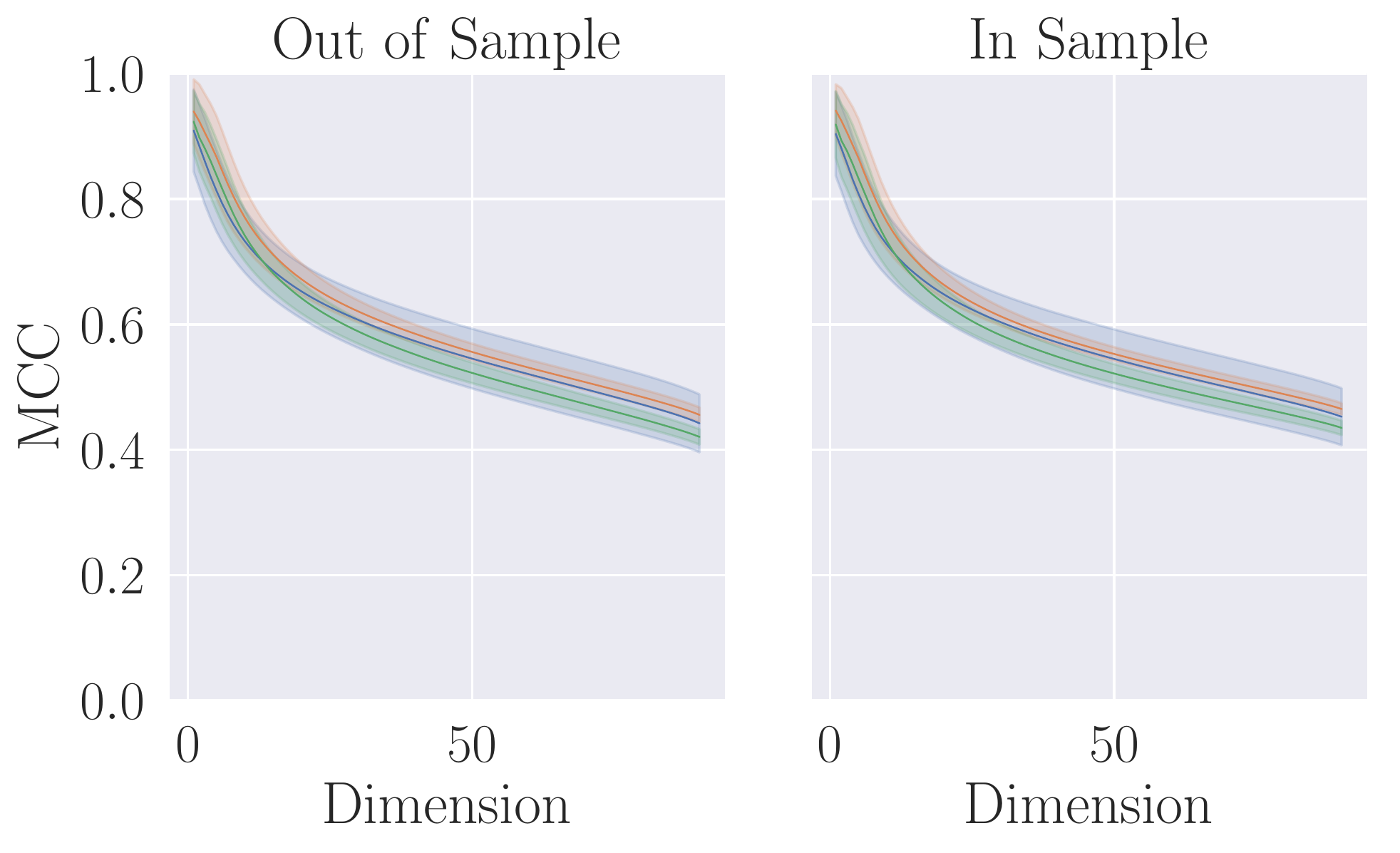}&
\includegraphics[height=3.0cm]{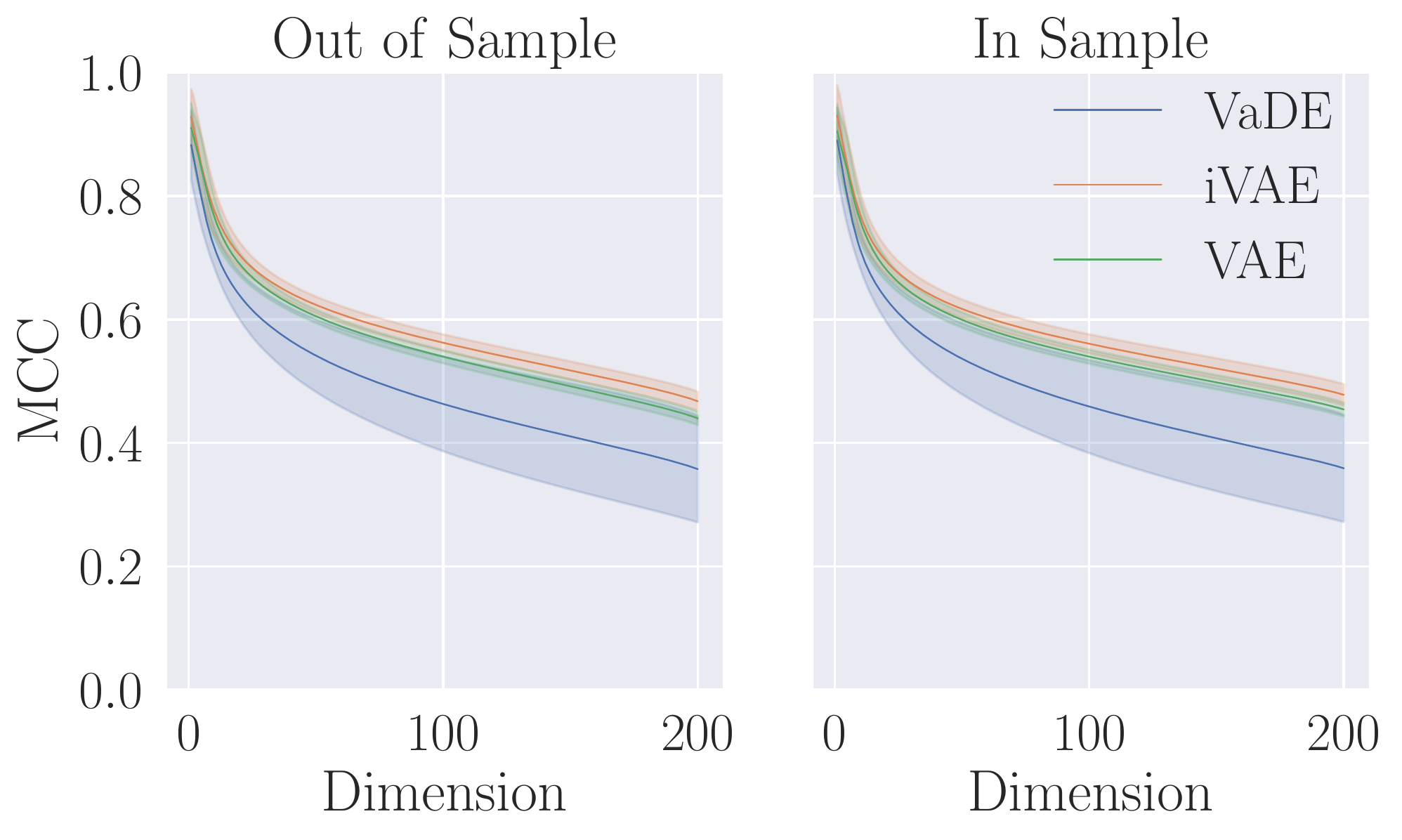}\\
\raisebox{2.1\height }{\rotatebox[origin=c]{90}{CIFAR10}}&
\includegraphics[height=3.0cm]{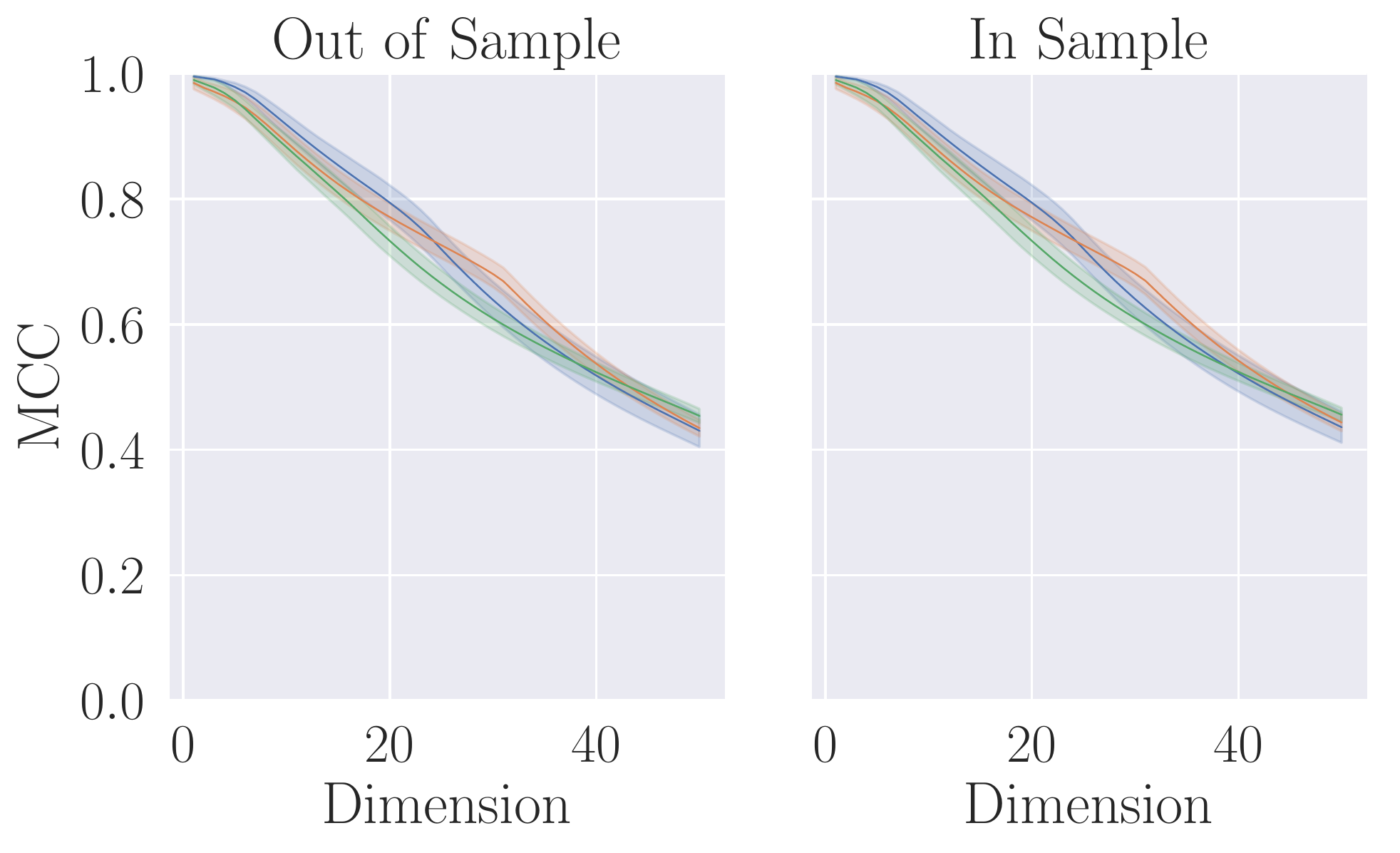}&
\includegraphics[height=3.0cm]{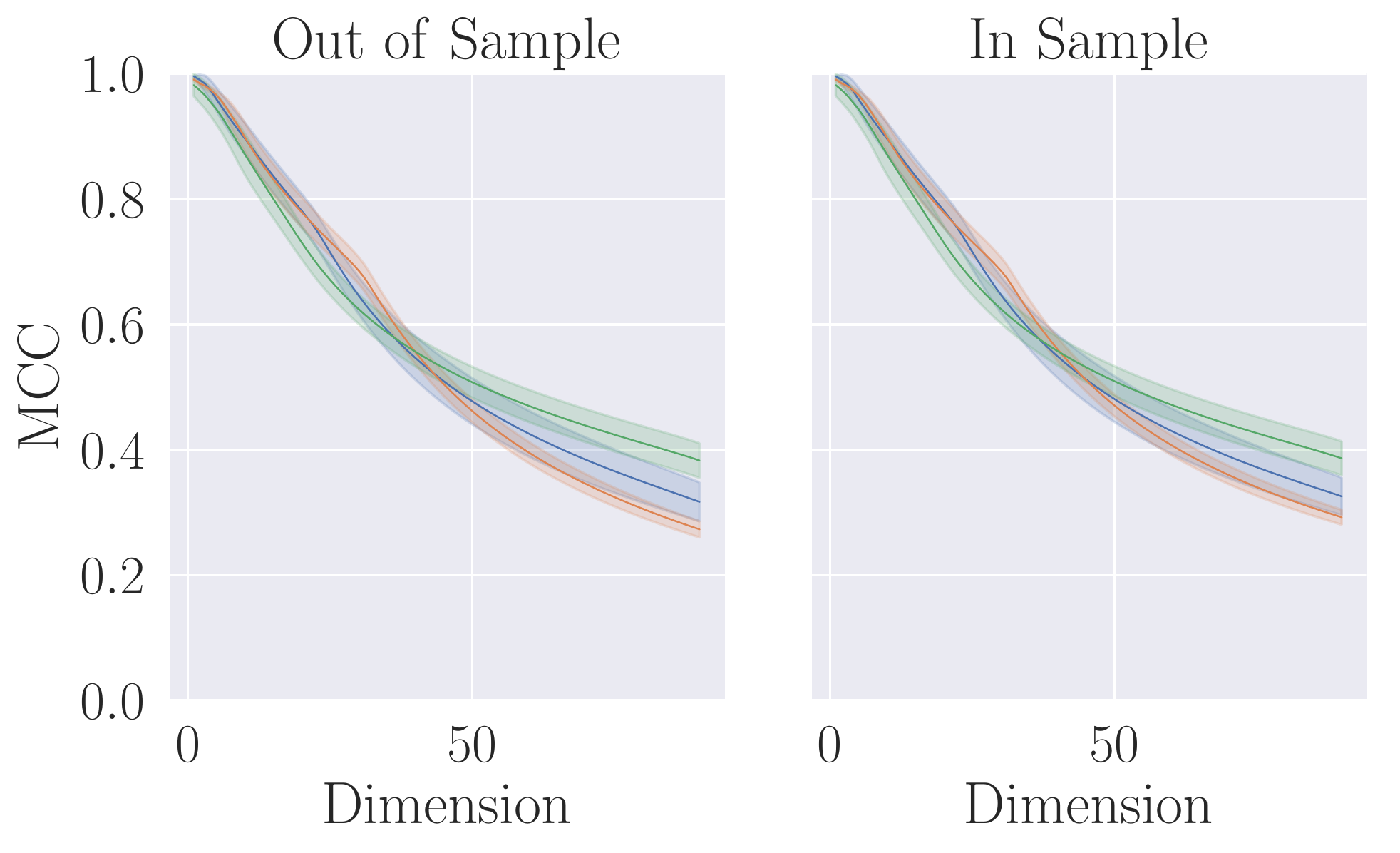}&
\includegraphics[height=3.0cm]{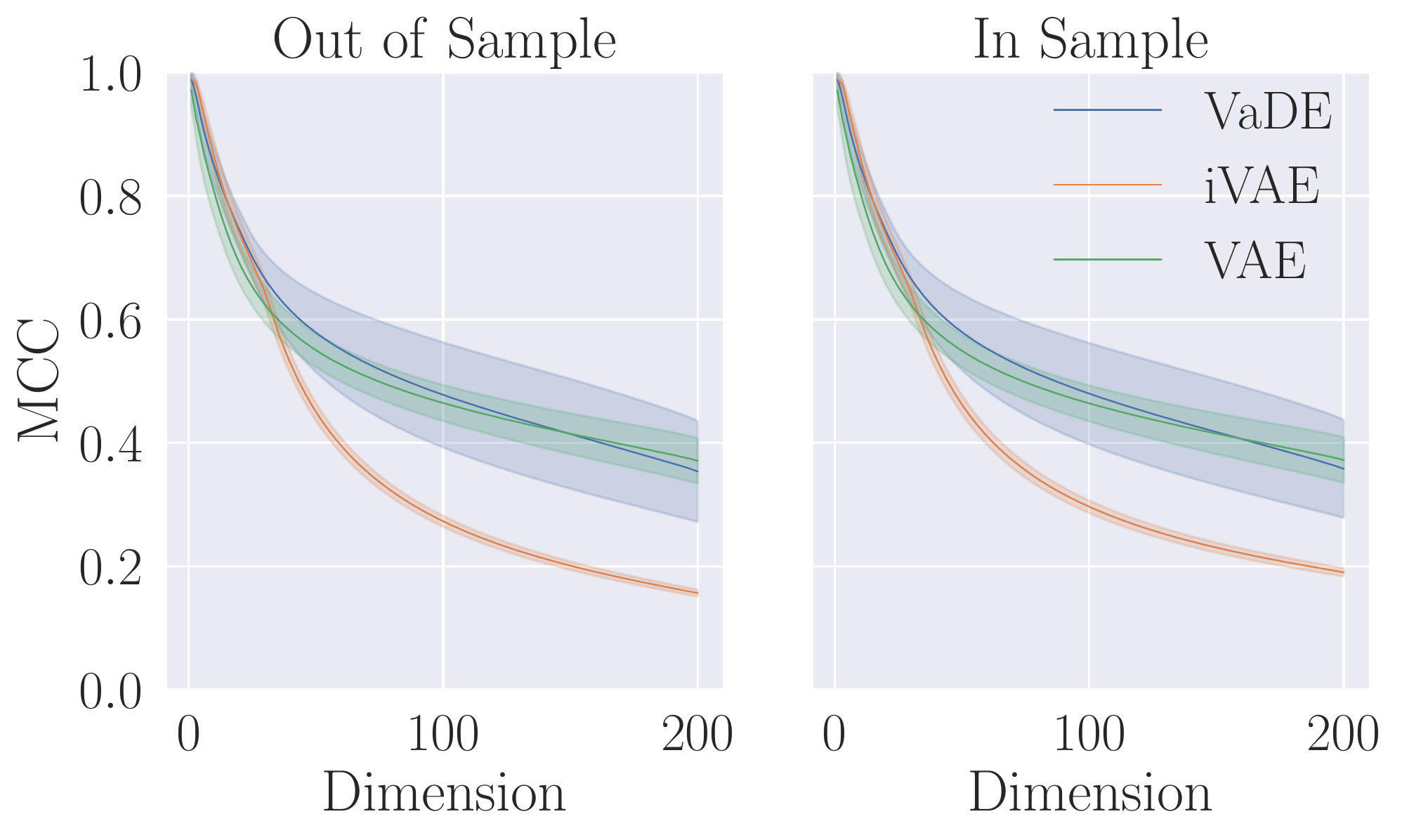}\\
\raisebox{2.8\height }{\rotatebox[origin=c]{90}{SVHN}}&
\includegraphics[height=3.0cm]{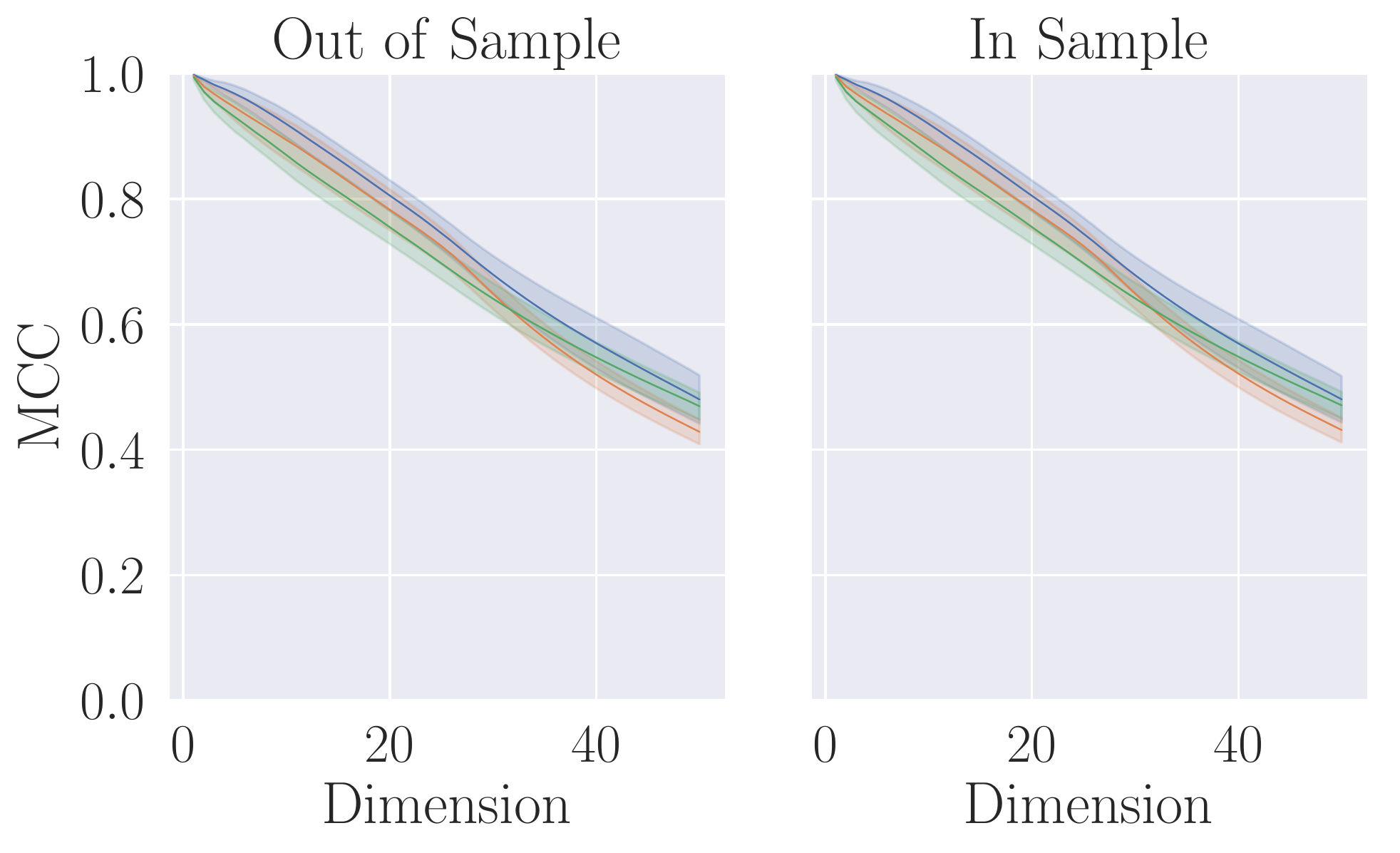}&
\includegraphics[height=3.0cm]{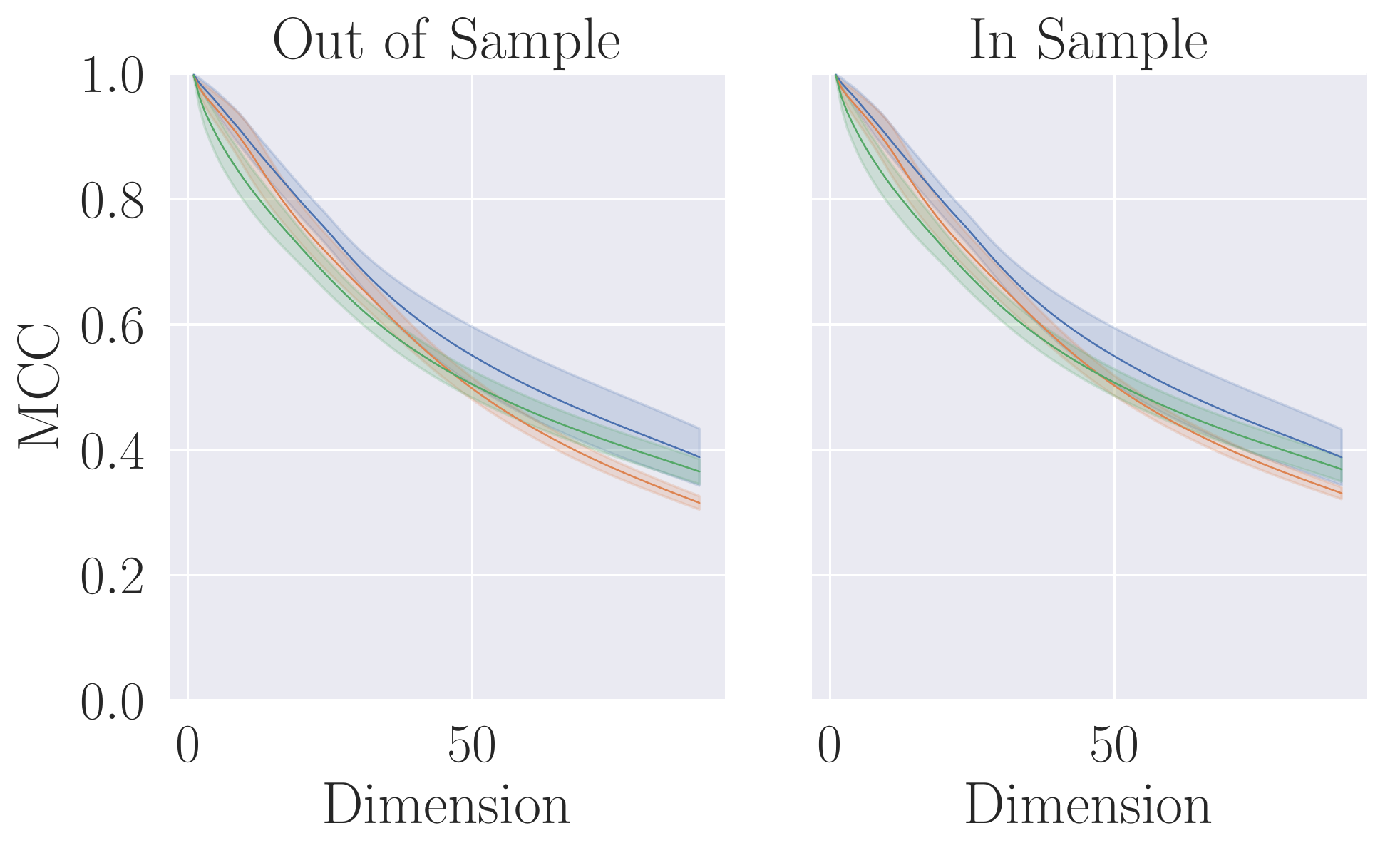}&
\includegraphics[height=3.0cm]{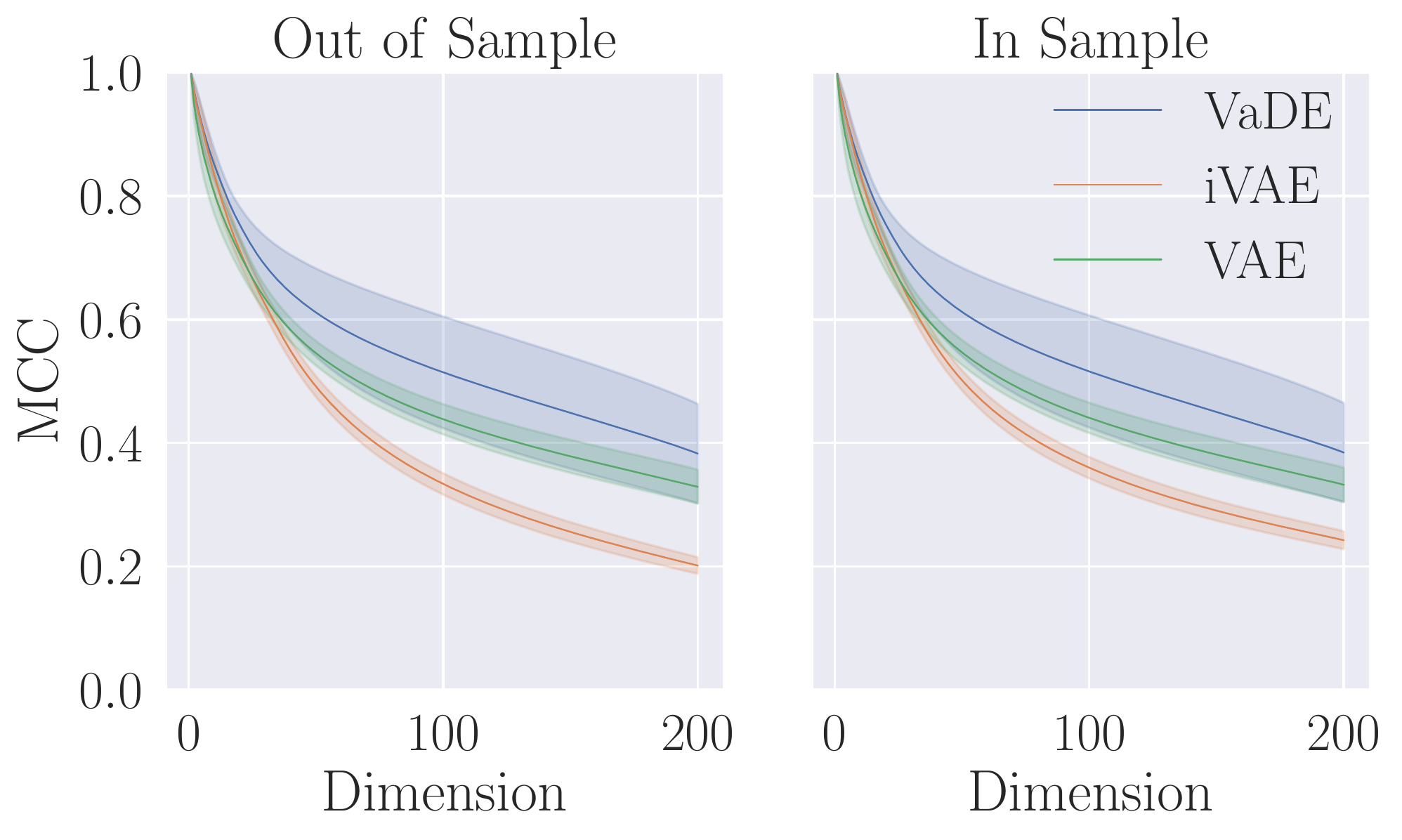}\\
\end{tabularx}
\caption{Plots showing the strong identifiability of $\v{\mu}_\phi(\v x)$ for MLP models, with $d_z\in \{50,90,200\}$, as measured by MCC (measured cumulatively over dimensions).
We show MCC values of the permutation-aligned spaces over the test set for both `In Sample' (MCC calculated over the half of the test set $\v z$ values used to find the optimal permutation) and `Out of Sample' (MCC calculated over the remaining $\v z$ values).
Shading is $\pm$ one standard deviation, calculated over all pairs of initial seeds.
}
\label{fig:strong_mlp_mcc}
\vspace{0.3cm}
    \centering
\begin{tabularx}{\textwidth}{cCCC}
\centering
& \hspace{2em} $d_z=50$  & \hspace{2.4em} $d_z=90$  & \hspace{2.3em} $d_z=200$  \\
\raisebox{2.4\height }{\rotatebox[origin=c]{90}{MNIST}}&
\includegraphics[height=3.0cm]{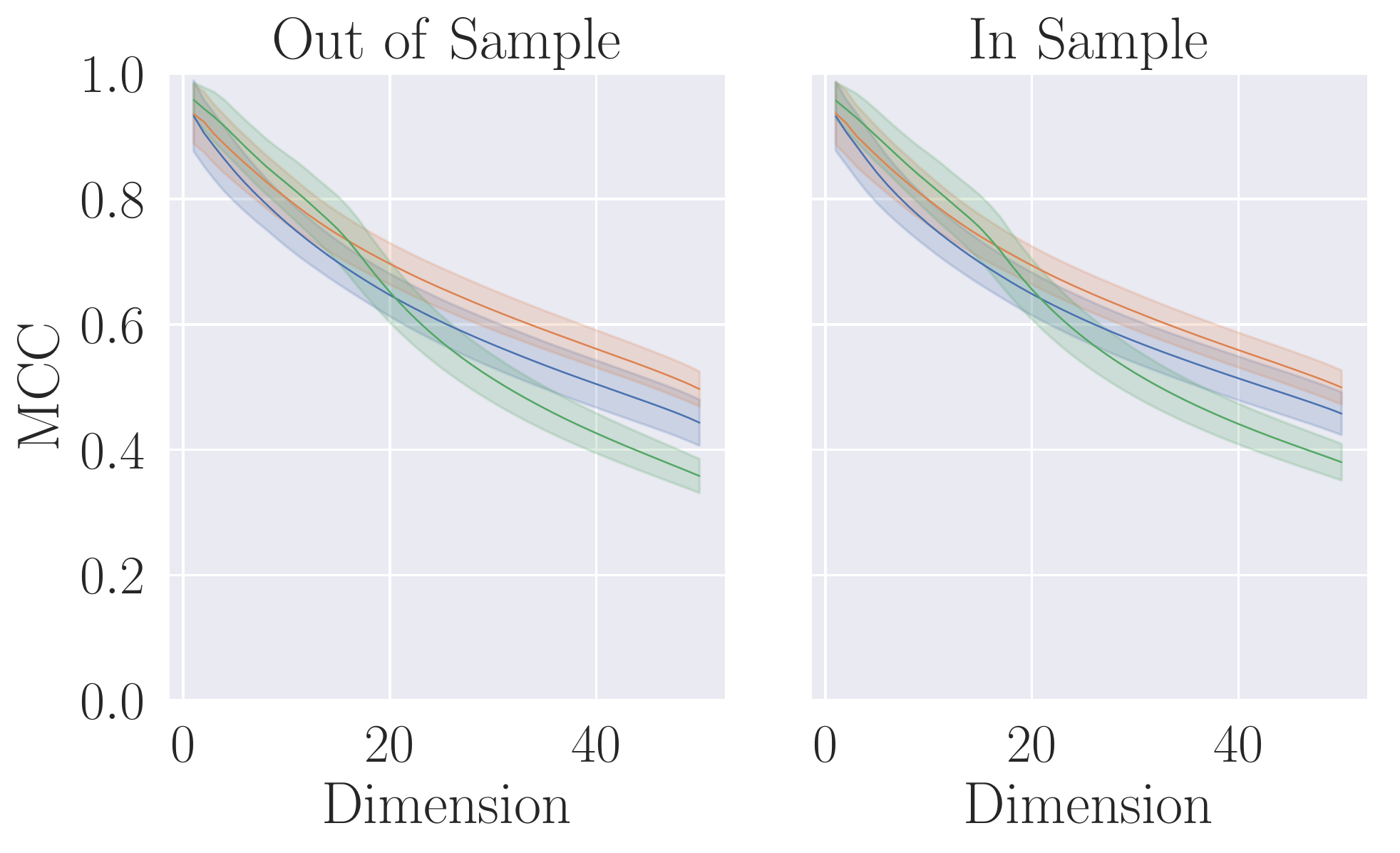}&
\includegraphics[height=3.0cm]{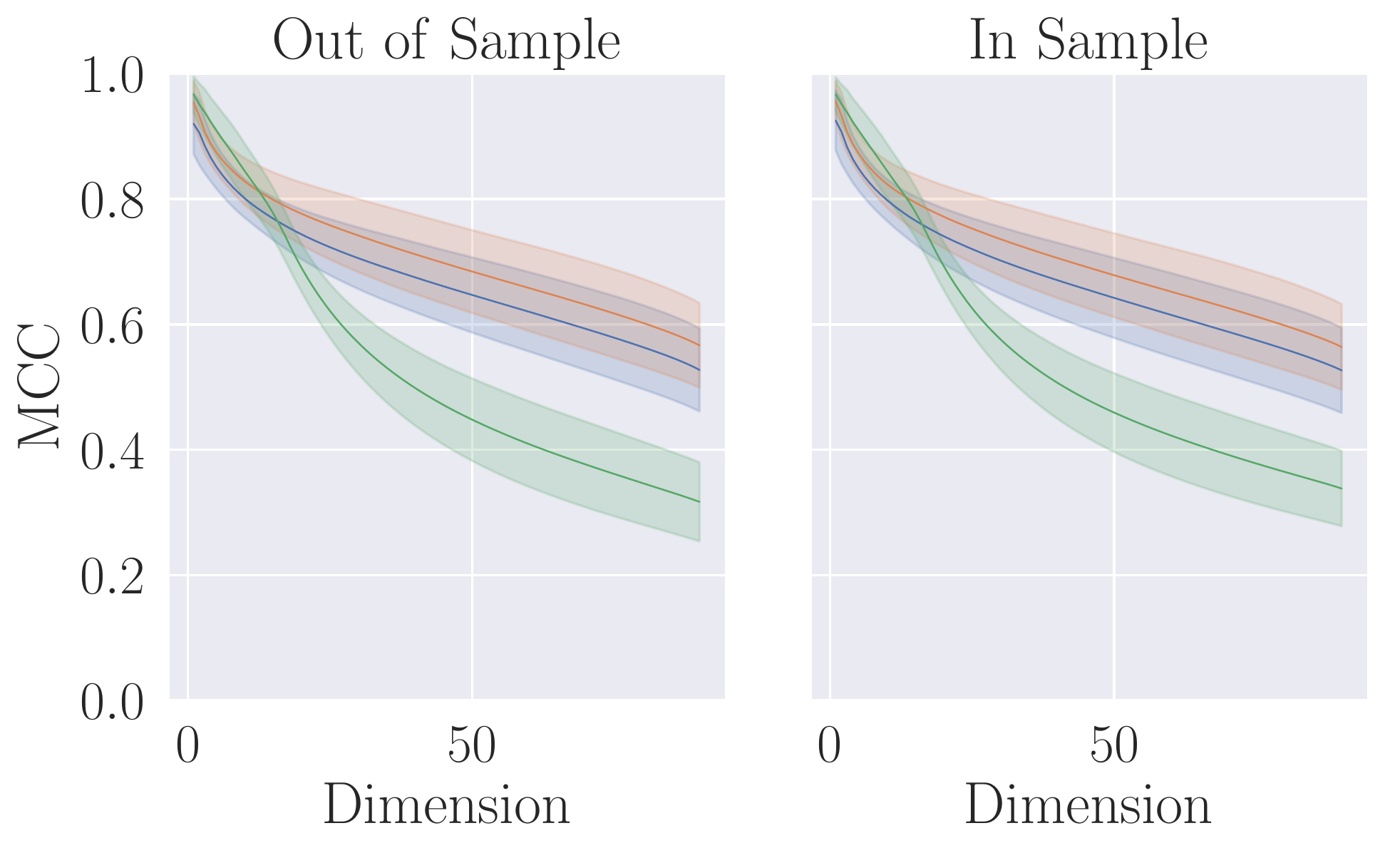}&
\includegraphics[height=3.0cm]{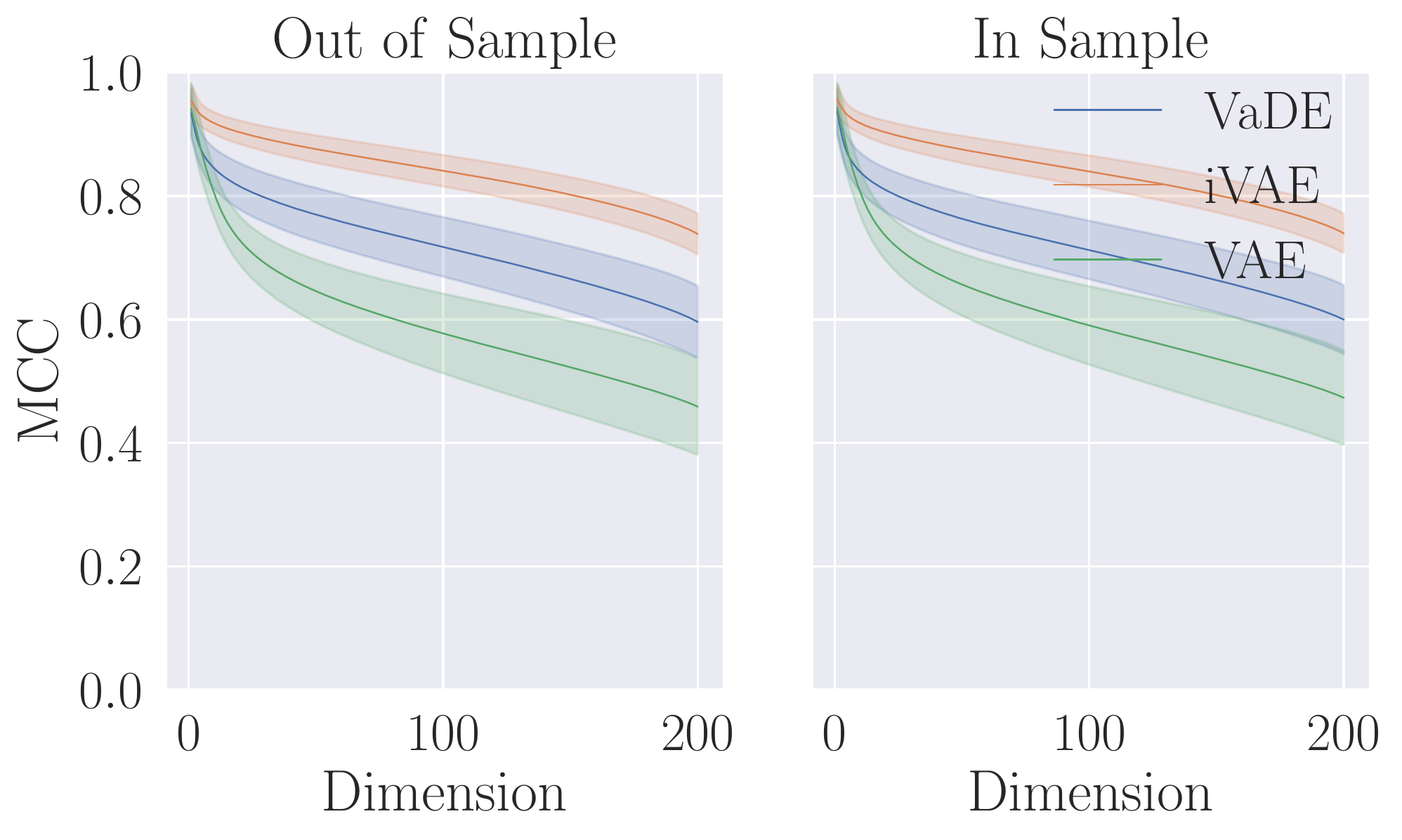}\\
\raisebox{2.1\height }{\rotatebox[origin=c]{90}{CIFAR10}}&
\includegraphics[height=3.0cm]{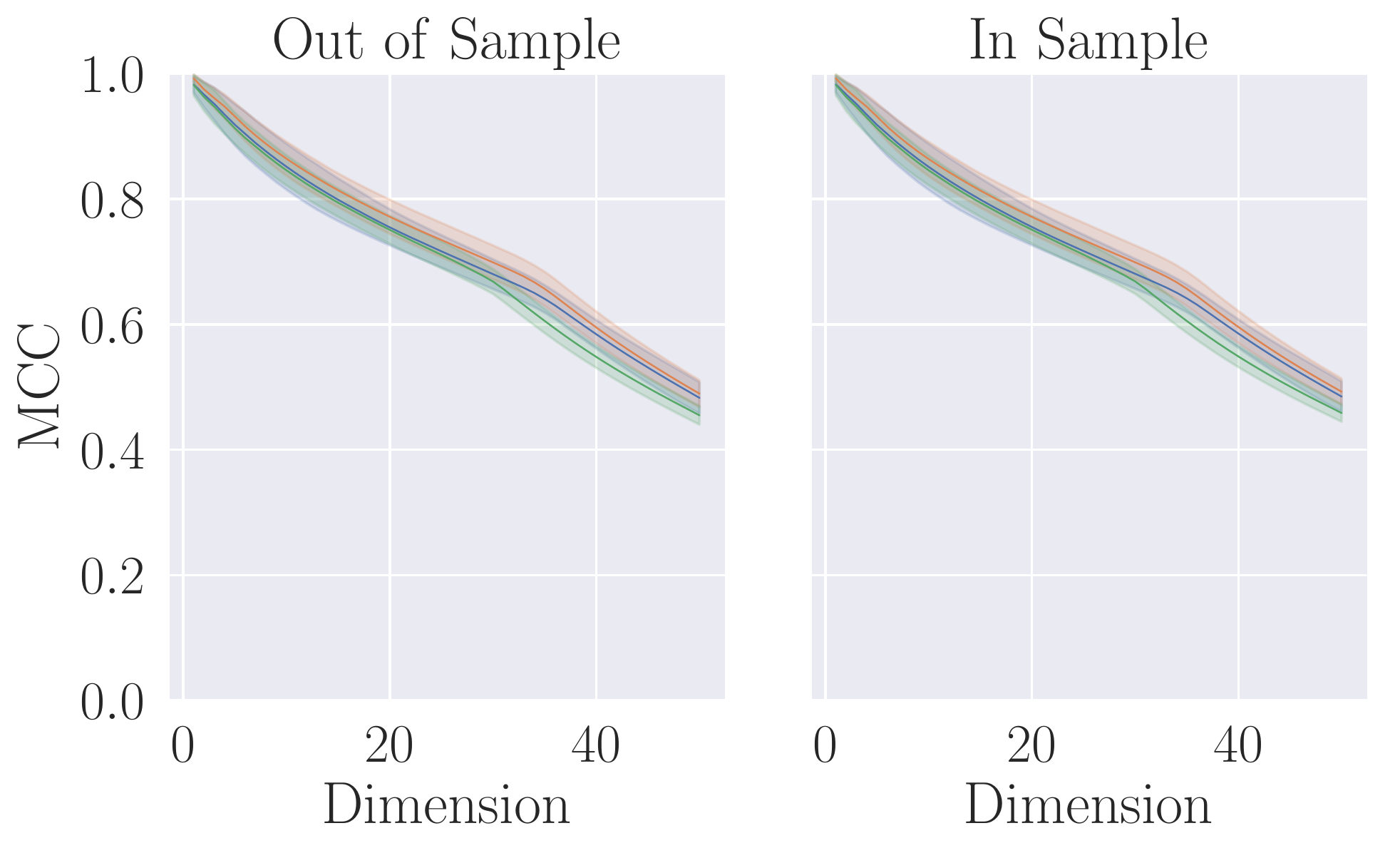}&
\includegraphics[height=3.0cm]{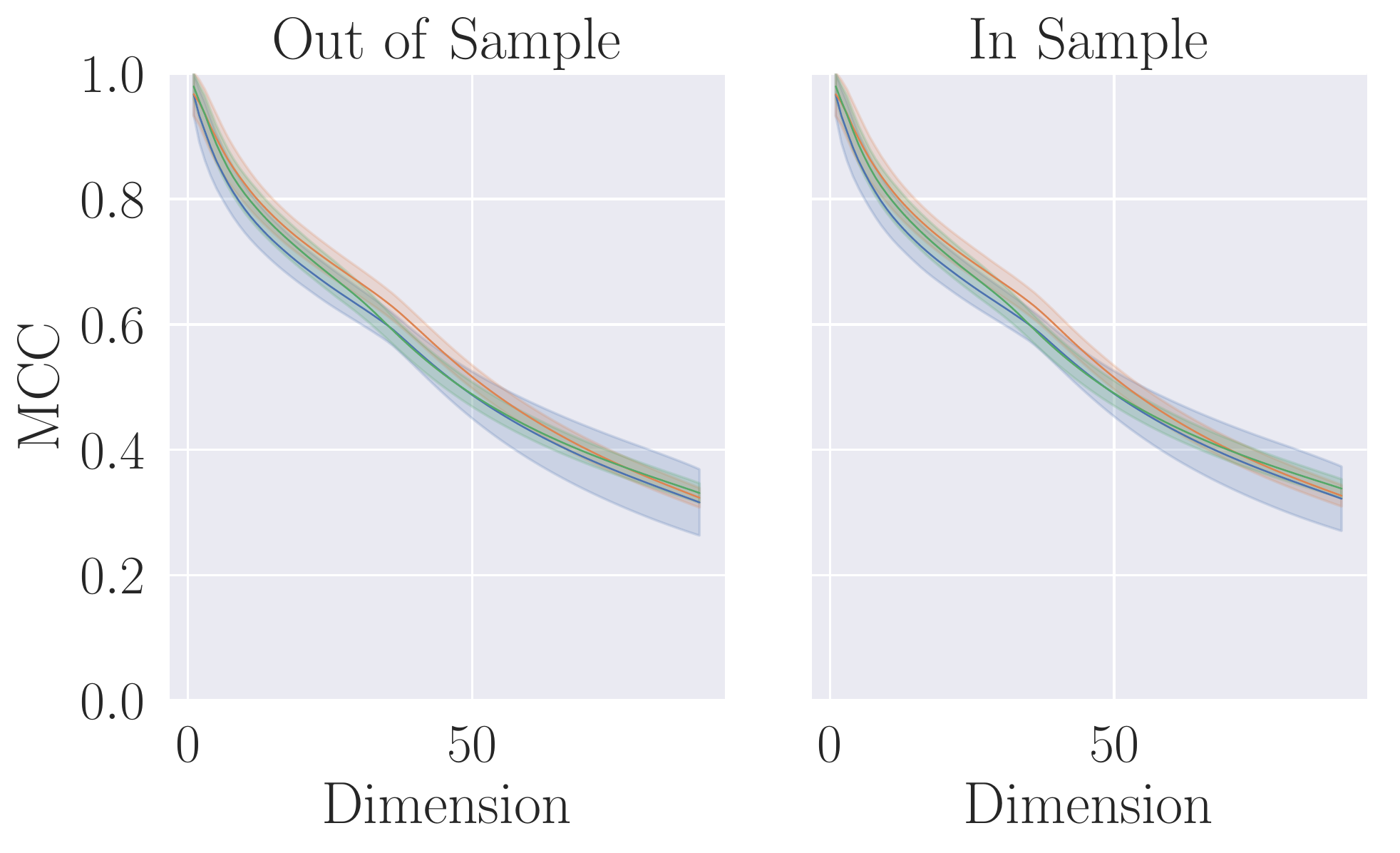}&
\includegraphics[height=3.0cm]{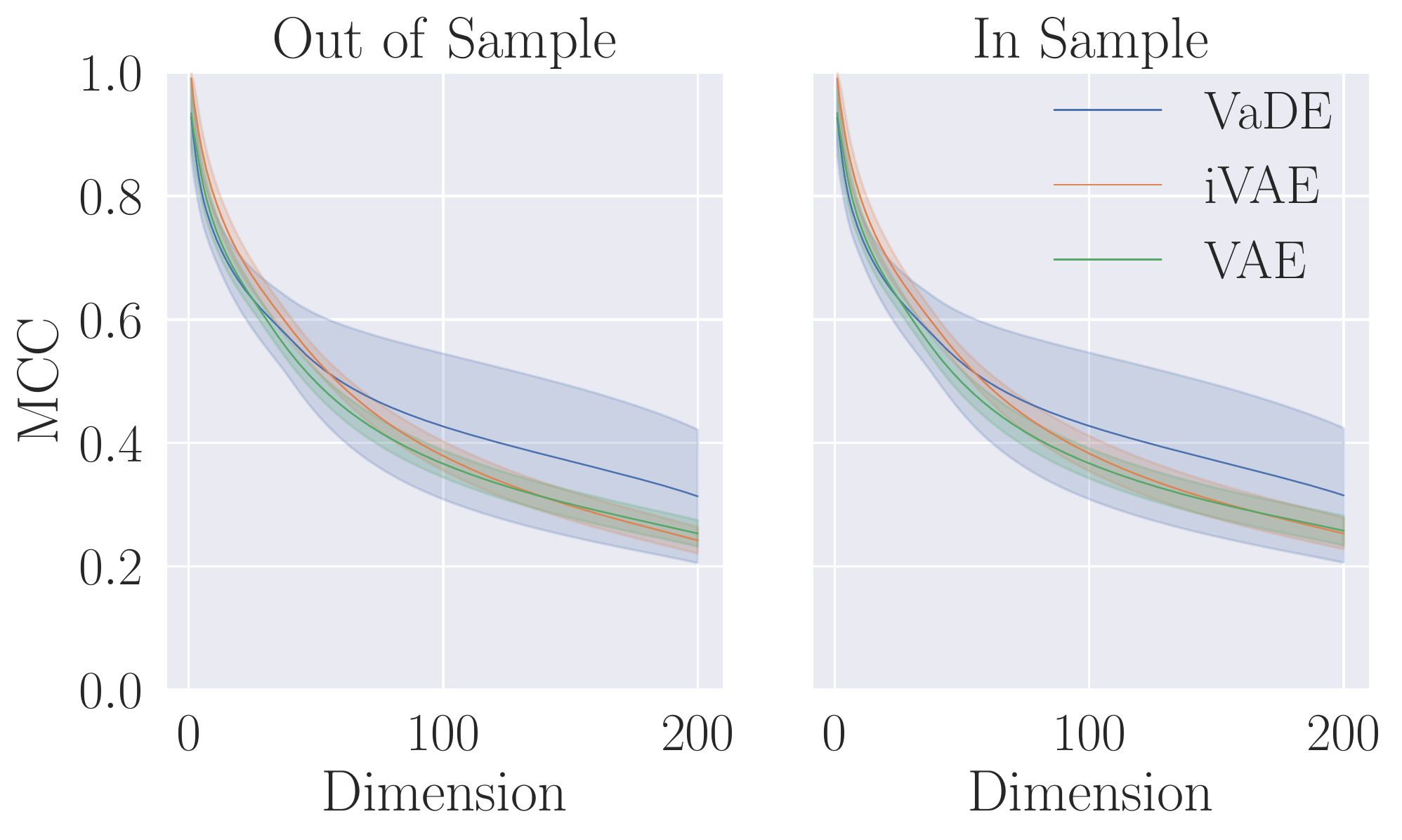}\\
\raisebox{2.8\height }{\rotatebox[origin=c]{90}{SVHN}}&
\includegraphics[height=3.0cm]{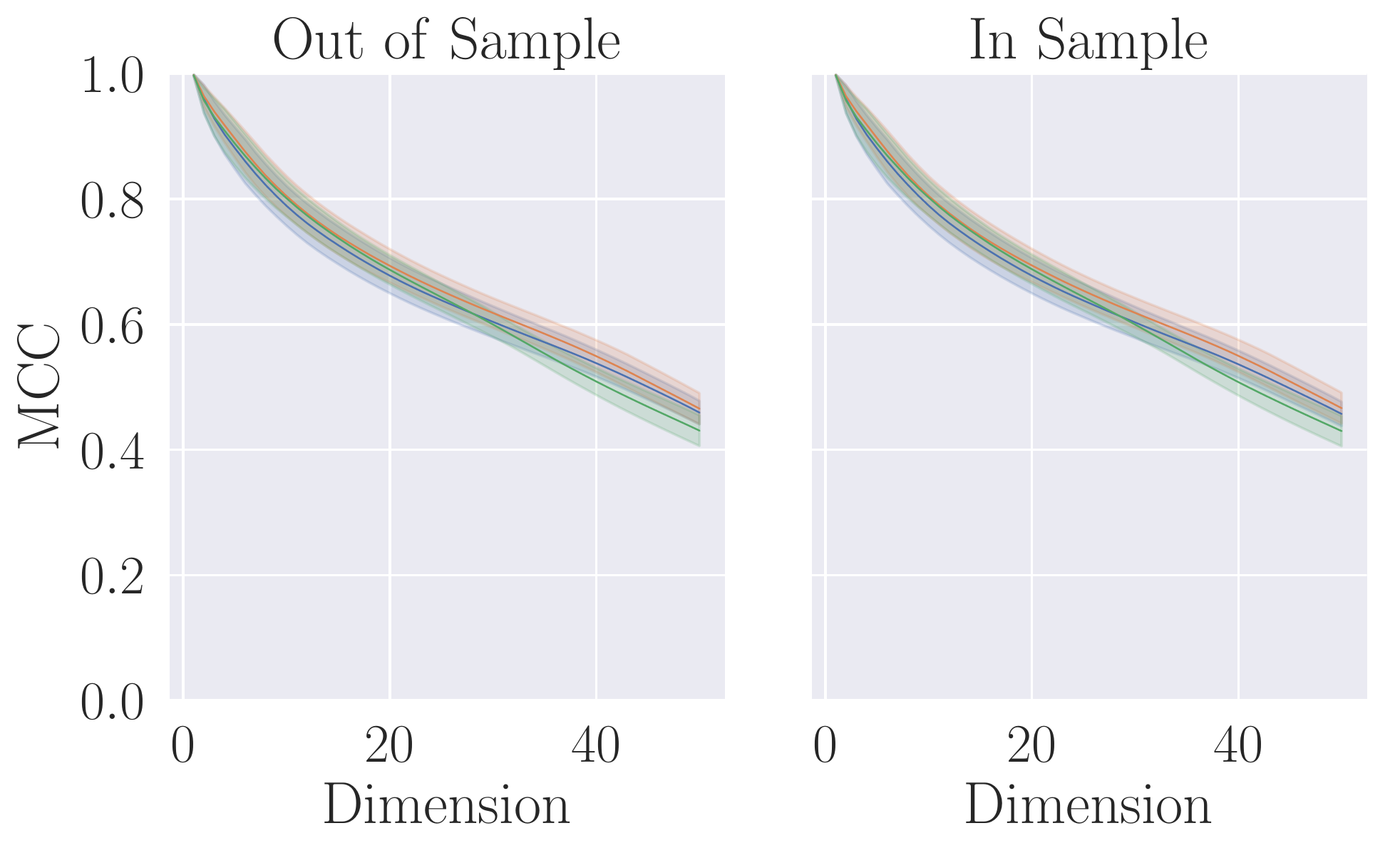}&
\includegraphics[height=3.0cm]{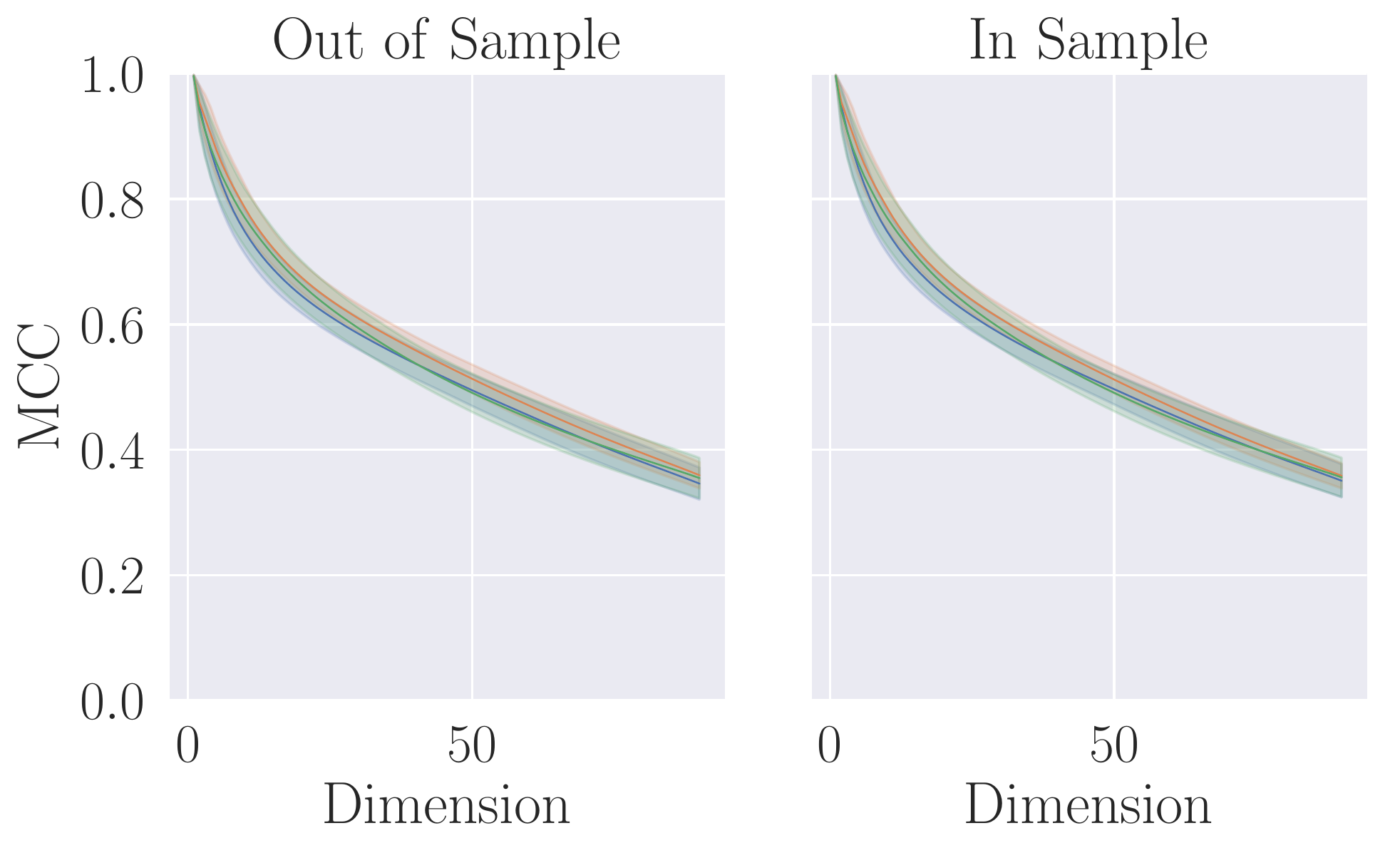}&
\includegraphics[height=3.0cm]{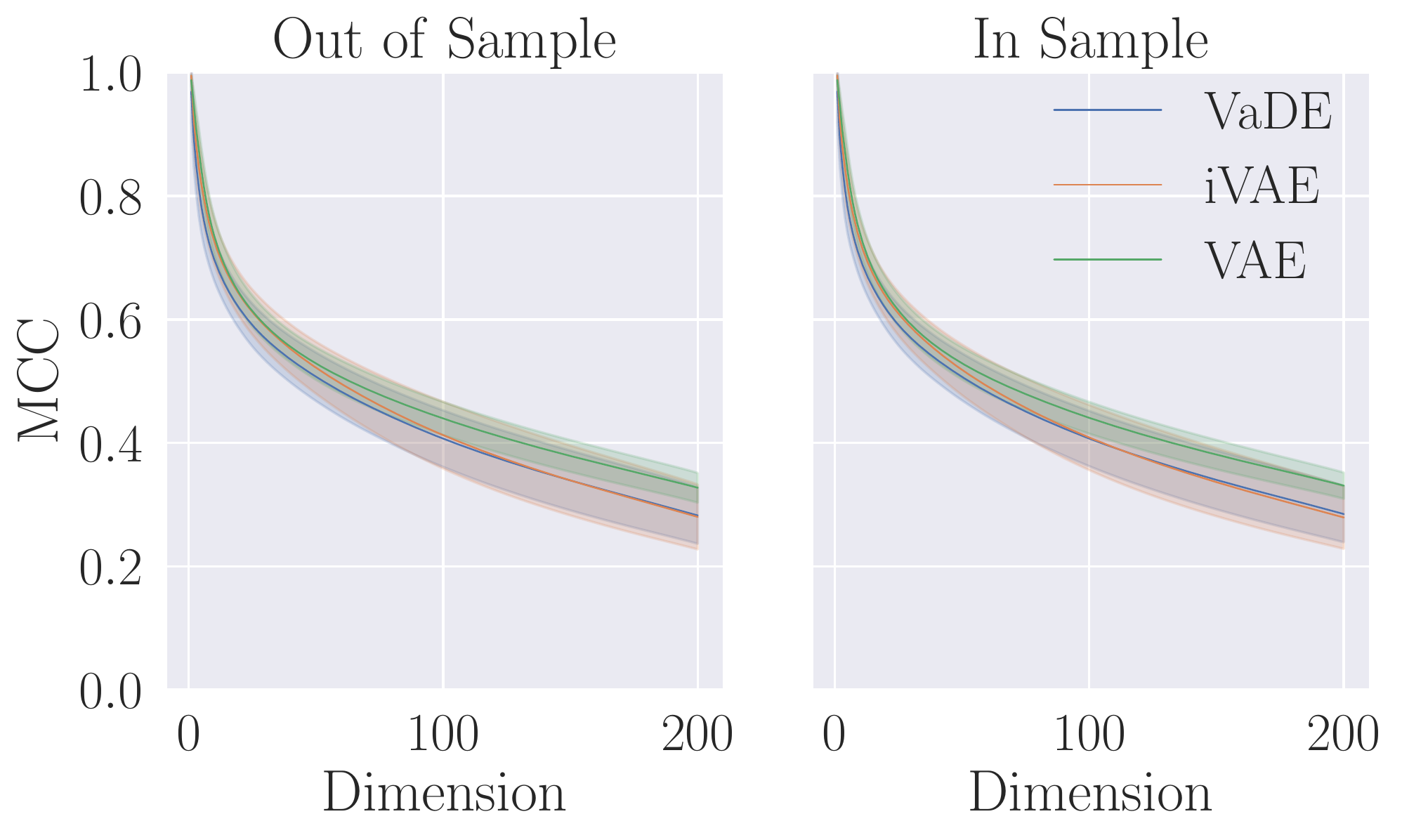}\\
\end{tabularx}
\caption{The same as Fig~\ref{fig:strong_mlp_mcc}, but for ConvNet models, with $d_z\in \{50,90,200\}$.
}
\label{fig:strong_conv_mcc}
  \end{figure*}
  
  \begin{figure*}
      \centering
    \centering
\begin{tabularx}{\textwidth}{cCCC}
\centering
& \hspace{2em} $d_z=50$  & \hspace{2.4em} $d_z=90$  & \hspace{2.3em} $d_z=200$  \\
\raisebox{2.1\height }{\rotatebox[origin=c]{90}{CIFAR10}}&
\includegraphics[height=3.0cm]{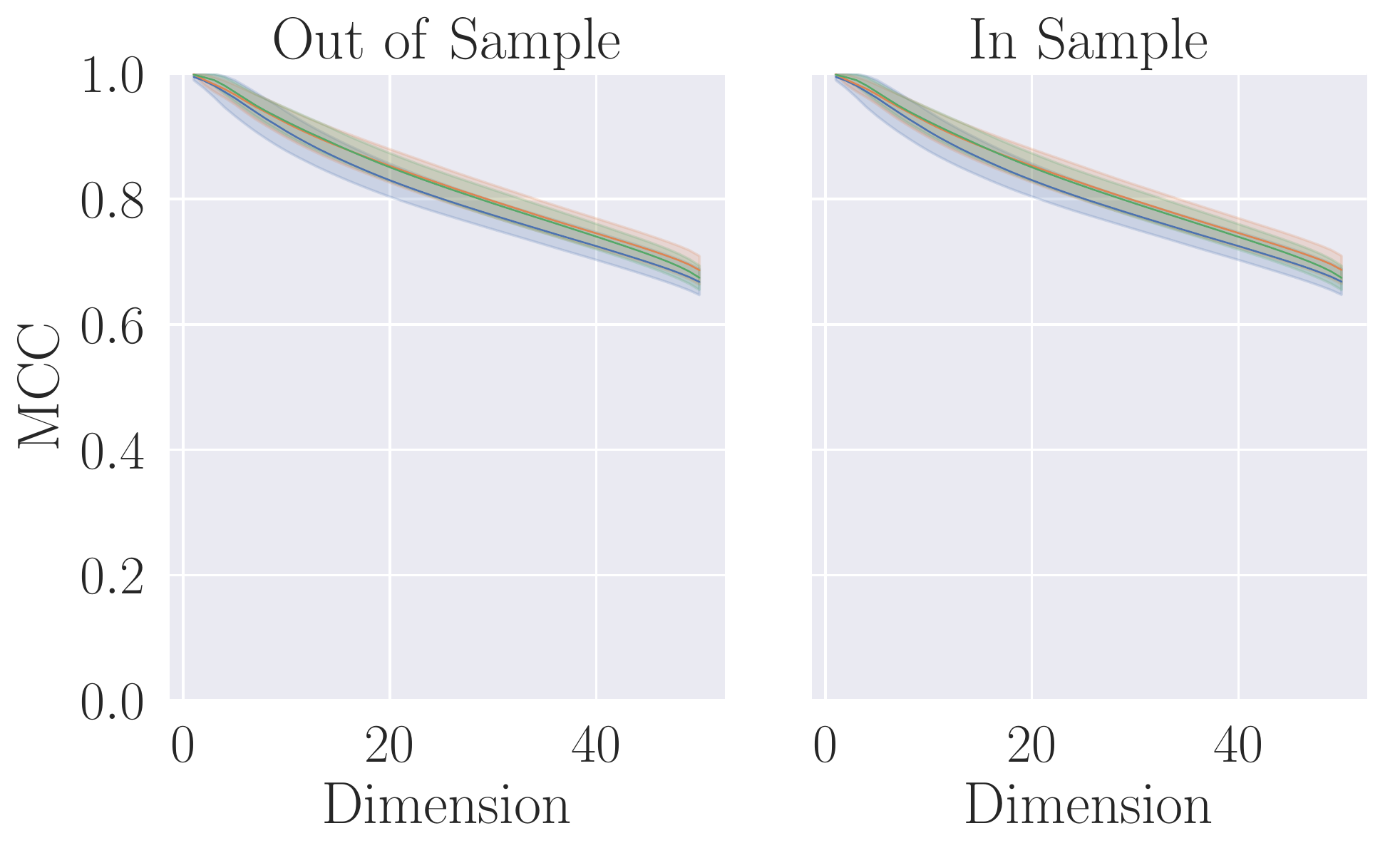}&
\includegraphics[height=3.0cm]{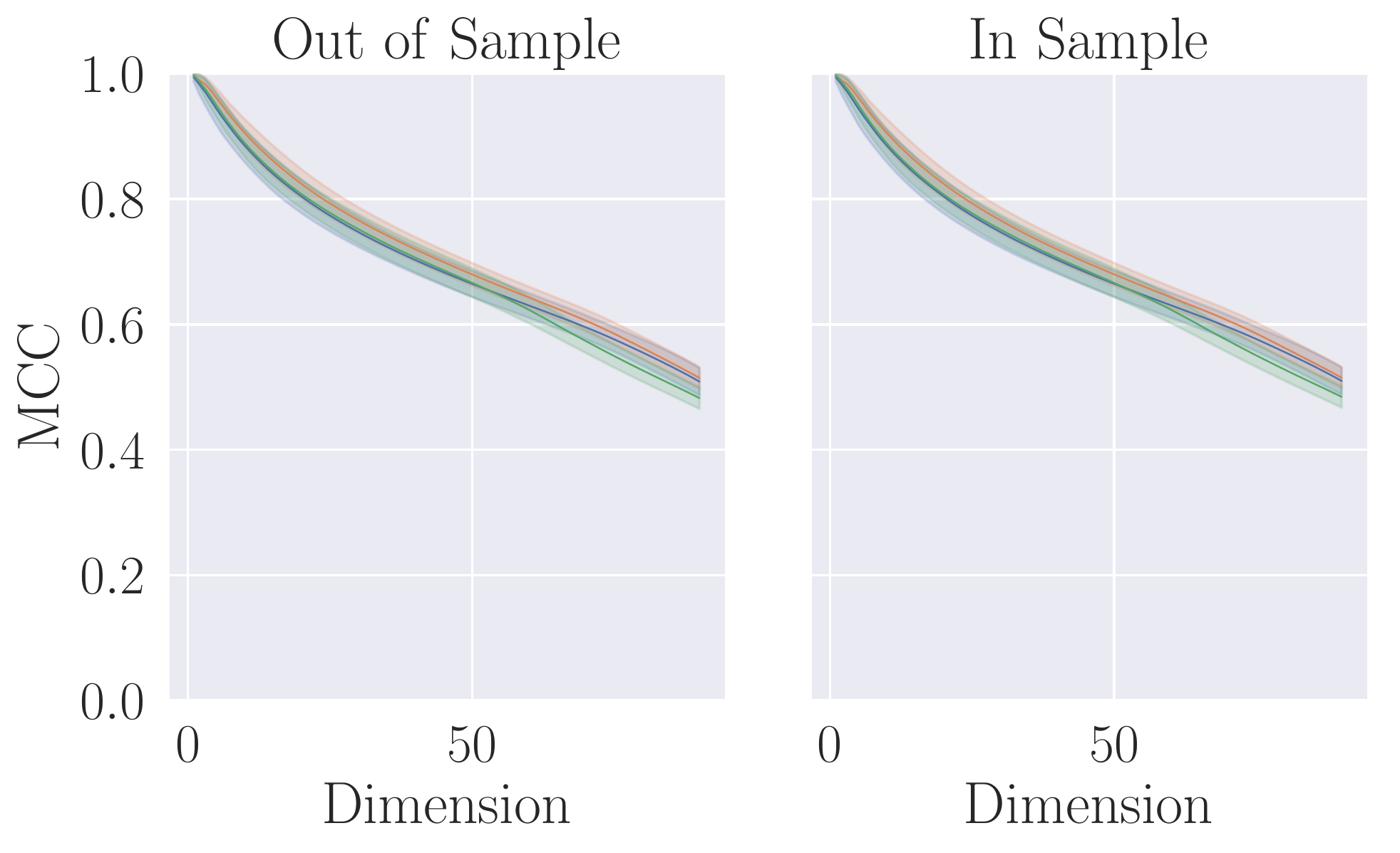}&
\includegraphics[height=3.0cm]{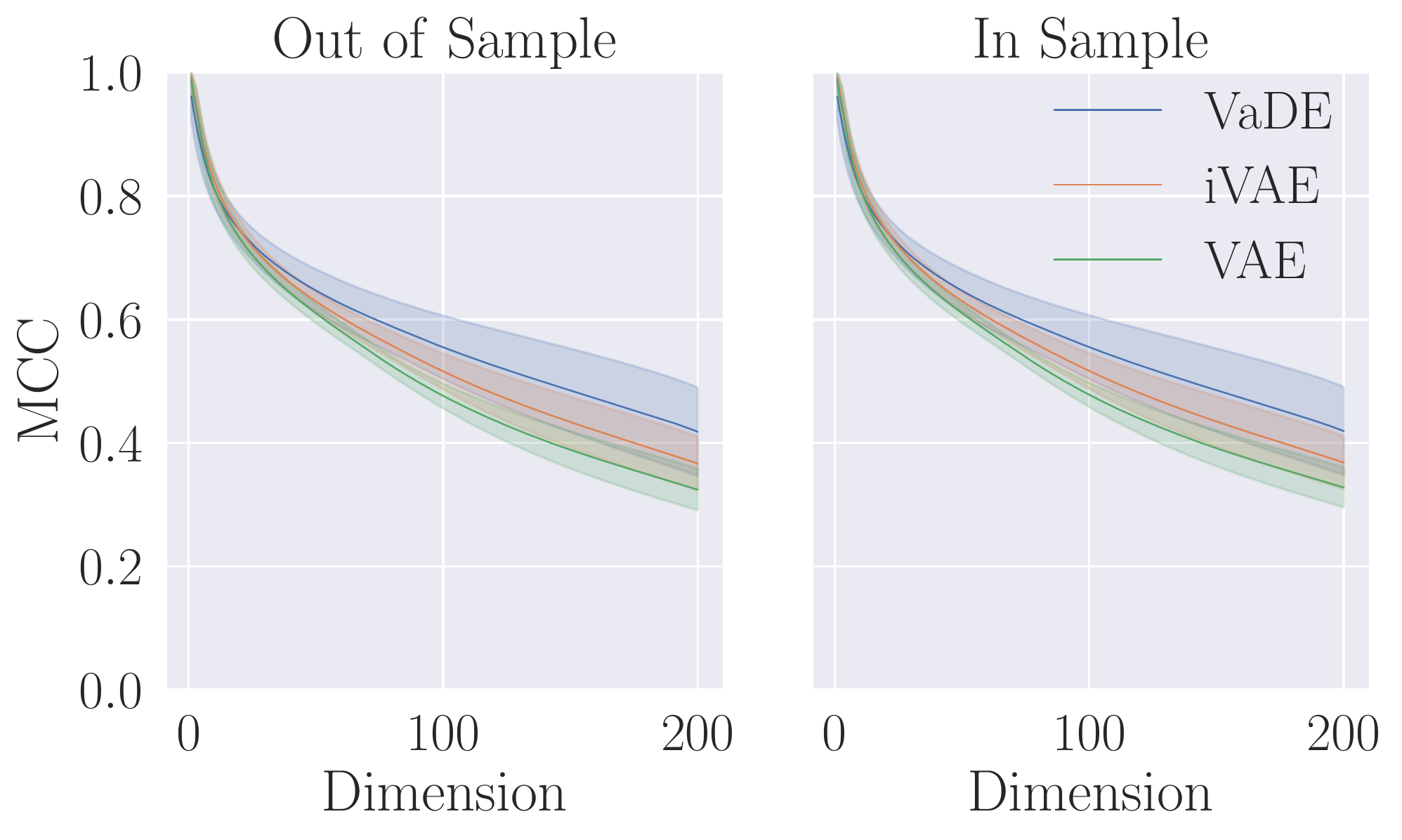}\\
\raisebox{2.8\height }{\rotatebox[origin=c]{90}{SVHN}}&
\includegraphics[height=3.0cm]{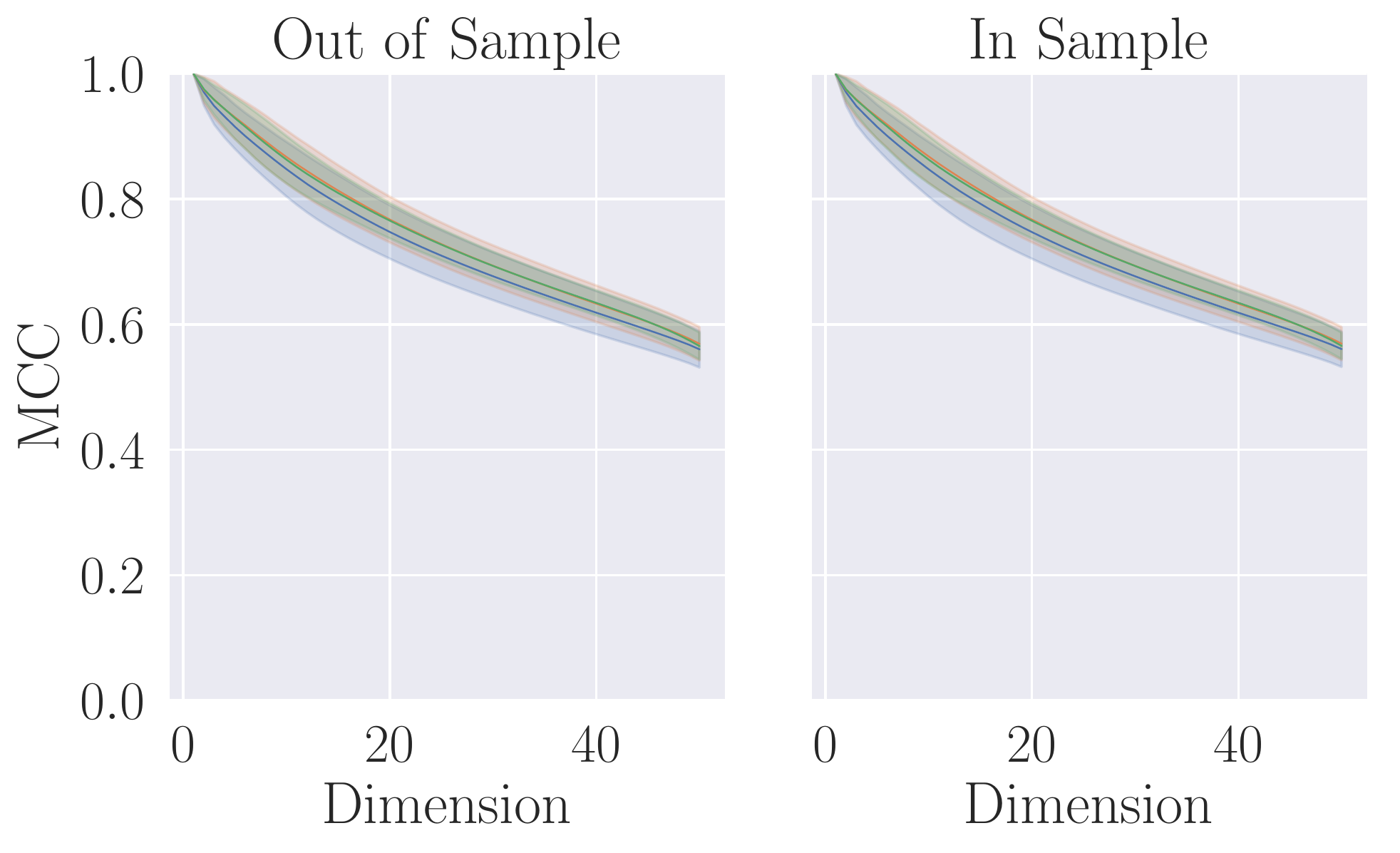}&
\includegraphics[height=3.0cm]{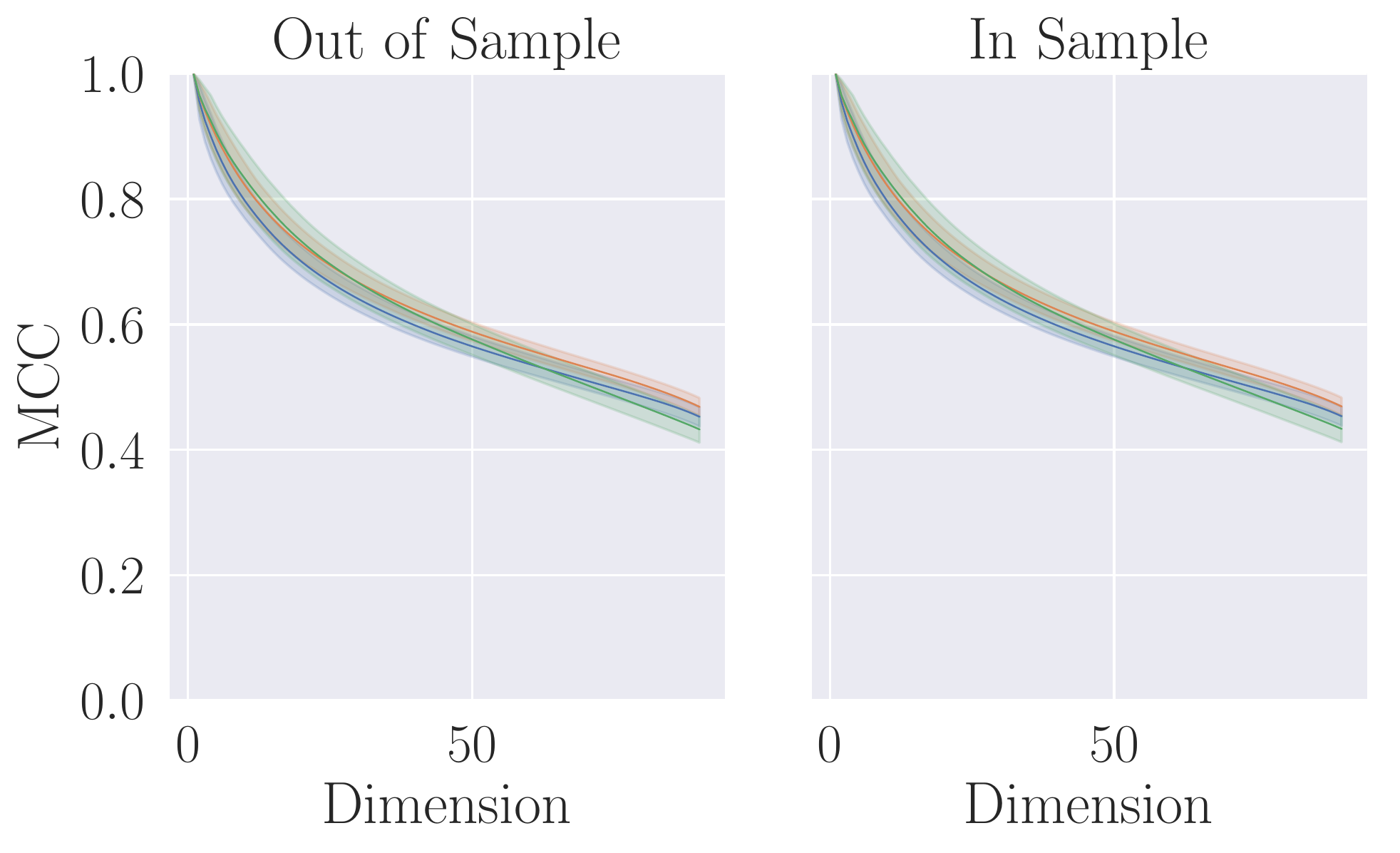}&
\includegraphics[height=3.0cm]{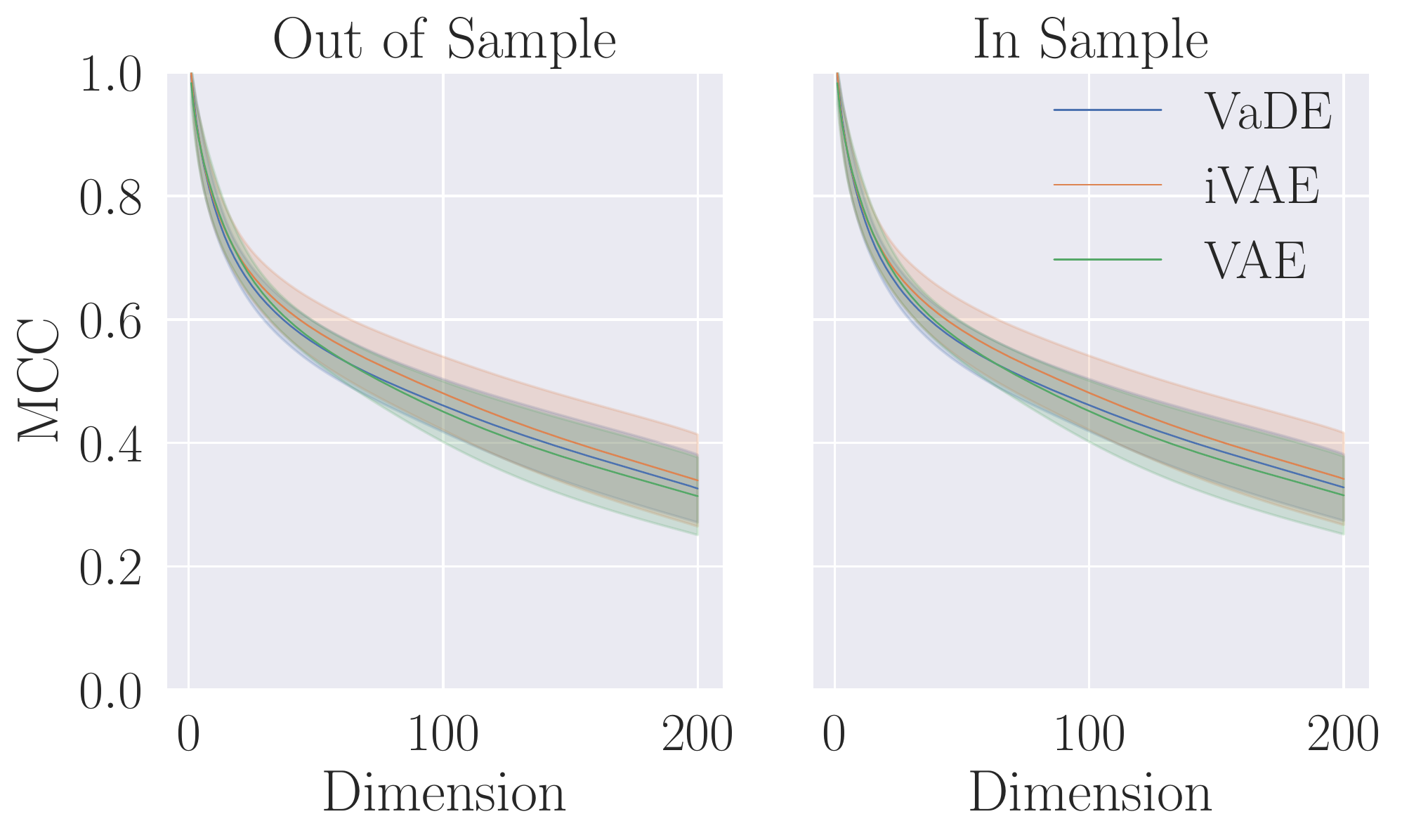}\\
\end{tabularx}
\caption{The same as Fig~\ref{fig:strong_mlp_mcc}, but for ResNet models, with $d_z\in \{50,90,200\}$.
}
\label{fig:strong_res_mcc}
  \end{figure*}

\subsection{Image Data}
We are particularly interested in the identifiability of VaDEs and vanilla VAEs, benchmarked against iVAEs, when trained on standard image datasets and over a range of architectures.
We train iVAEs, VaDEs and VAEs with the encoders and decoders being MLPs, ConvNets and ResNets.
With these architectures we are trying to capture a range of model richness and representational power.
See Appendix~\ref{app:archs} for detailed descriptions of the neural architectures.
We trained 480 DGMs for the experiments in this section\footnote{With each run taking on average $\sim8$h on an M60 GPU, this means that if performed in series these experiments would take $\sim5$ months.}.

For each model-architecture-dataset combination we train $d_z\in\{50,90,200\}$, and for each of these we train with 10 different seeds.
We train on MNIST, CIFAR10 and SVHN.
For iVAE, we use the provided class label as $\v u$; for VaDE we introduce a 40-component mixture model for MNIST and a 100-component one for CIFAR-10 and SVHN.

We wish to underline that we carried out no architecture tuning or tuning of optimisation method, beyond finding a learning rate that provided stable optimisation in terms of training set $\ELBO$.
Rather, we implemented three reasonable architectures and then trained them, as iVAEs, VaDEs and VAEs.
The idea in these experiments is not to tune some lucky architecture that happens to lead to consistently-learnt representations, but to test how robust learnt VaDE, VAE and iVAE representations are to different neural parameterisations.
All training was done with a batch size of 64 using ADAM.
We had to make small changes to the models to handle different datasets: 1) those needed to handle $28\times28\times1$ images (MNIST) vs $32\times32\times3$ images (CIFAR10 \& SVHN) and 2) the choice of likelihood function, Bernoulli for MNIST and Logistic~\cite[\S C.5]{Kingma} for CIFAR10 \& SVHN.

As we do not have recorded the true underling latent sources that generated these images, in order to evaluate these models we follow~\citep{Khemakhem2020} in measuring the MCC between matching runs with different seeds.
For 10 different random restarts we get 55 pairs of seeds.
To measure how our clustering approach compares to iVAEs, we test both strong ($\sim_P$) and weak ($\sim_A$) formulations.

In the case of testing strong identifiability, where we calculate the MCC under an optimal alignment of coordinate directions between the two sets of representations, we give values as cumulative means over match coordinates, in descending order of MCC values.
We show these results in Figures~\ref{fig:strong_mlp_mcc}-\ref{fig:strong_res_mcc}.
Some of our $d_z$ values, as a proportion of $d_x$, are relatively large here; for example, for $d_z=200$ on MNIST with $d_x=784$, $\frac{d_z}{d_x}\approx\frac{1}{4}$.
We might expect that some subspace in $\mathcal{Z}$ could contain mere noise, as is observed in iFlows~\citep{iflows} where $d_z=d_x$.

Recall that in the case of weak identifiability one learns an affine transform using CCA to maximally-align two sets of learnt representations.
As such we follow~\cite{Khemakhem2020} (where also $d_z=d_x$) in learning a maximally-aligned space that is of lower dimensionality than the raw representations themselves.
As in~\cite{Khemakhem2020}, we have $d_\mathrm{CCA}=20$.
We show the results for all these models in Figures~\ref{fig:mlp_mcc}-\ref{fig:res_mcc} for affine-aligned representations ($\sim_A$).
In these experiments, for MLPs VaDE consistently reaches higher MCC values than iVAEs do.
For ConvNets and ResNets, performance is very similar between VaDE and iVAE.

For each formulation, strong and weak, the optimal permutation/affine alignment for each pair of runs is calculated over half the test set (`In Sample') and re-used for the second half (`Out of Sample').

To summarise, the results we show (Figs~\ref{fig:strong_mlp_mcc}--\ref{fig:strong_res_mcc} for strong identifiability, Figs~\ref{fig:mlp_mcc}--\ref{fig:res_mcc} for weak identifiability) indicate that
VaDEs and VAEs learn representations which score similarly to iVAEs in terms of MCC, despite being trained unsupservised, across a range of datasets and model architectures.
(Given the qualitative similarity of these curves, we perform statistical analysis in the next section.)

Empirically, these results indicate identifiability, as measured by MCC, that is similar for iVAEs, VaDEs and VAEs.
As well as providing empirical evaluation of VaDEs and VAEs, this is also the first large-scale evaluation of the identifiability of iVAEs on image data.

\subsubsection{Statistical Testing} We performed a two-sided Wilcoxon signed-rank test on the final MCC values (i.e. the MCC calculated over all dimensions in the latent space, the rightmost values in each plot in Figs~\ref{fig:strong_mlp_mcc}--\ref{fig:strong_res_mcc}) to see if VaDE, VAE and iVAE MCC values can be mutually distinguished, i.e. for every pair test whether each model's values are drawn from the same distribution.
We find that we cannot reject the null hypothesis that VaDE and iVAE MCC values come from the same distribution as we get $p=0.422$ under this test.
Performing the same statistical test, we can, however, reject the null hypothesis that VAE and iVAE MCCs come from the same distribution with $p=0.029$, and similarly for VAE and VaDE MCCs with $p=0.039$.
See Appendix~\ref{app:stats} for more information.
These results suggest that VaDE MCCs are indistinguishable from those from iVAEs, despite VaDE's lack of side information. 
These results further support the improved identifiability of iVAE and VaDE relative to a vanilla VAE.

\section{Related Work}
\label{sec:relatedwork}
Our discussion of the range of potential $\v u$-tasks in the context of non-linear ICA is mirrored in the deep clustering literature, that for a given dataset there are multiple ways that one could plausibly cluster the data~\citep{multidimcluster,Willetts2019d,mfcvae}.
For instance, for MNIST one might aim to cluster over digit identity, but also one could aim to cluster over stroke thickness, or the left-right slant of the written digits, or some more baroque task like how many genera (in a topological sense) each written digit contains. (This last task would, for instance, separate the two ways of writing `4'.)

This phenomena, that different clustering algorithms lead to different natural clusterings of the data, has been discussed in the context of semi-supervised learning~\citep{Nigam2000,Corduneanu2002}.
This is a problem in that setting, as then additional unlabelled data can degrade rather than improve classification performance.
In terms of Figure~\ref{fig:clustering}, the situation would be having a small amount of data annotated with the shape of objects, say, but a model that naturally partitions over colour.
For us, however, any informative clustering of the data serves our purposes as long as it is consistently learnt by the model and fulfills the technical requirements of~\cite{Khemakhem2019}.

As mentioned above, classifier-based approaches to non-linear ICA~\citep{tcl,Hyvarinen2019} are the cousins of contrastive methods for self-supervised learning~\citep{Gutmann2012,cpc,dataefficientcpc,moco}.
The $\v u $-tasks used in the context of non-linear ICA are exactly the tasks one can use in contrastive self-supervised learning, and vice versa.
Examples include separating the channels of colour images and aiming to match the pair of different channels that came from the same image~\citep{cpc}, or matching different modalities or augmentations of the same datapoint~\citep{Wu2018a, Bachman2019, Chen2020, moco}.
See~\cite{makesgood} for a recent analysis of what makes a good $\v u$-task in contrastive learning.
Further, the connection between contrastive learning and classifier-based methods to non-linear ICA has been formalised in the `Incomplete Rosetta stone problem'~\citep{Gresele2019}.
Recently~\cite{hmmica} have shown that unsupervised non-linear ICA is possible when the data is not i.i.d. but instead has some temporal structure.

There is a strand, also, of self-supervised learning that aims to learn a clustering over the data simultaneously with the representations~\citep{Yang2016a, Caron2018, Huang2019, Zhuang2019, selflabelling, Caron2020}.
These approaches so far are mostly theoretically-distinct from constrastive methods (though \cite{Caron2020} is perhaps the start of a unification).
We think that two lenses we explore, clustering DGMs and non-linear ICA, might provide a way for the two self-supervised approaches of clustering and contrastive methods to be unified.
Finally, recently work has been done that explicitly aims to learn two representations simultaneously where the objective is to minimise the cross-correlation between the pair of representations~\citep{barlowtwins, vicreg}, which is done in our work and others~\citep{Khemakhem2020} using linear canonical-correlation analysis to find alignment (prior to calculating MCC values) between the representations of random restarts of a model.

\section{Conclusion}
Here we have shown how DGMs with no side information can score similarly to iVAEs in terms of MCC values, across a range of datasets and neural parameterisations.
In this analysis we have found that the inductive biases of DGMs are sufficient to induce structure in their latent representation that itself is sufficiently consistent run-to-run to provide consistently-learnt representations, as measured by MCC values.
This work opens the question of when our parametric models are identifiable in practice.

\newpage
\bibliographystyle{apalike2}
\bibliography{references.bib}

\newpage

 \onecolumn
\appendix
\setcounter{equation}{0}
\renewcommand\theequation{\thesection.\arabic{equation}}
\setcounter{figure}{0}
\renewcommand\thefigure{\thesection.\arabic{figure}}
\setcounter{table}{0}
\renewcommand{\thetable}{\thesection.\arabic{table}}
 \hsize\textwidth
  \linewidth\hsize {\centering
  {\Large\bfseries Appendix\par}}

\section{Synthetic Data}
\label{app:synth}
\begin{figure}[h]
    \centering
    \includegraphics[width=0.45\textwidth]{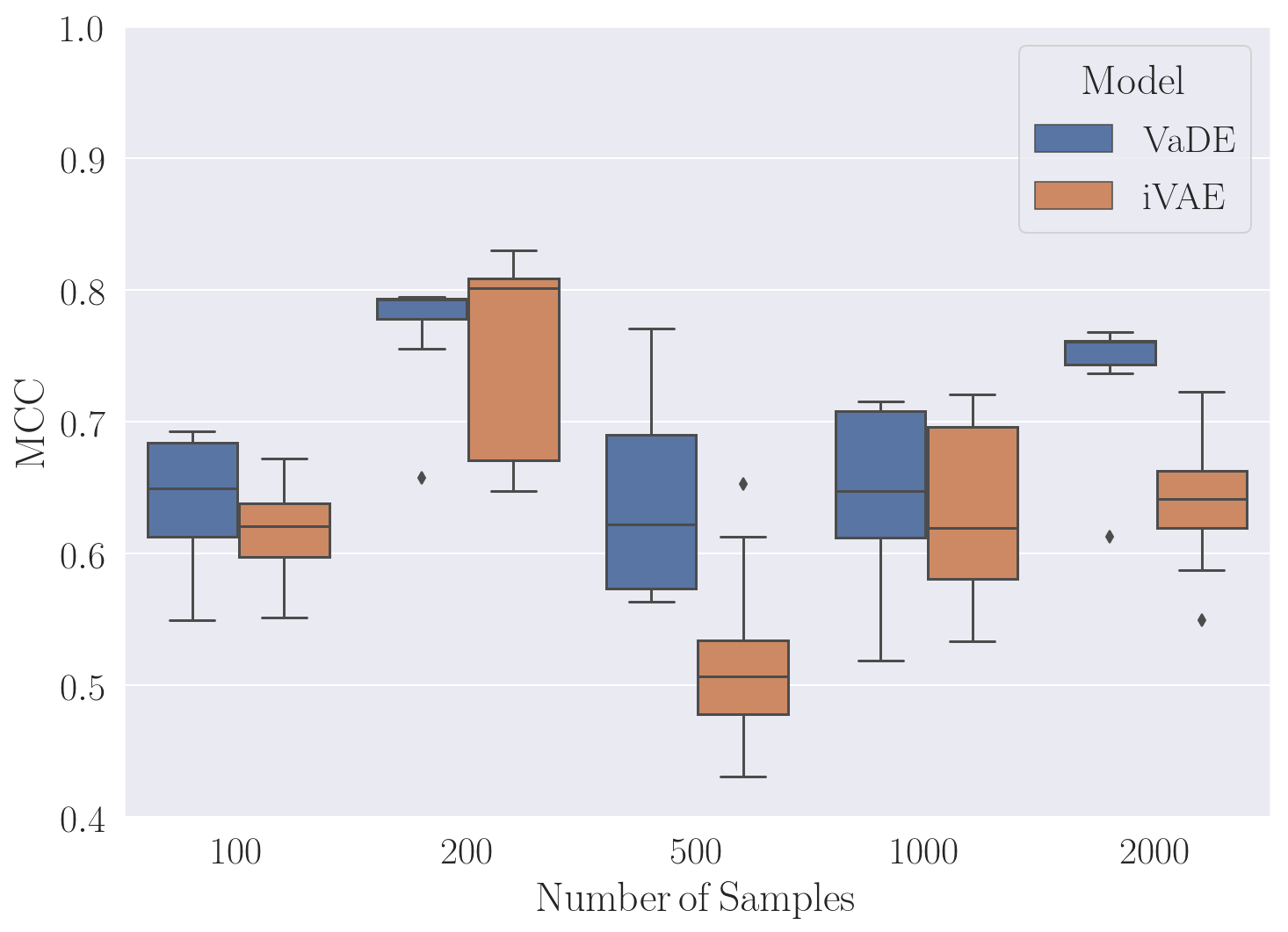}
    \caption{MCC for VaDE and iVAE for synthetic data generated using $L=4$ mixing layers.}
    \label{fig:l4_mcc}
\end{figure}
We create synthetic data to evaluate model performance using the same generating mechanism and mode of analysis as in previous in works on non-linear ICA~\citep{tcl, Hyvarinen2019, Khemakhem2019,gin, iflows, Khemakhem2020}.
We generate Time-Constrastive Learning data with $d_x=d_z=5$, constructing data in 20 segments.
Each segment's data is made by sampling from a random product of Gaussians with means with uniformly distributed on $[-3, 3]$ and standard deviations uniformly distributed on $[0.01, 3]$.
These sources were mixed by a 4-layer random MLP, where the weight matrices were constructed to be full-rank, with $\mathrm{LeakyRelu}$ activations functions.
We constructed five different datasets, varying by the number of samples per segment in $\{100,200, 500, 1000,2000\}$.

We trained iVAE and VaDE models 10 times with different seeds on these five datasets, with iVAEs and VaDEs using the same neural architectures as each other for both the encoder and decoder.
For both networks we used 3-layer MLPs with $\mathrm{LeakyRelu}$ activations functions and trained using ADAM~\citep{Kingma2015} for 70,000 steps, with an initial learning rate of $0.001$ that we decay on plateau.
For VaDE $p_\theta(\v z)$ is a 40-component Gaussian mixture model, with the means and log variances Xavier-uniformly initialised~\citep{glorot}.

We show boxplots of the resulting MCC values, Figure~\ref{fig:l4_mcc}.
We find that VaDE achieves MCC results that are as high or higher than iVAE does, indicating that it is as successful or more successful at discovering the true latents as iVAE is.

\begin{figure}[h]
\centering
    \includegraphics[width=1.0\textwidth]{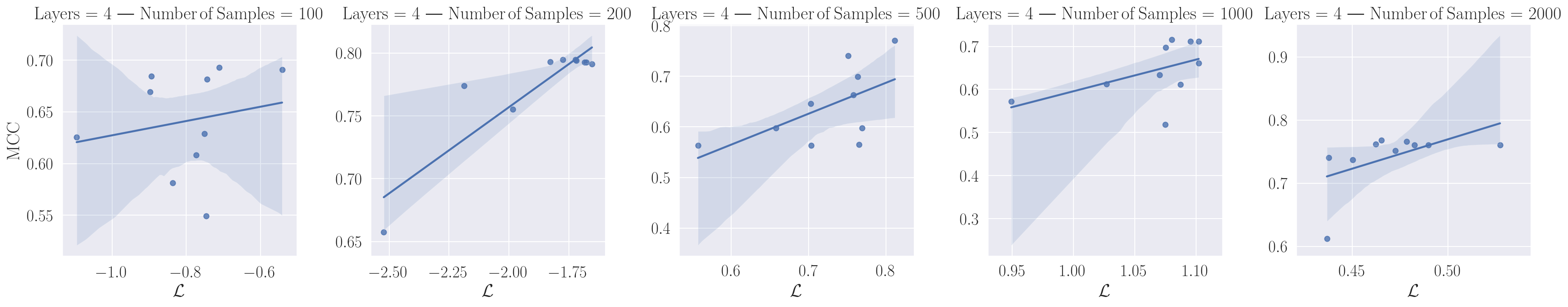}
    \caption{MCC (to ground-truth sources) as a function of $\ELBO$ for VaDE models trained on synthetic data, for five different datasets of different sizes.
    Different points correspond to different random restarts. Shading is over the 95\% confidence interval.}
    \label{fig:l4_elbo}
\end{figure}

We can also ask, is model quality, as measured by $\ELBO$ on the training data, a predictor of successful non-linear un-mixing in VaDE models on this task?
We plot $\ELBO$ against MCC for VaDE, Figure~\ref{fig:l4_elbo}, and find that yes $\ELBO$ and MCC are positively correlated.

\newpage
\section{Neural Network Architectures}
\label{app:archs}
Here we describe the neural network architectures used for our experiments.
We include code to reproduce our experiments in the supplementary material.

We trained iVAEs and VaDEs on CIFAR10, SVHN and MNIST, using MLPs, ConvNets and ResNets for each of the encoder and decoder.
\begin{enumerate}
  \item \label{arch3} \emph{MLP}: A series of fully connected layers, for encoder and decoder.
  \item \label{arch2} \emph{ConvNet}: A series of convolutional layers followed by some fully connected layers for encoder. For decoder, some fully connected layers followed by a series of transposed convolutional layers.
  \item \label{arch1}  \emph{ResNet}: A series of downsampling ResNet blocks followed by some fully connected layers for encoder. For decoder, some fully connected layers followed by a series of upsampling ResNet blocks.
\end{enumerate}

For detailed breakdowns of MLP architectures see Table~[\ref{tab:configs_MLP}], for ConvNets Table~[\ref{tab:configs_conv}] and for ResNets Table~[\ref{tab:configs_res}]

For each neural parameterisation, we chose the dimensionality of the latent space $d_z$ to be $\in\{50,90,200\}$, as discussed in the main paper.
For both iVAEs and VaDEs we chose the set of conditionals $\{p_\theta(\v z | \v u)\}_{\v u \in\mathcal{U}}$ to be Gaussians with diagonal covariance.

In all experiments we trained using ADAM \citep{Kingma2015}, with $(\beta_1, \beta_2) = (0.9, 0.999)$.
We used mini-batches sizes of $64$ in all runs, and we trained for 200 epochs, decaying learning rate on plateau.
Our code is based on that included with papers~\citep{Khemakhem2019,Khemakhem2020}, and is released under a GPL v3.0 license.

\begin{table}[h!]
\begin{minipage}{.5\linewidth}
\caption{\label{tab:configs_data} Dataset parameters}
\centering
\begin{tabular}{c}
\toprule
Sizes\\
\midrule
  Data: $d_x = w\times w\times n_c$\\
  $n_c$: channels, $w$: width \\
  MNIST: $n_c=1$, $w=28$ \\
  CIFAR10: $n_c=3$, $w=32$ \\
  SVHN: $n_c=3$, $w=32$ \\
\bottomrule
\end{tabular}
\end{minipage}
\begin{minipage}{.5\linewidth}
\caption{\label{tab:configs_MLP} MLP Architecture in detail}
\centering
\begin{tabular}{c l}
\toprule
Architecture & Layer Sequence\\
 \midrule
\emph{MLP Encoder} & Input: $d_x=w \times w \times c$  \\
 \midrule
  & FC $512$, LeakyReLU($0.1$)  \\
  & FC $384$, LeakyReLU($0.1$)  \\
  & Dropout($0.1$)  \\
  & FC $256$, LeakyReLU($0.1$)  \\
  & FC $256$, LeakyReLU($0.1$)  \\
  & $\mu_\phi(\v x)$: FC $d_z$  \\
  & $\sigma_\phi(\v x)$: FC $d_z$  \\

\midrule
\emph{MLP Decoder} & Input: $d_z$ \\
  & FC $256$, LeakyReLU($0.1$)  \\
  & FC $256$, LeakyReLU($0.1$)  \\
  & Dropout($0.1$)  \\
  & FC $384$, LeakyReLU($0.1$)  \\
  & FC $512$, LeakyReLU($0.1$)  \\
  & FC $d_x$  \\
\bottomrule
\end{tabular}
\end{minipage}
\end{table}

\begin{table}
\caption{\label{tab:configs_conv} ConvNet Architecture in detail}
\centering
\begin{tabular}{c l r}
\toprule
Architecture & Layer Sequence & Notes\\
 \midrule
\emph{ConvNet Encoder} & Input: $d_x=w \times w \times c$ & stride $1$ for convs, unless stated\\
  & Conv $w\times w \times 8$, BatchNorm, ELU & padding $1$, filter size $3$ \\
  & Conv $w\times w \times 16$, BatchNorm, ELU & padding $1$, filter size $3$ \\
  & MaxPool $\frac{w}{2}\times \frac{w}{2} \times 16$ & \\
  & Conv $\frac{w}{2}\times \frac{w}{2} \times 32$, BatchNorm, ELU & padding $1$, filter size $3$ \\
  & Conv $\frac{w}{2}\times \frac{w}{2} \times 64$, BatchNorm, ELU & padding $1$, filter size $3$ \\
  & MaxPool $\frac{w}{4}\times \frac{w}{4} \times 64$ & \\
  & Conv $1\times1 \times 64$ & padding $0$, filter size $\frac{w}{4}$ \\
  & FC $64$, LeakyReLU($0.1$) & \\
  & $\mu_\phi(\v x)$: FC $d_z$ & \\
  & $\sigma_\phi(\v x)$: FC $d_z$ & \\
 \midrule
 \emph{ConvNet Decoder} & Input: $d_z$& stride $1$ for convs, unless stated\\
   & FC $64$, LeakyReLU($0.1$) & \\
   & FC $64$ & \\
  & ConvTranspose $\frac{w}{4}\times\frac{w}{4} \times 64$, BatchNorm, ELU & padding $0$, filter size $\frac{w}{4}$\\
  & ConvTranspose $\frac{w}{2}\times\frac{w}{2} \times 32$, BatchNorm, ELU & padding $1$, filter size $4$, stride $2$ \\
  & ConvTranspose $\frac{w}{2}\times\frac{w}{2} \times 16$, BatchNorm, ELU & padding $1$, filter size $3$ \\
  & ConvTranspose $w \times w \times 8$, BatchNorm, ELU & padding $1$, filter size $4$, stride $2$  \\
    & Conv $w\times w \times c$ & padding $1$, filter size $3$ \\
\bottomrule
\end{tabular}
\end{table}

\begin{table}
\caption{\label{tab:configs_res} ResNet Architecture in detail. ResNet blocks all have weightnorm applied to them, and in encoder downscale by a factor of $2$ spatially, in the decoder they upscale by a factor of $2$.}
\centering
\begin{tabular}{c l r}
\toprule
Architecture & Layer Sequence & Notes\\
 \midrule
\emph{ResNet Encoder} & Input: $d_x=w \times w \times c$ & stride $1$ for all conv. layers\\
  & Conv $w\times w \times 16$ & padding $0$, filter size $1$ \\
  & ResNet Block $\frac{w}{2}\times \frac{w}{2} \times 32$ \\
  & ResNet Block $\frac{w}{4}\times \frac{w}{4} \times 64$ \\
  & ResNet Block $\frac{w}{8}\times \frac{w}{8} \times 128$ \\
  & ResNet Block $\frac{w}{16}\times \frac{w}{16} \times 256$ \\
  & ResNet Block $\frac{w}{32}\times \frac{w}{32} \times 512$, ReLU \\
  & $\mu_\phi(\v x)$: FC $d_z$ & \\
  & $\sigma_\phi(\v x)$: FC $d_z$ & \\
 \midrule
 \emph{ResNet Decoder} & Input: $d_z$\\
   & FC $512$\\
  & ResNet Block $\frac{w}{16}\times \frac{w}{16} \times 256$ \\
  & ResNet Block $\frac{w}{8}\times \frac{w}{8} \times 128$ \\
   & ResNet Block $\frac{w}{4}\times \frac{w}{4} \times 64$ \\
    & ResNet Block $\frac{w}{2}\times \frac{w}{2} \times 32$ \\
  & ResNet Block $w\times w \times 16$, ReLU \\
   & Conv $w\times w \times c$ & padding $0$, filter size $1$ \\

\bottomrule
\end{tabular}
\end{table}

\clearpage
\newpage
\section{Weak-Identifiability Results}
\vspace{10em}
\begin{figure*}[h!]
    \centering
\begin{tabularx}{\textwidth}{cCCC}
\centering
& \hspace{2em} $d_z=50$  & \hspace{2.4em} $d_z=90$  & \hspace{2.3em} $d_z=200$  \\
\raisebox{2.4\height }{\rotatebox[origin=c]{90}{MNIST}}&
\includegraphics[height=3.0cm]{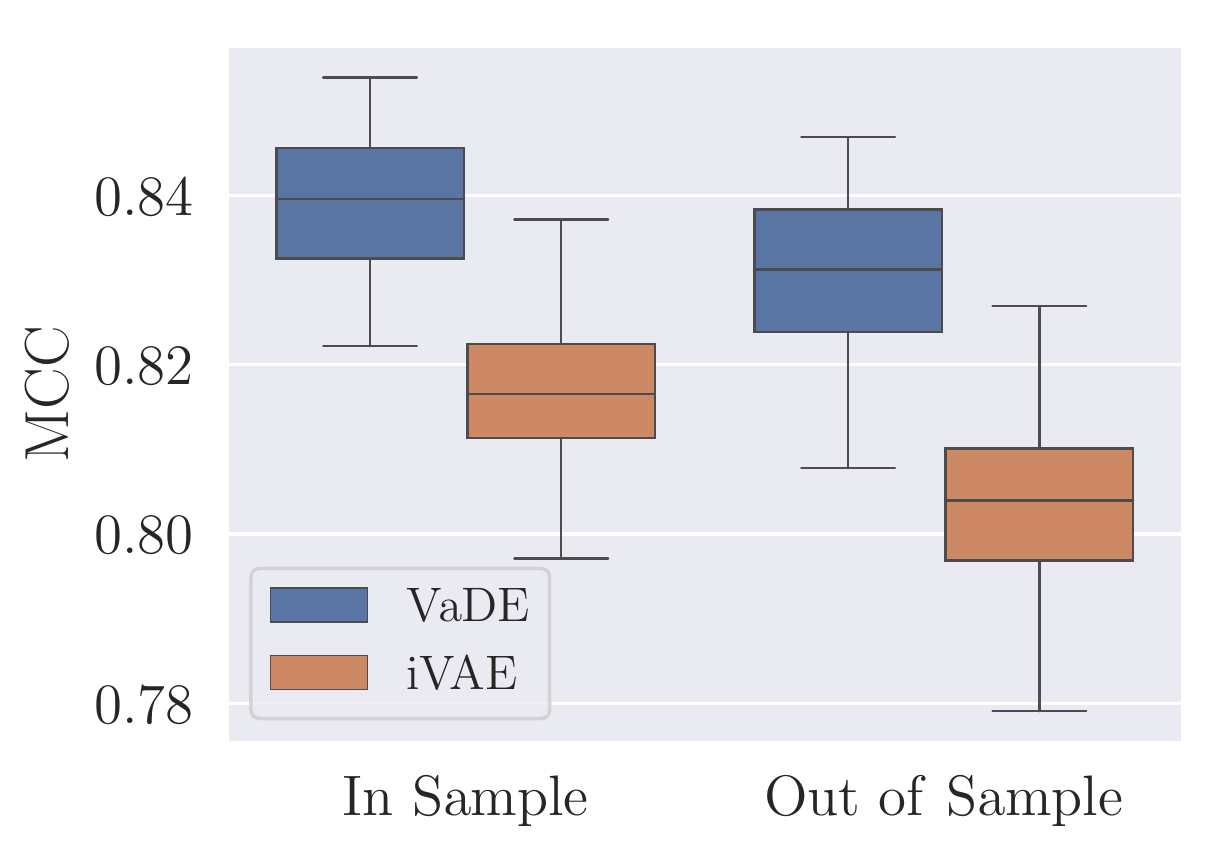}&
\includegraphics[height=3.0cm]{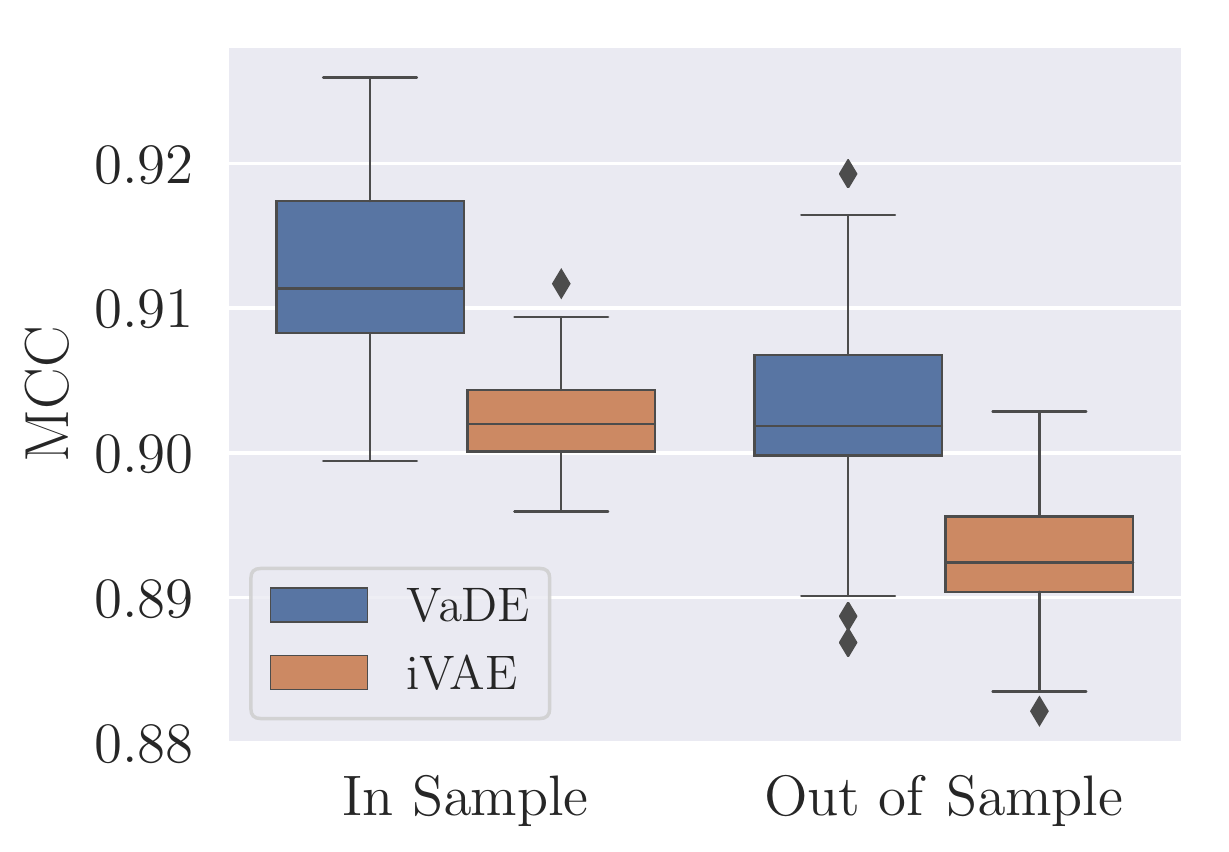}&
\includegraphics[height=3.0cm]{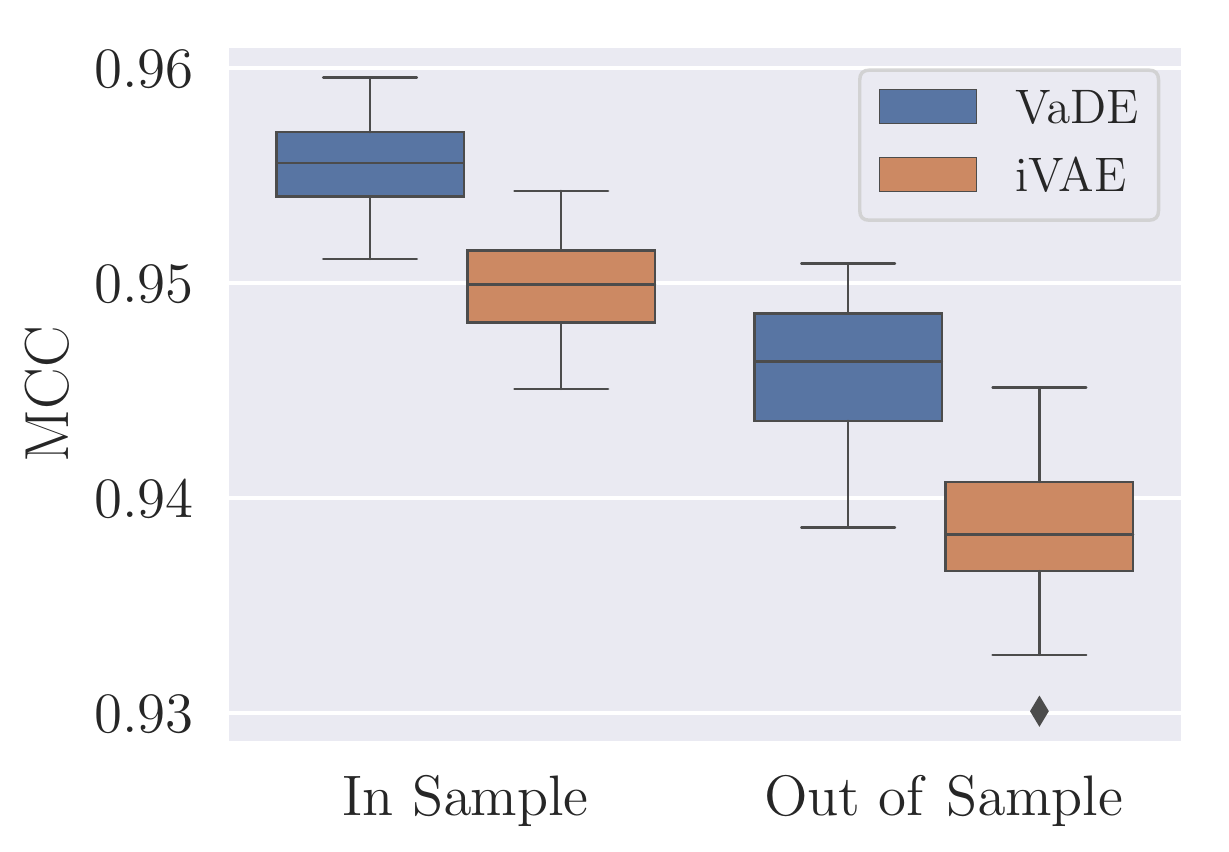}\\
\raisebox{2.1\height }{\rotatebox[origin=c]{90}{CIFAR10}}&
\includegraphics[height=3.0cm]{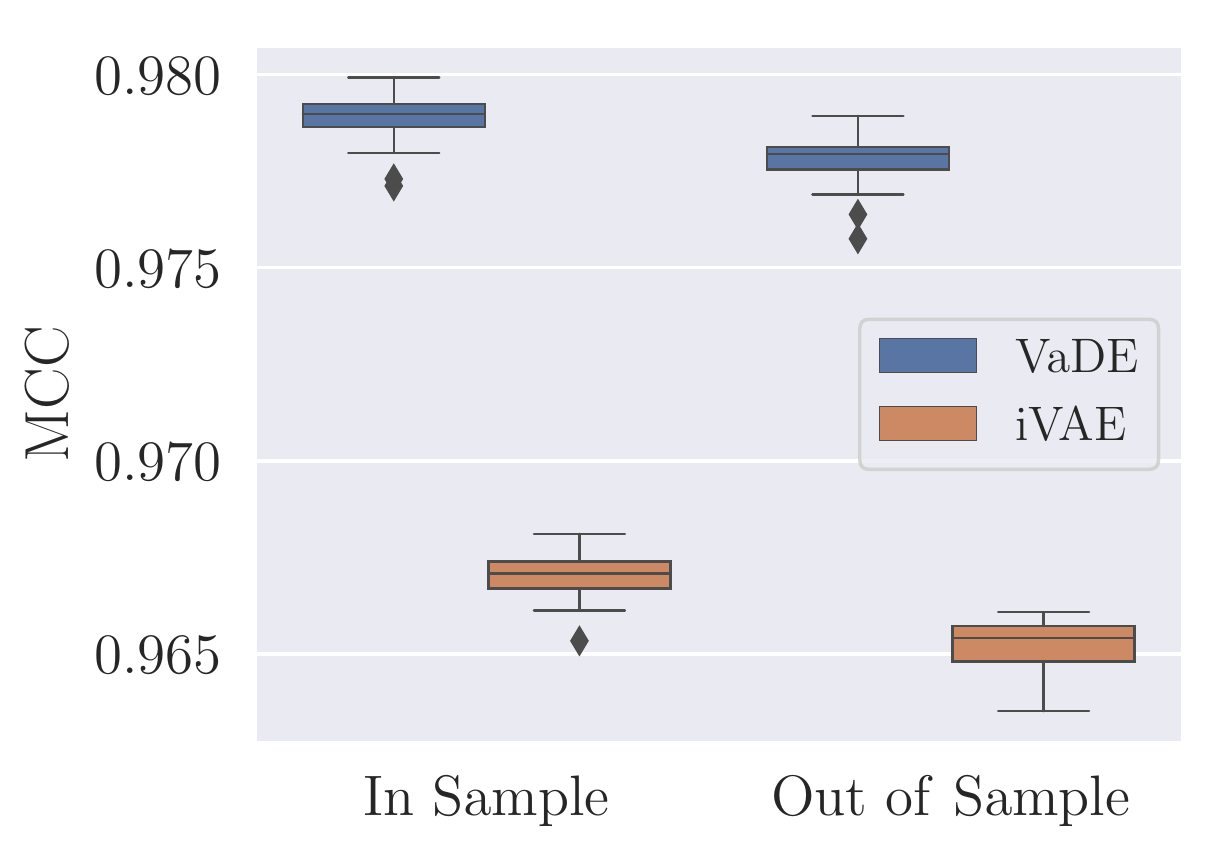}&
\includegraphics[height=3.0cm]{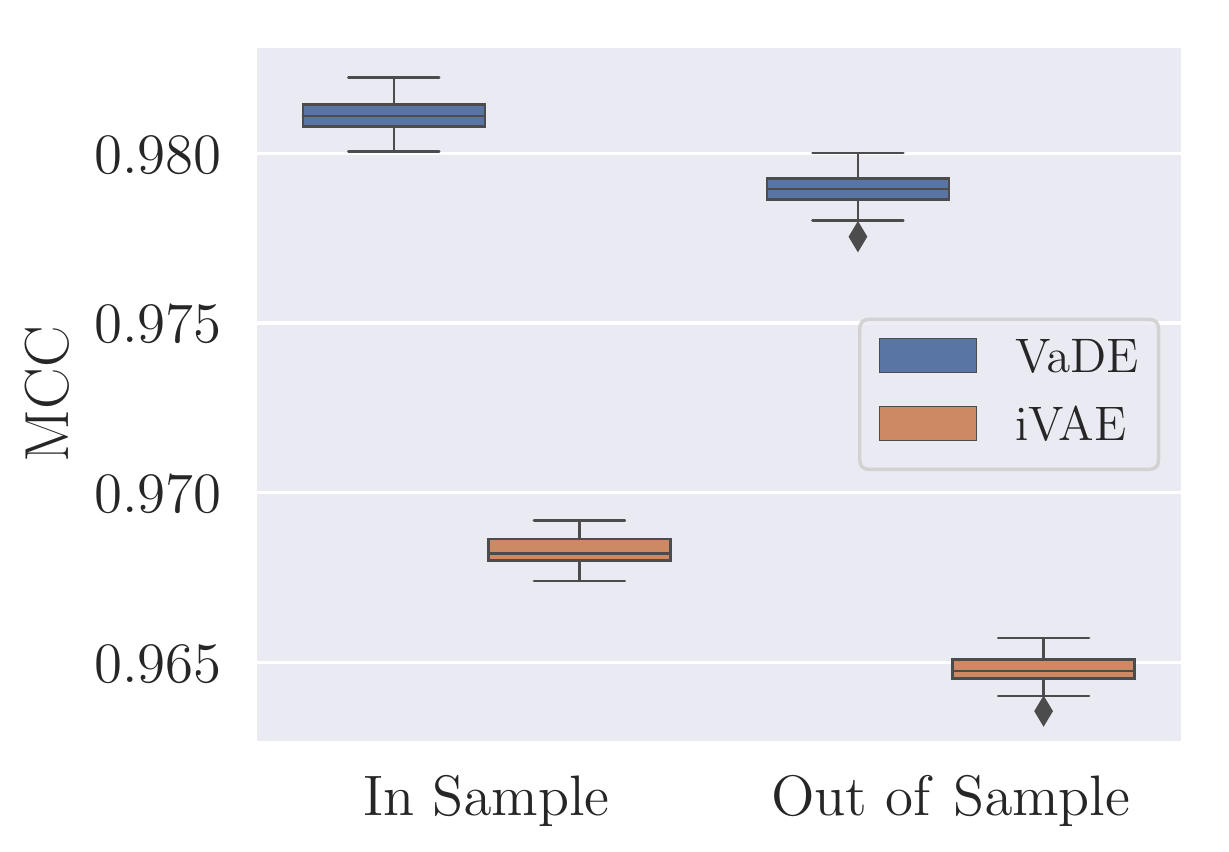}&
\includegraphics[height=3.0cm]{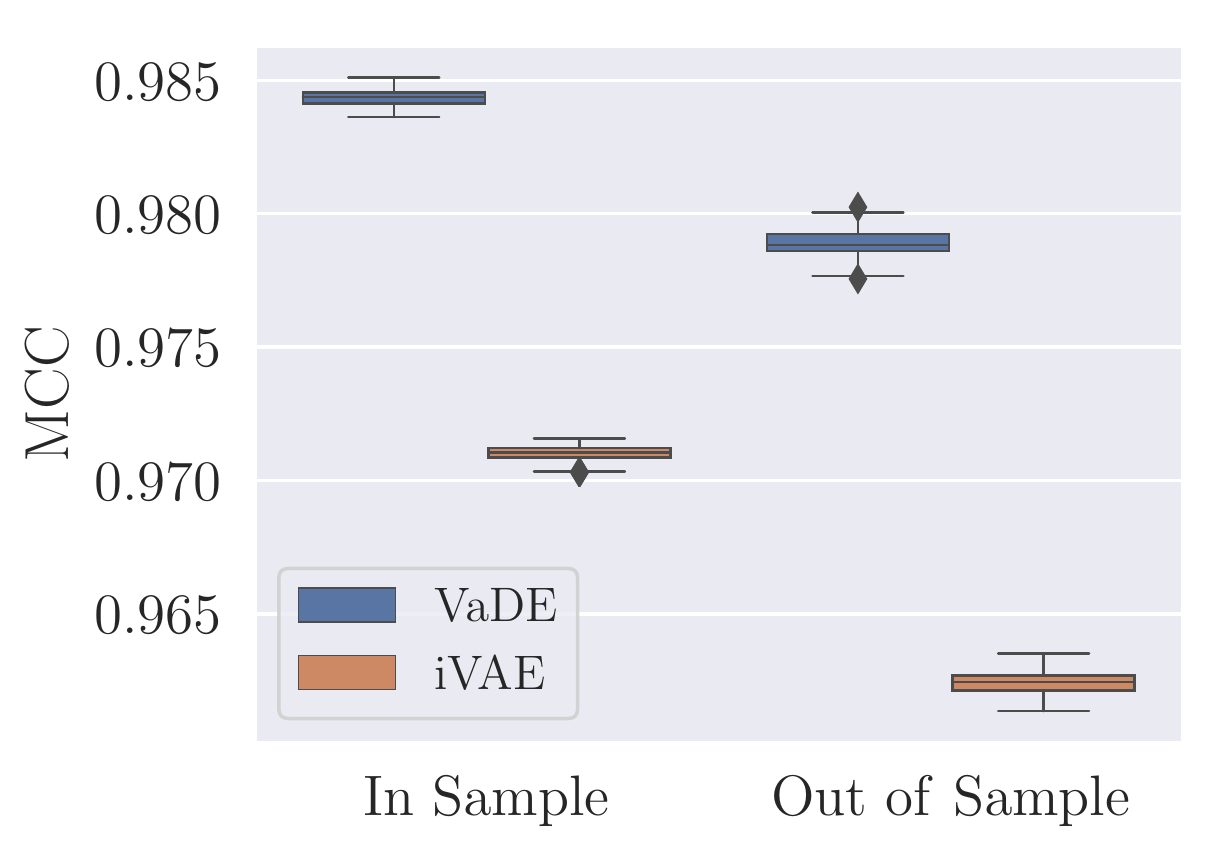}\\
\raisebox{2.8\height }{\rotatebox[origin=c]{90}{SVHN}}&
\includegraphics[height=3.0cm]{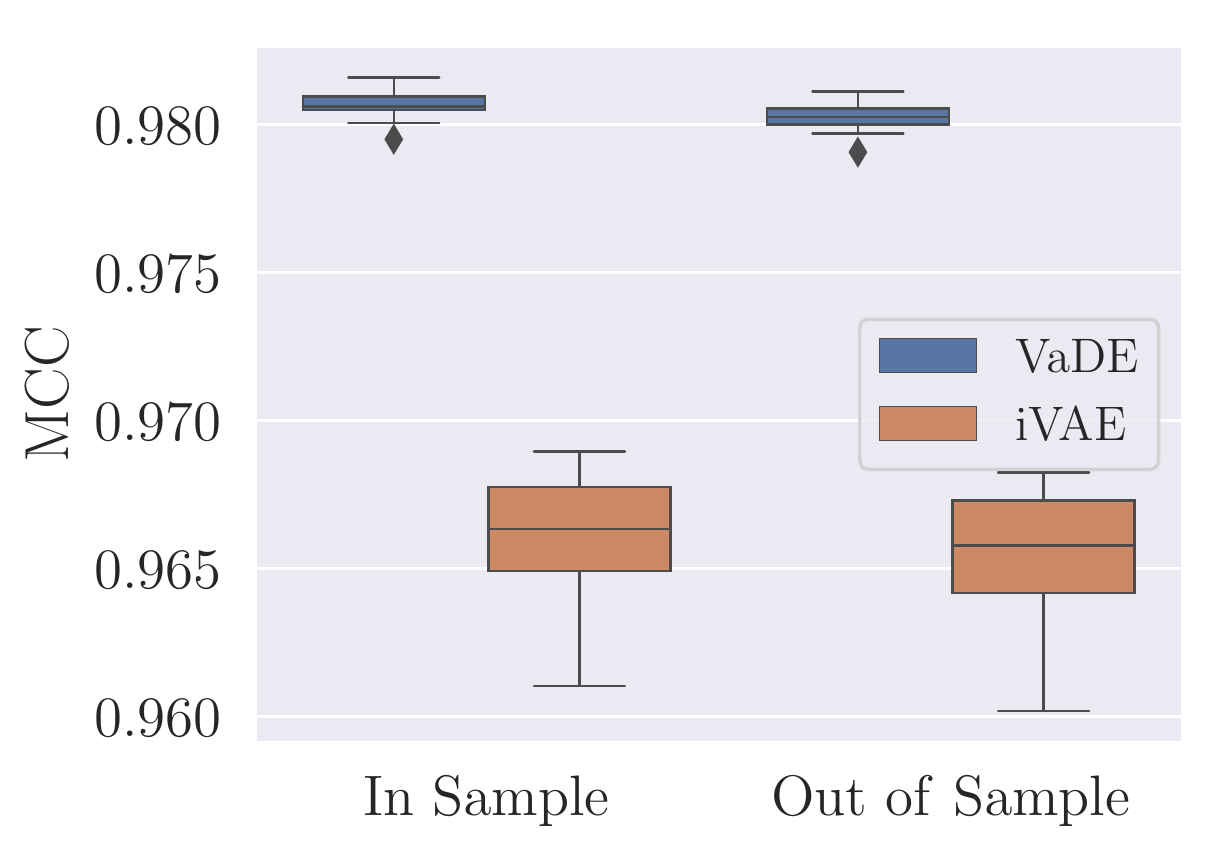}&
\includegraphics[height=3.0cm]{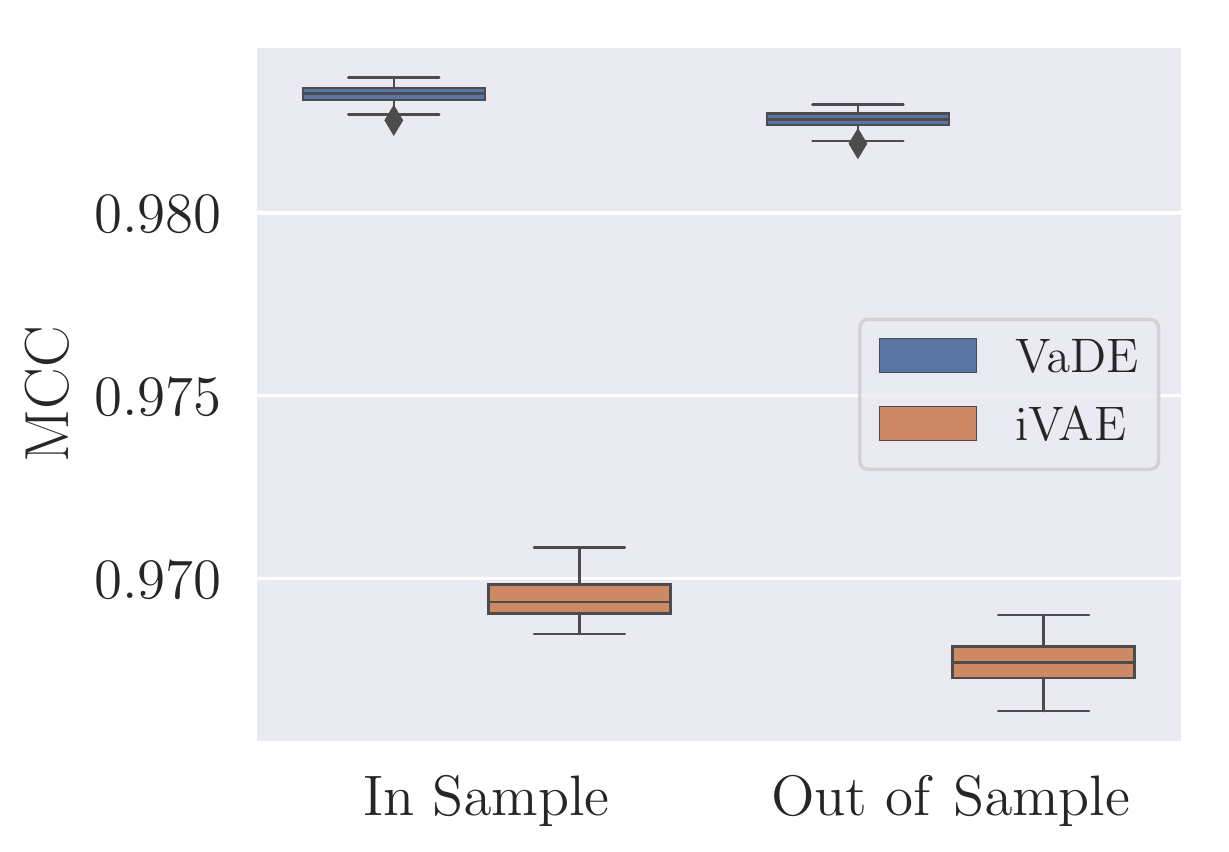}&
\includegraphics[height=3.0cm]{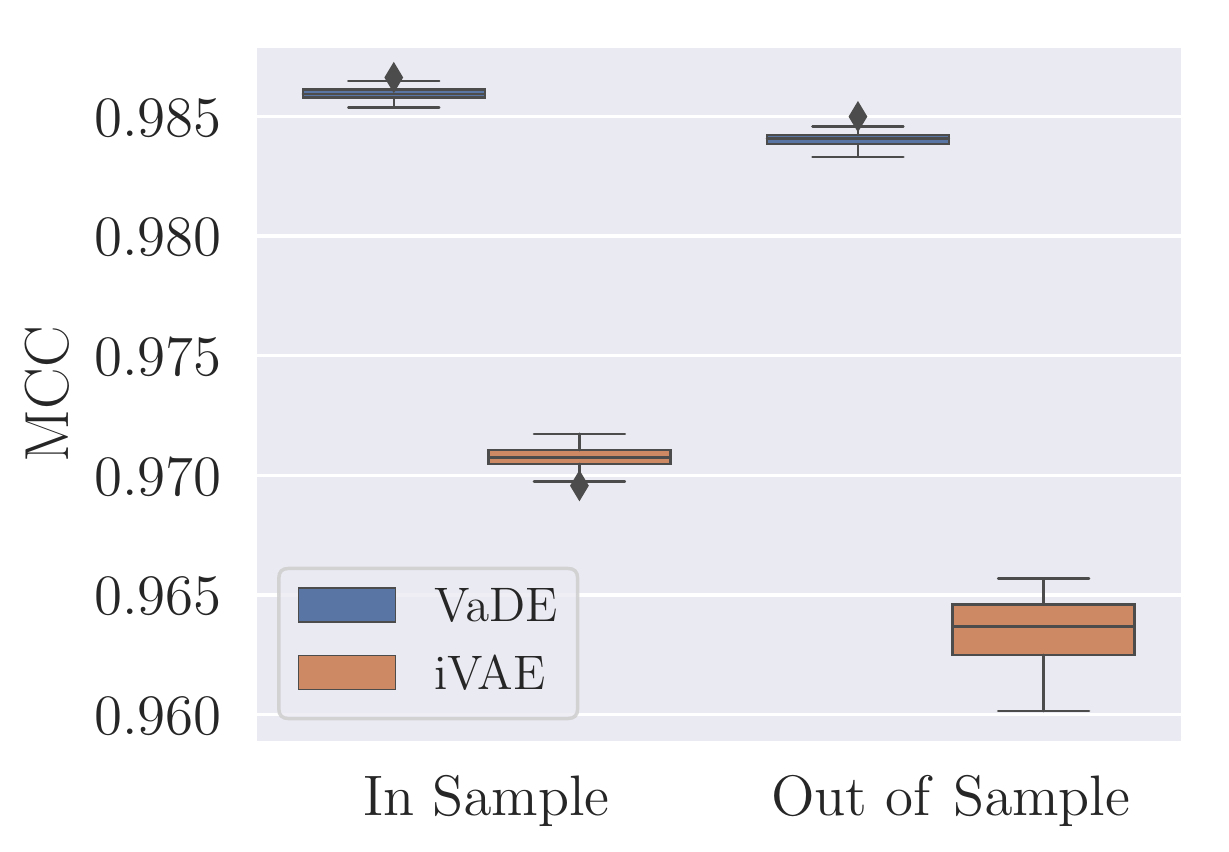}\\
\end{tabularx}
\caption{Plots showing the linear identifiability of $\v{\mu}_\phi(\v x)$ for MLP models with $d_z\in \{50,90,200\}$ as measured by MCC.
We trained each model-$d_z$-dataset combination 10 times with different seeds.
We used half of the test set to find the best linear mapping $\v A$ between between representations for each pair of seeds.
We do this alignment using Canonical Correlation Analysis (CCA)~\citep{CCA}.
We show MCC values of the aligned spaces over the test set for both `In Sample' (MCC calculated over the half of the test set $\v z$ values used to find the CCA mapping) and `Out of Sample' (MCC calculated over the remaining $\v z$ values).
VaDE, the purely unsupervised clustering approach, is in all cases scores about as well as iVAE, which is given a ground-truth $\v u$ value, or better. Note that y-axes vary in scale.
}    \label{fig:mlp_mcc}
\end{figure*}

\begin{figure*}[p]
    \centering
\begin{tabularx}{\textwidth}{cCCC}
\centering
& \hspace{2em} $d_z=50$  & \hspace{2.4em} $d_z=90$  & \hspace{2.3em} $d_z=200$  \\
\raisebox{2.4\height }{\rotatebox[origin=c]{90}{MNIST}}&
\includegraphics[height=3.0cm]{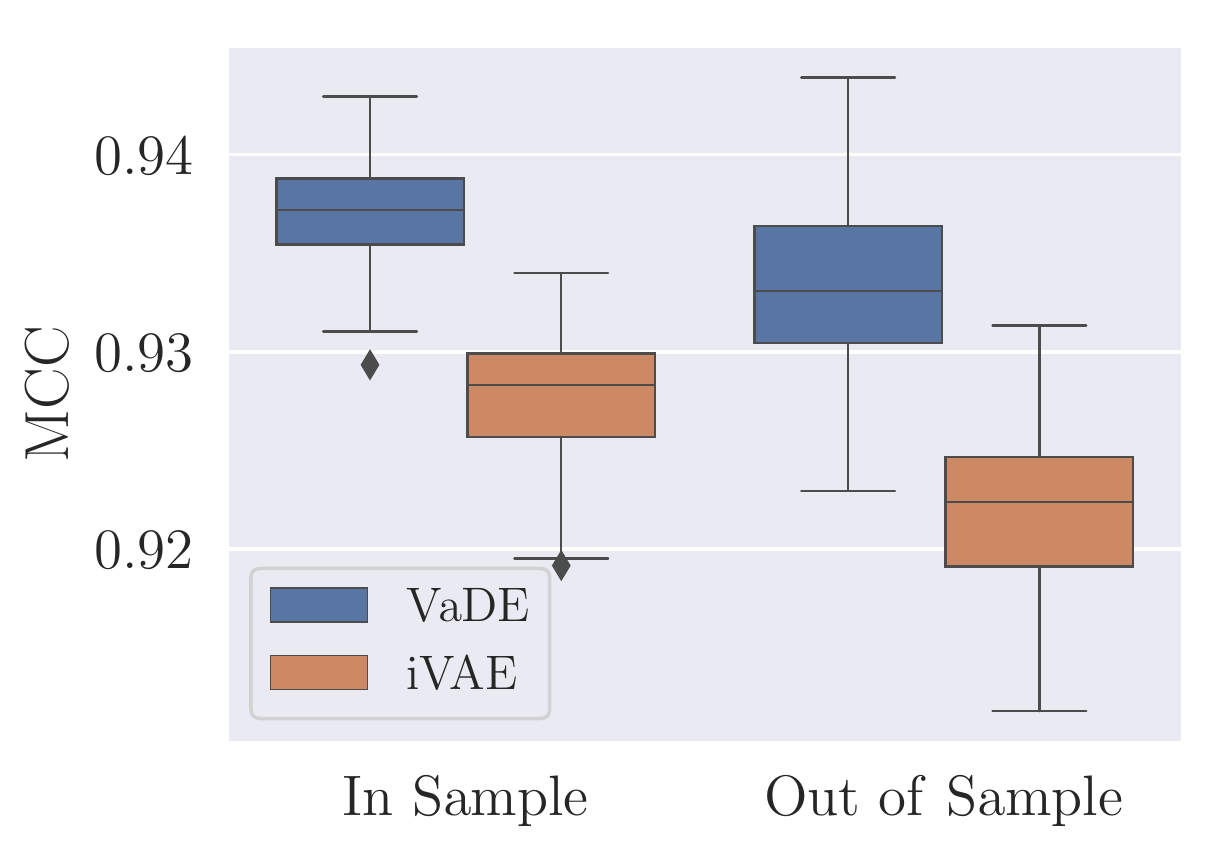}&
\includegraphics[height=3.0cm]{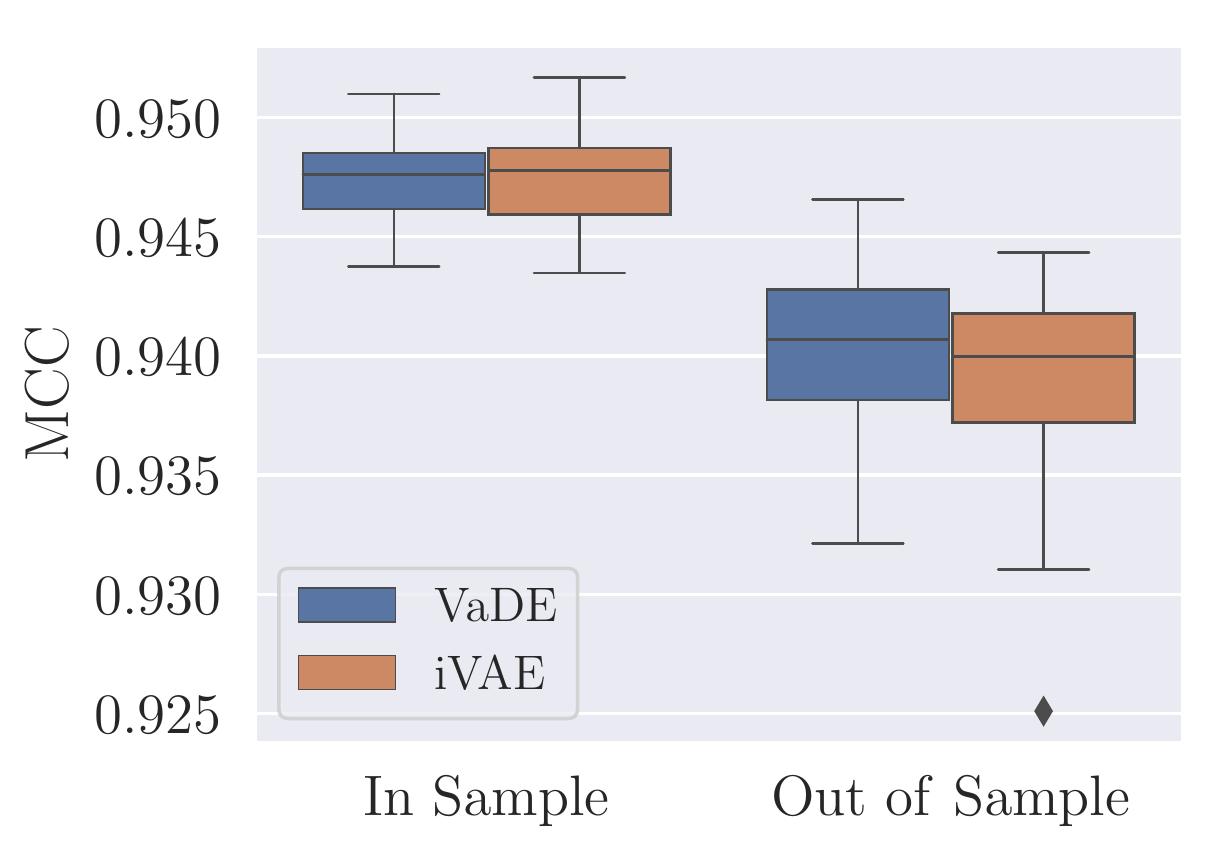}&
\includegraphics[height=3.0cm]{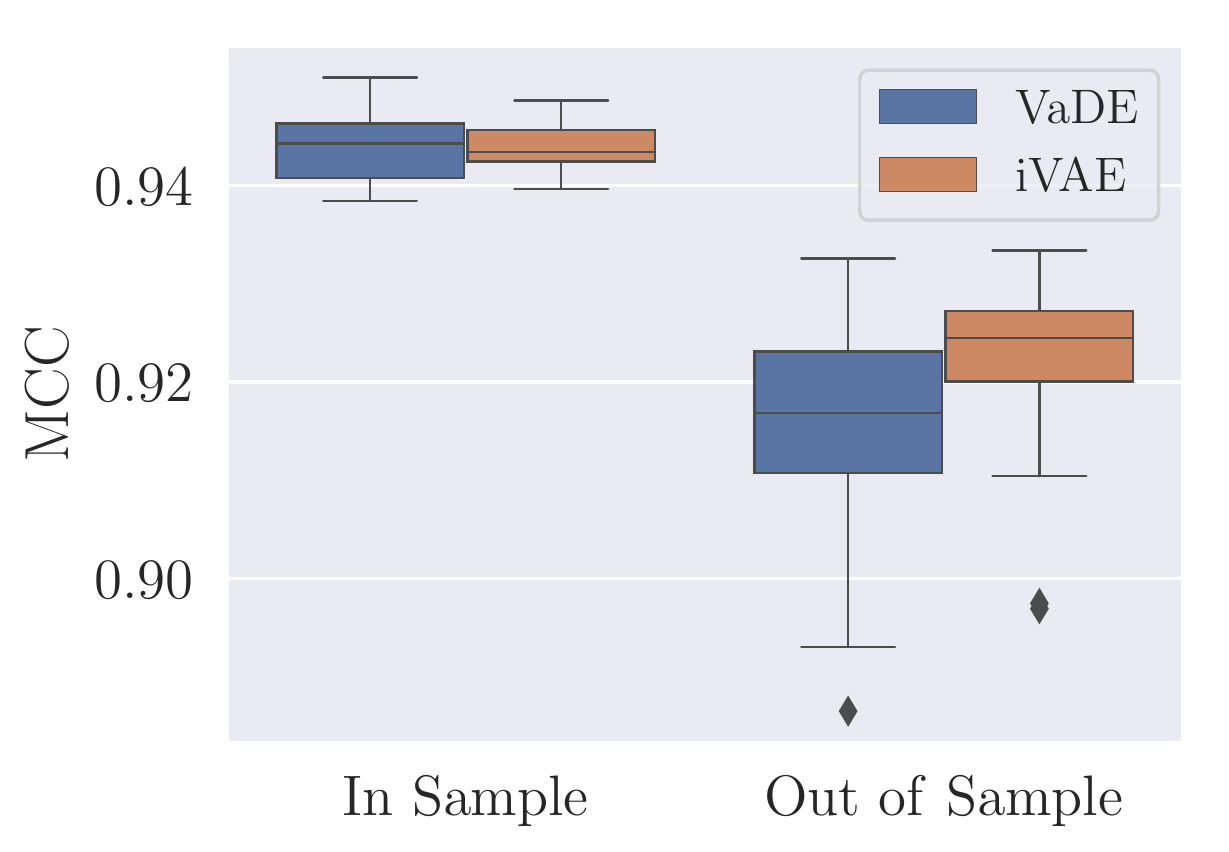}\\
\raisebox{2.1\height }{\rotatebox[origin=c]{90}{CIFAR10}}&
\includegraphics[height=3.0cm]{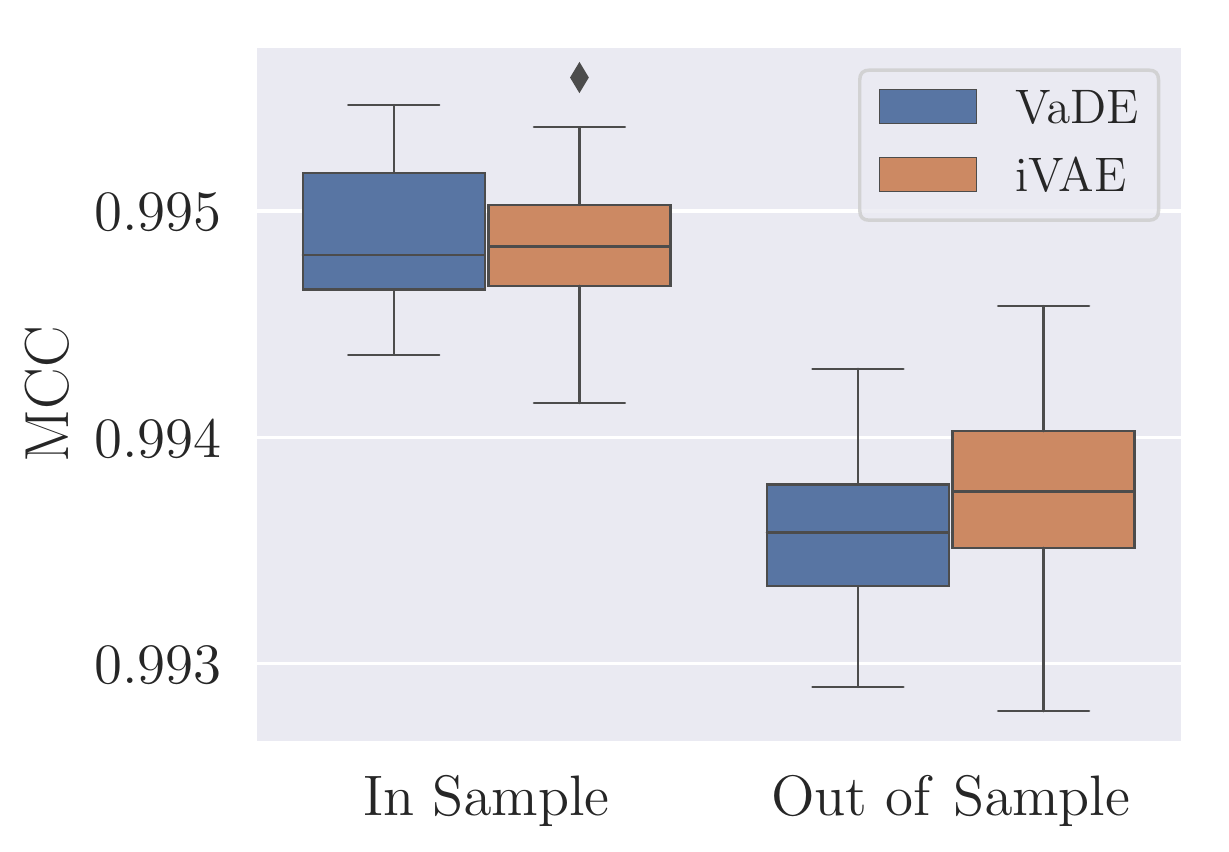}&
\includegraphics[height=3.0cm]{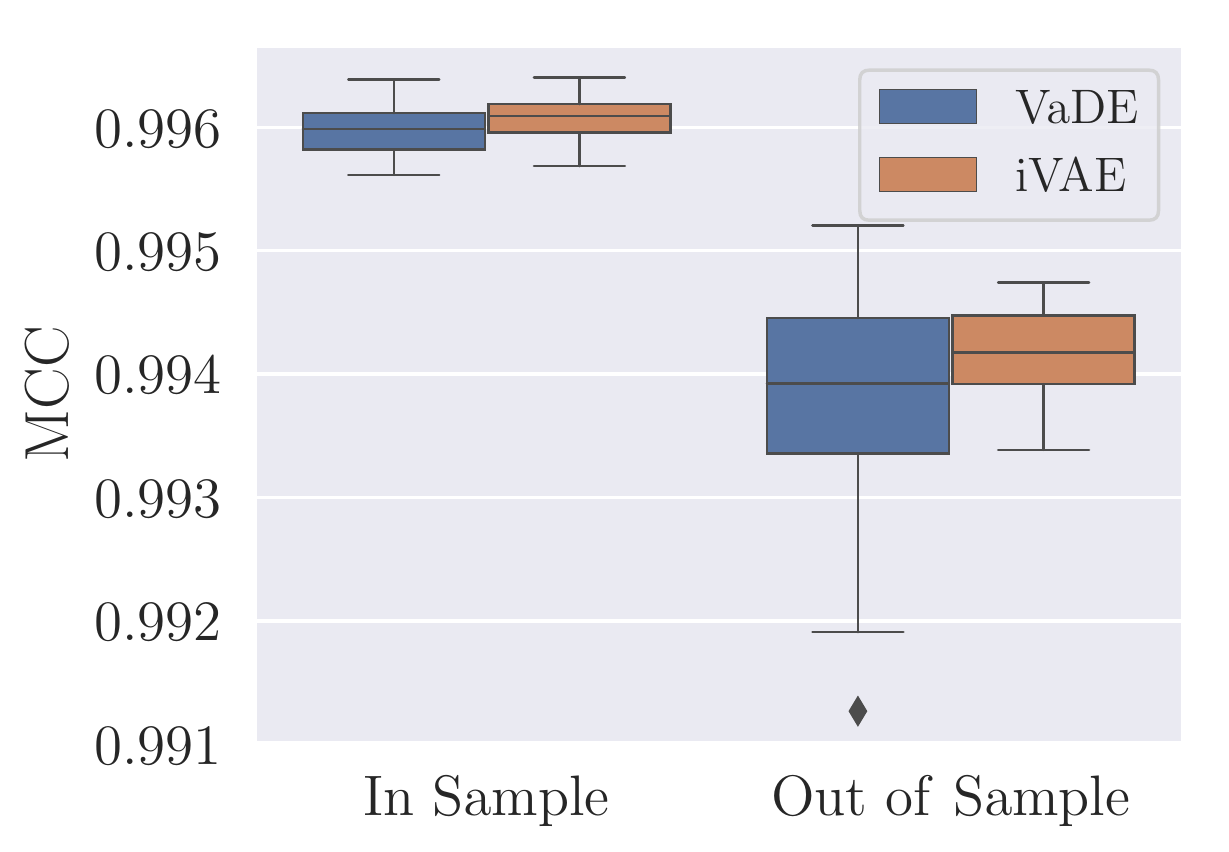}&
\includegraphics[height=3.0cm]{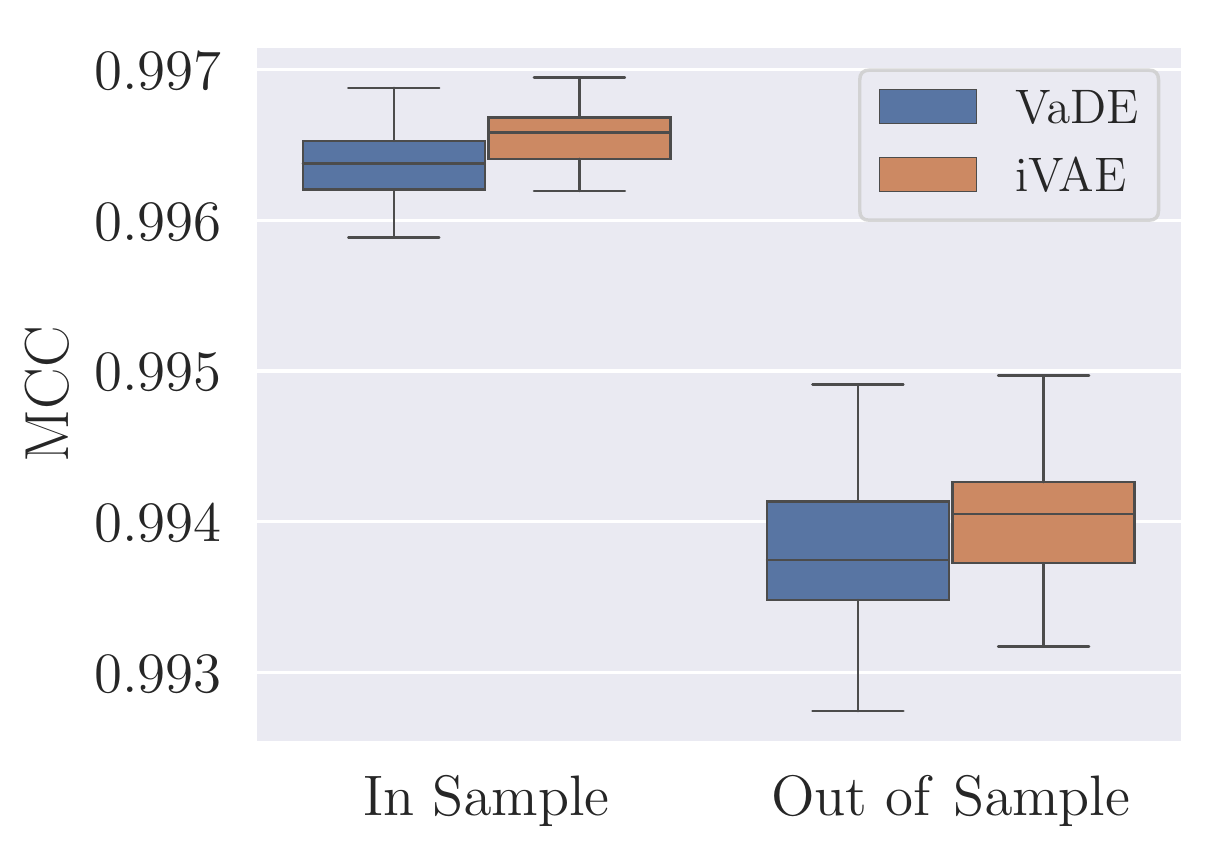}\\
\raisebox{2.8\height }{\rotatebox[origin=c]{90}{SVHN}}&
\includegraphics[height=3.0cm]{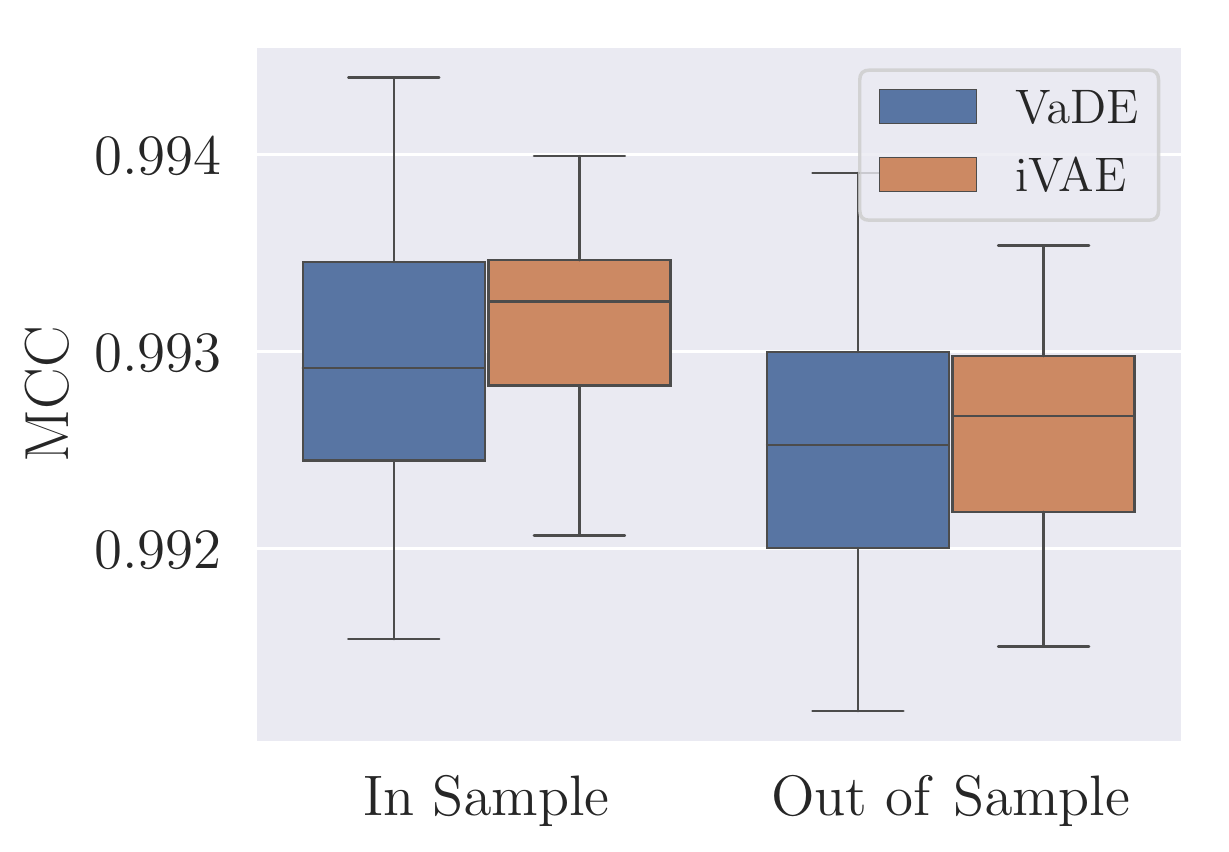}&
\includegraphics[height=3.0cm]{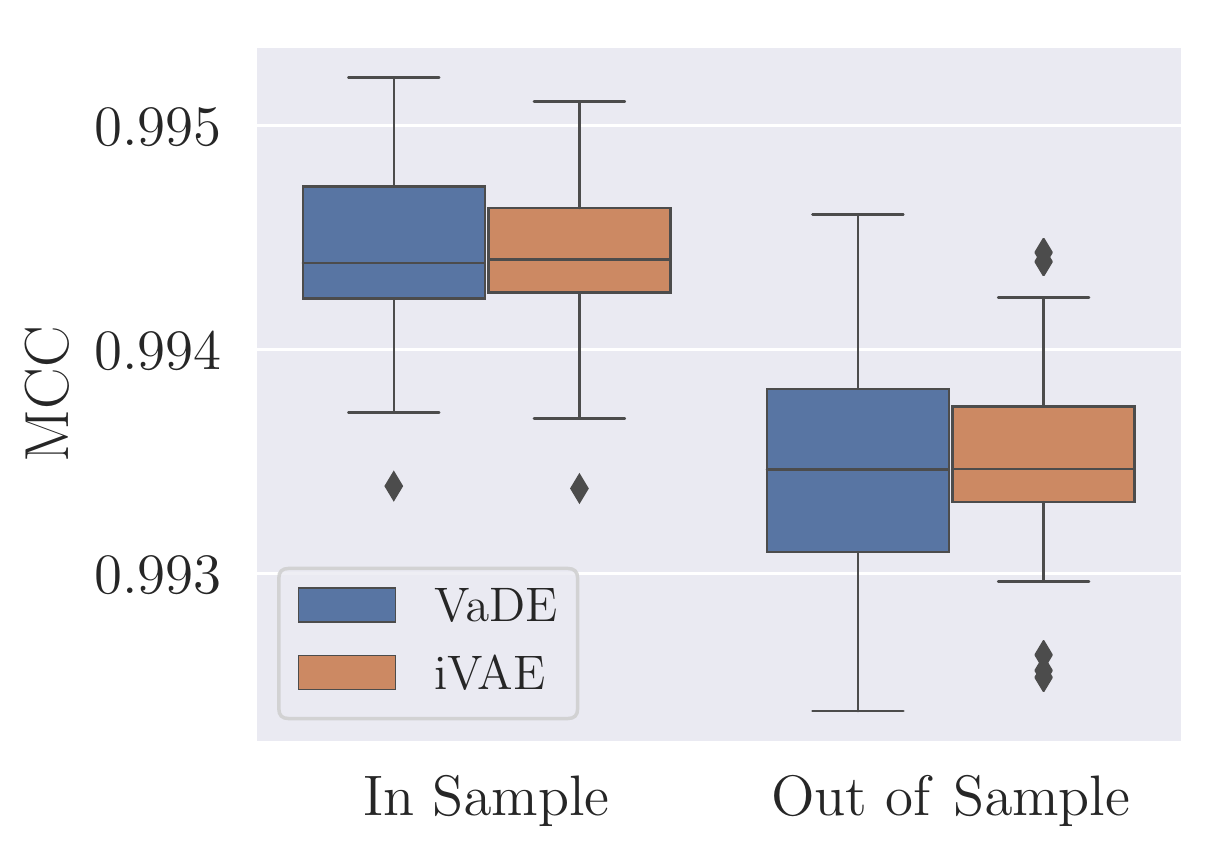}&
\includegraphics[height=3.0cm]{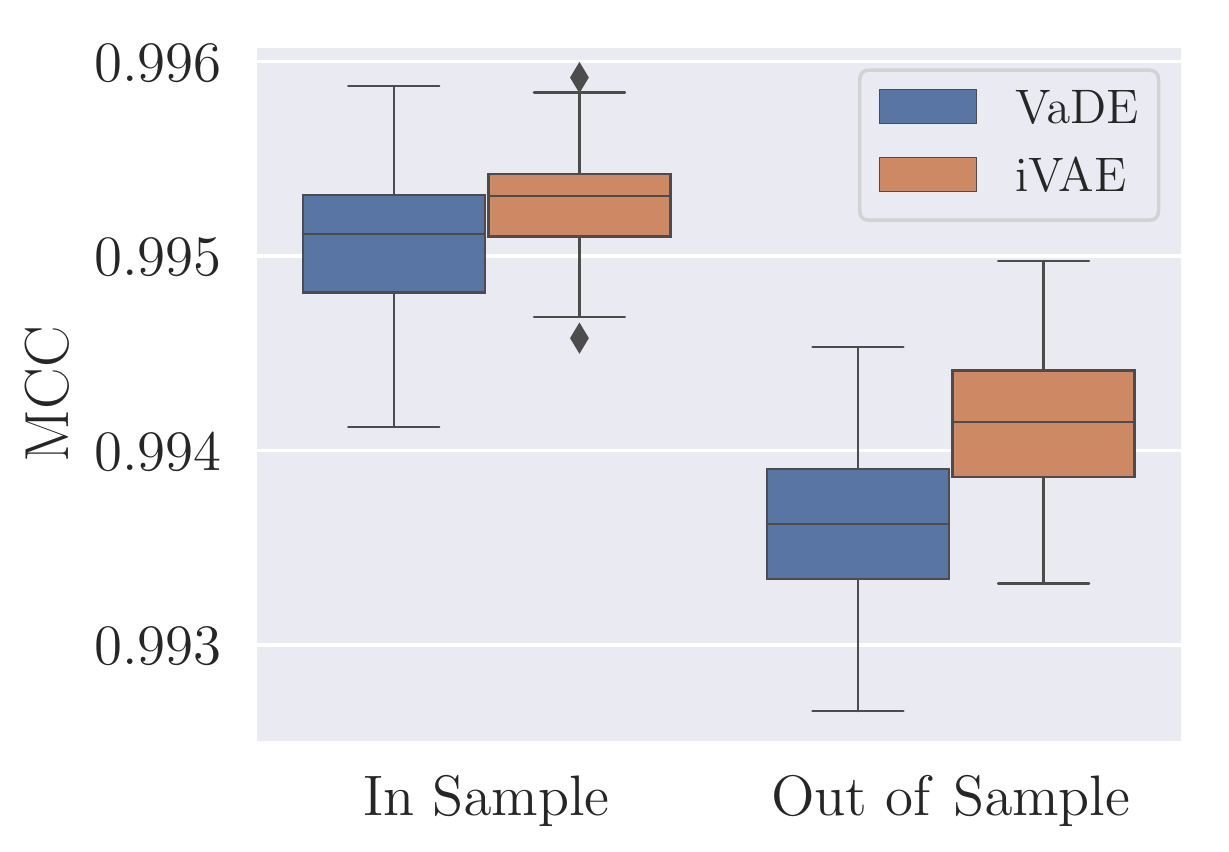}\\
\end{tabularx}

\caption{Plots showing the linear identifiability of $\v{\mu}_\phi(\v x)$ for ConvNet models with $d_z\in \{50,90,200\}$ as measured by MCC.
The overall method of analysis is the same as for MLP results.
VaDE, the purely unsupervised clustering approach, in all cases scores about as well as iVAE, which is given a ground-truth $\v u$ value.
 Note that y-axes vary in scale.
}
\label{fig:conv_mcc}
\vspace{0.5cm}
    \centering
\begin{tabularx}{\textwidth}{cCCC}
\centering
& \hspace{2em} $d_z=50$  & \hspace{2.4em} $d_z=90$  & \hspace{2.3em} $d_z=200$  \\
\raisebox{2.1\height }{\rotatebox[origin=c]{90}{CIFAR10}}&
\includegraphics[height=3.0cm]{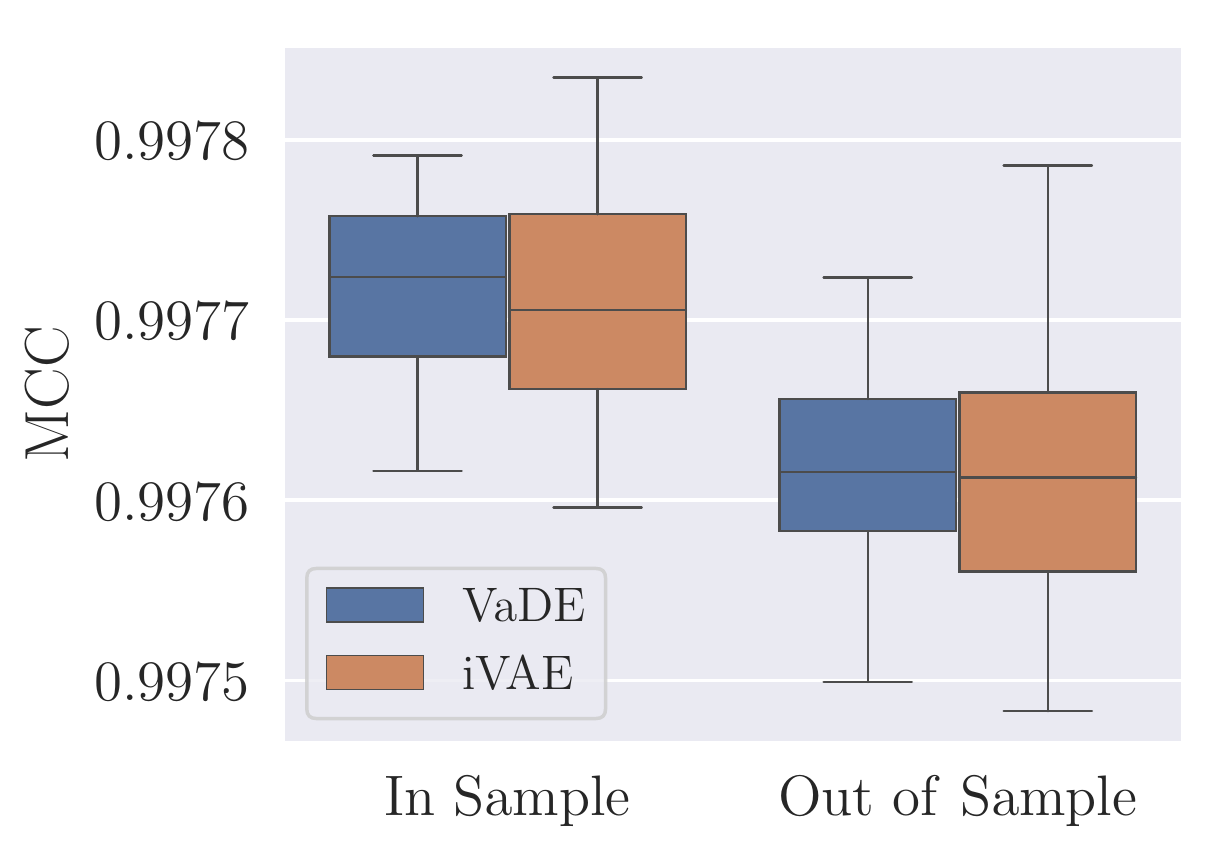}&
\includegraphics[height=3.0cm]{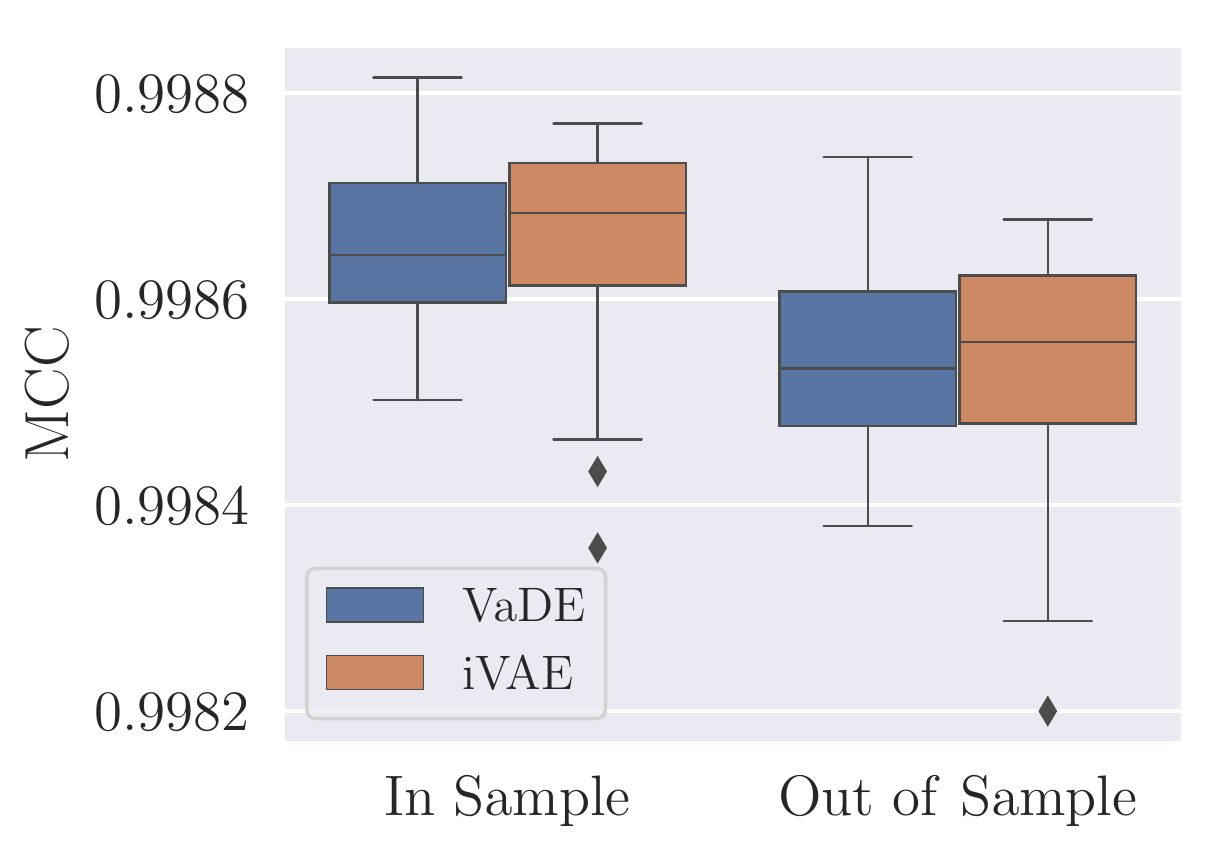}&
\includegraphics[height=3.0cm]{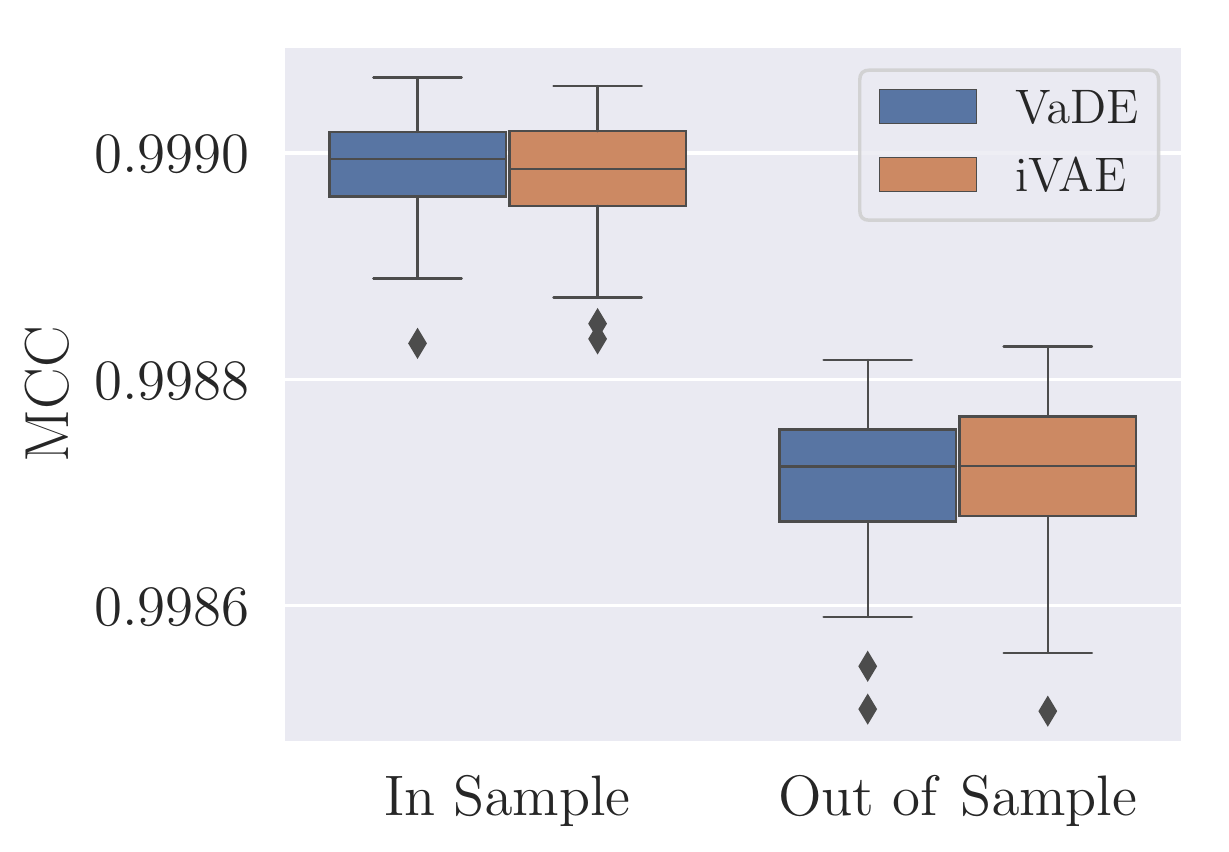}\\
\raisebox{2.8\height }{\rotatebox[origin=c]{90}{SVHN}}&
\includegraphics[height=3.0cm]{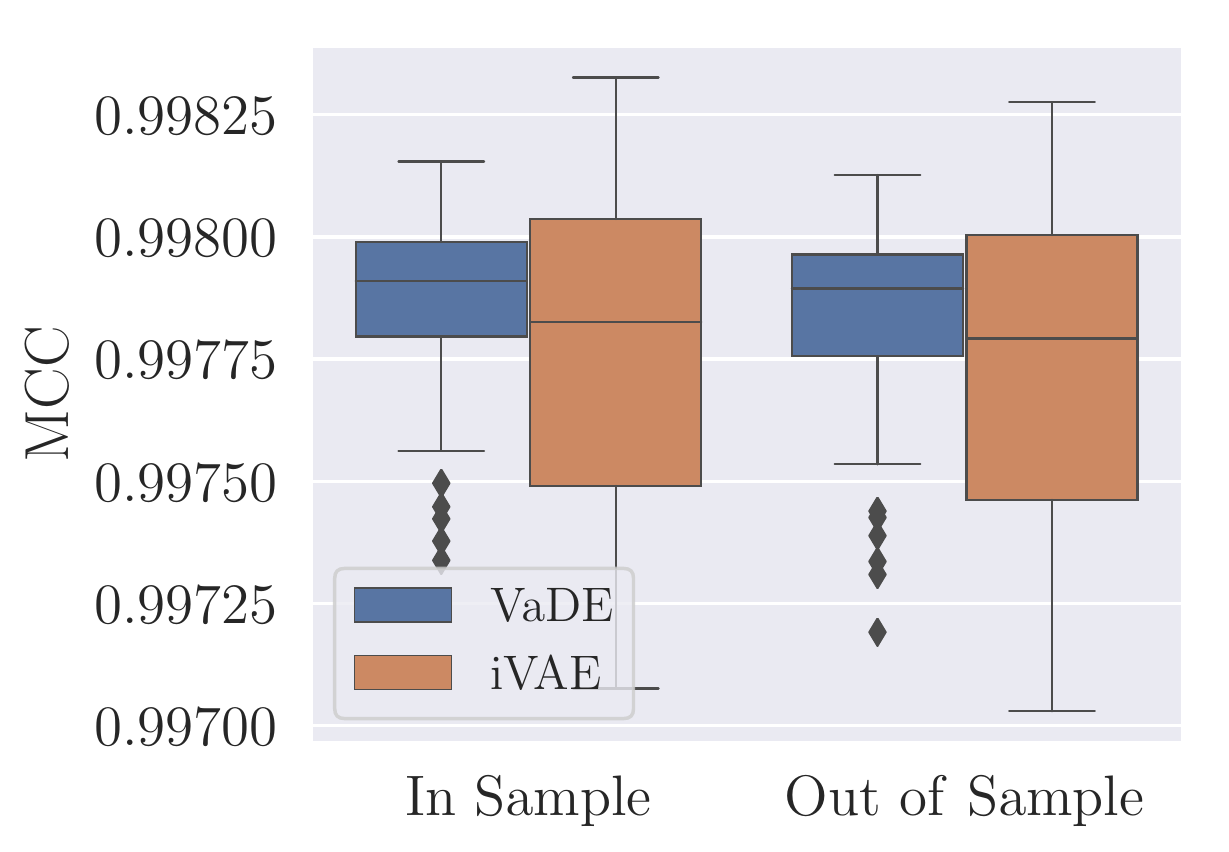}&
\includegraphics[height=3.0cm]{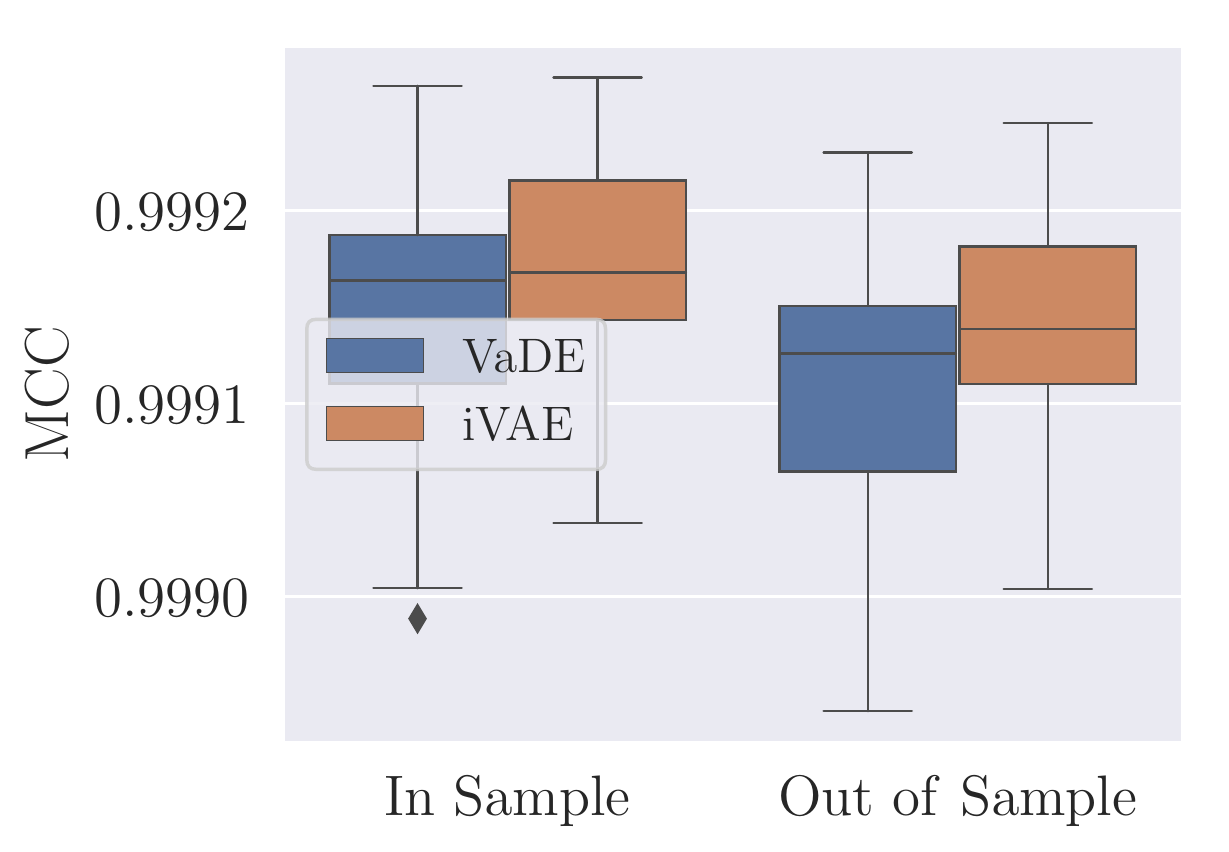}&
\includegraphics[height=3.0cm]{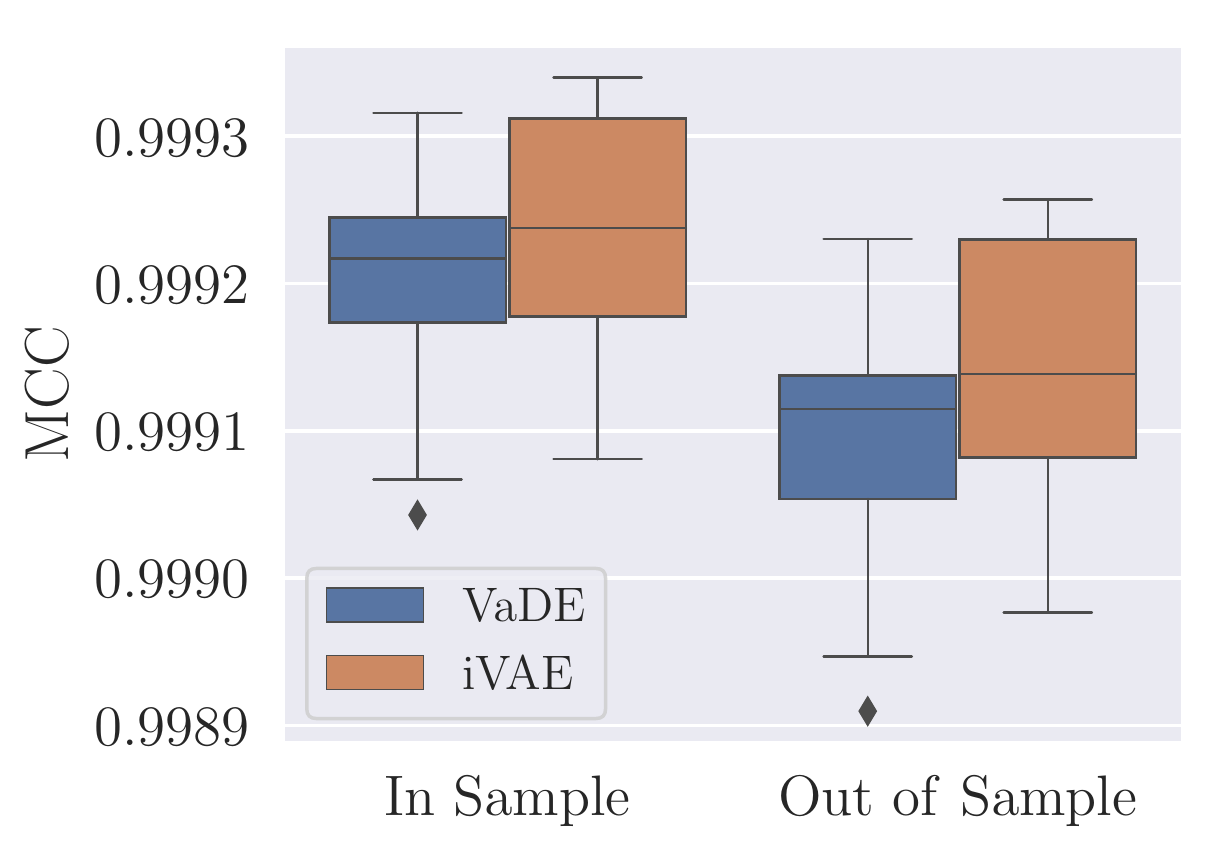}\\
\end{tabularx}

\caption{Plots showing the linear identifiability of $\v{\mu}_\phi(\v x)$ ResNet models with $d_z\in \{50,90,200\}$ as measured by MCC.
The overall method of analysis is the same as for MLP and ConvNet results.
VaDE, the purely unsupervised clustering approach, in all cases scores about as well as iVAE, which is given a ground-truth $\v u$ value.
 Note that y-axes vary in scale.
}
\label{fig:res_mcc}
  \end{figure*}

\newpage
\section{VQ-VAEs as Mixture models}

There is another class of DGMs to which we can easily extend this analysis.
Vector Quantised-VAEs (VQ-VAEs)~\citep{VQVAE} have a spatially-arrange grid of discrete latent variables (for images---for time series data a line would be used and for video a 3D block).
These latent variables index over a shared codebook of embeddings.
For each latent position the (deterministic) posterior is formed via an encoder that maps from inputs to the continuous space of the codebook vectors.
The posterior over codebooks is simply defined to be one-hot on the closest codebook entry to the encoder's embedding.
As the lookup operation is not differentiable, an exponential moving average method with close links to K-means~\citep{macqueen1967} is used to learn the codebook.
VQ-VAEs are a kind of mixture model is disguise, with the learnt codebook embeddings acting as means.

While VQ-VAEs are deterministic models, as their posteriors over codebook indexes are one-hot, they do have a probabilistic relaxation~\citep{Sonderby2017}, relaxed-VQ-VAEs (rVQ-VAEs).
Here the posterior over codebook indexes is exactly the responsibilities used in learning a GMM using variational inference~\cite[\S10.2]{Bishop2006}, albeit a mixture model where all components are hardcoded to have identity covariance.

\begin{figure}[h]
    \centering
    \includegraphics[width=0.45\textwidth]{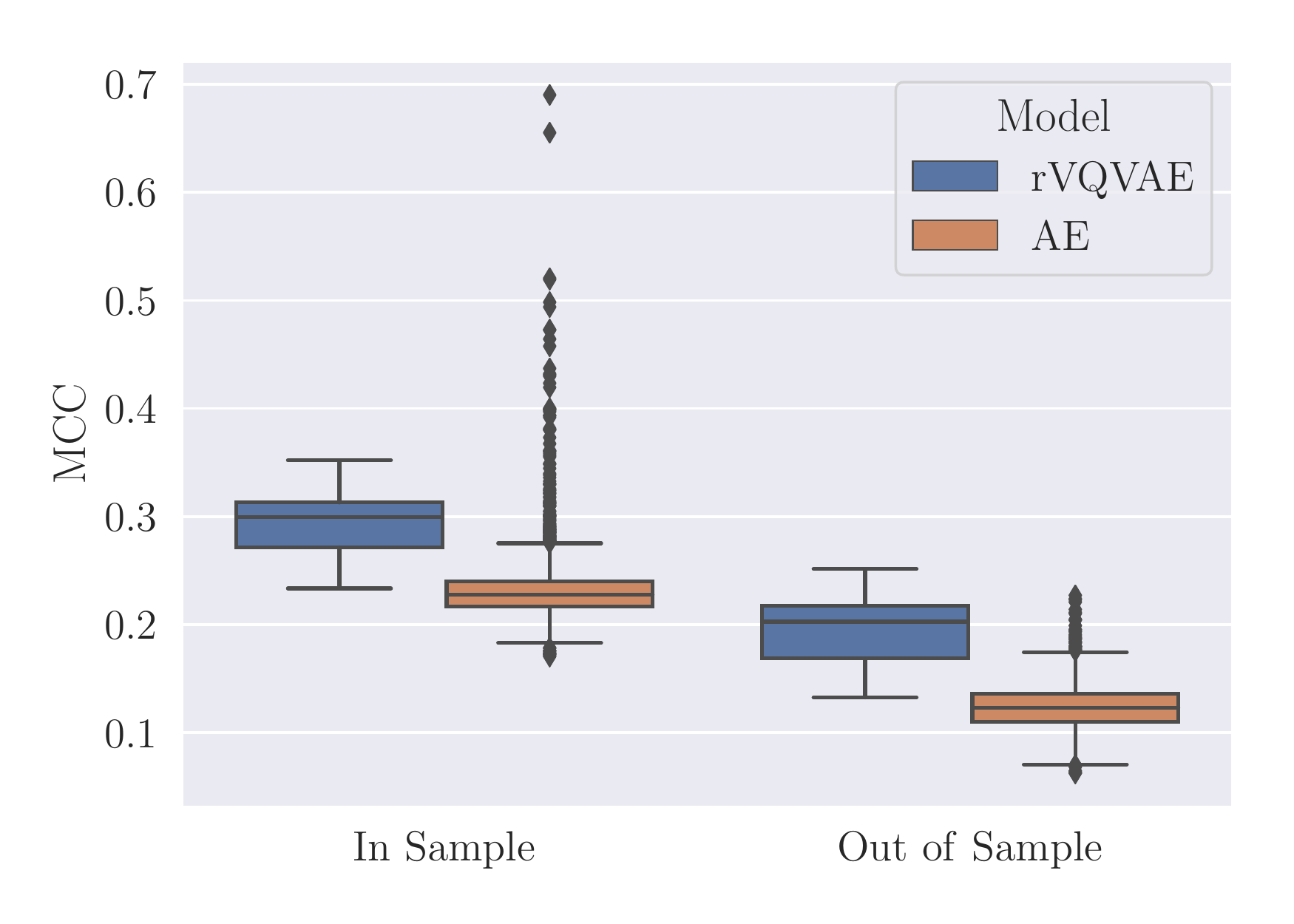}
    \caption{MCC of rVQ-VAE vs baseline AE (where no quantisation is carried out) from 10 random restarts, trained on CIFAR10. Note that while it seems there are a lot of outliers for the AE baseline, recall that we are showing boxplots over 3250 MCC values for each model.}
    \label{fig:my_label}
\end{figure}

We train relaxed-VQ-VAEs and a matching autoencoder baseline (where the continuous representations are directly given to the decoder, ie they are not quantised) 10 times each on CIFAR10 and measured the MCCs between the resulting $\v z$ representations.
Of course the representations in these models form a grid, in the standard implementation we used\footnote{\href{https://github.com/bshall/VectorQuantizedVAE}{github.com/bshall/VectorQuantizedVAE}} this is an $8\times8$ grid, each entry in $\mathbb{R}^{d_z}$, so when calculating the MCC values we compare each latent grid position against its matching counterparts from other seeds.
We then plot all these values, $8\times8$ latent positions $\times55$ pairs of seeds from 10 restarts giving $3250$ MCC values for each of rVQ-VAE and the baseline AE.

The rVQ-VAE is moderately higher in MCC than the AE baseline, but neither scores particularly high.
So while explicitly having a mixture model in the latent representations does seem to increase MCC, it does not lead to the large values, often $> 0.9$, seen for the iVAE and VaDE models we show in the main paper.
This may be related to the fact that these rVQ-VAE and AE-baseline models have convolutional latents, so they are not parameterised by MLPs.
Recall that the identifiability of iVAEs is, formally, only guaranteed when one has the dimensionality of the latent space sufficiently small.
When one has a grid of spatially-arranged latents, as in these models, the effective latent dimensionality is much larger than for the models in the main paper.
This might be part of the reason for the small MCC values.
For a discussion of MLP vs convolutional (ie spatially-arranged) latents in VAEs, see~\citep{Willetts2020a}.

\newpage
\section{Rademacher-Hashed Posteriors for \texorpdfstring{$\v u$}{u}}
\label{app:rade}
In VaDE we learn our $\v u$ representations by learning to cluster our representations in $\mathcal{Z}$, learning both the $\v z$ and $\v u$ variables jointly.
The posterior for $\v u$ is defined to be Bayes-optimal.
However, one could define the posterior for $\v u$ in many different ways---using VaDE is a design choice we make, choosing it for both its naturalness and simplicity.
In this section, we explore a method of defining $q_\phi(\v u | \v x)$ using sketching/random projections/Johnson–Lindenstrauss (JL) transforms~\citep{Johnson1984,Achlioptas2003,Dasgupta2003,Woodruff2014} combined with simple hash.

We choose the number of cluster components, $K$, indexed by $\v u$ to be $2^N$.
We sample and then \emph{fix} a simple random projection $\v A\in\mathbb{R}^{N\times d_x}$ constructed from Rademacher random variables~\citep{Achlioptas2003} that projects from $\mathcal{X}=\mathbb{R}^{d_x}$ to $\mathbb{R}^N$:
\begin{align}
A_{i,j} = \begin{cases}
      +1, & \text{with probability } \frac{1}{2}  \\
      -1, & \text{with probability } \frac{1}{2}.
    \end{cases}
\label{eq:jlbin1}
\end{align}
Let $\v h(\v x) = \v A \v x$, our JL embedding in $\mathbb{R}^N$.
From this we can then define a binary representation $\v b$ of length $N$:
\begin{equation}
    b_i(\v x) = 0.5\times\left(\mathrm{sign}(h_i(\v x)) + 1\right).
\end{equation}
$\v u$ is then the one-hot representation of the binary vector $\v b$:
\begin{equation}
    \v{u}^{\mathrm{Rad}}(\v{x}) = \mathrm{onehot}\left(\sum_{i=0}^{N-1} 2^i \times b_i(\v x)\right).
\end{equation}
The `posterior' for $\v u$ is then simply this one-hot value: $q^{\mathrm{Rad}}(\v u | \v x) = \delta(\v u - \v u^{\mathrm{Rad}}(\v x))$.
While this construction is not differentiable, that is not important as we do not learn $\v A$ so no gradient propagation is needed.

This approach can be interpreted as constructing a Johnson-Lindenstrauss embedding out of $\v x$ of dimension $N$ where which orthant a data point falls into in this JL-space is then used to index over mixture components in $\mathcal{Z}$, the latent space of our learnt DGM.

\begin{figure*}[h!]
\centering
    \includegraphics[width=0.4\textwidth]{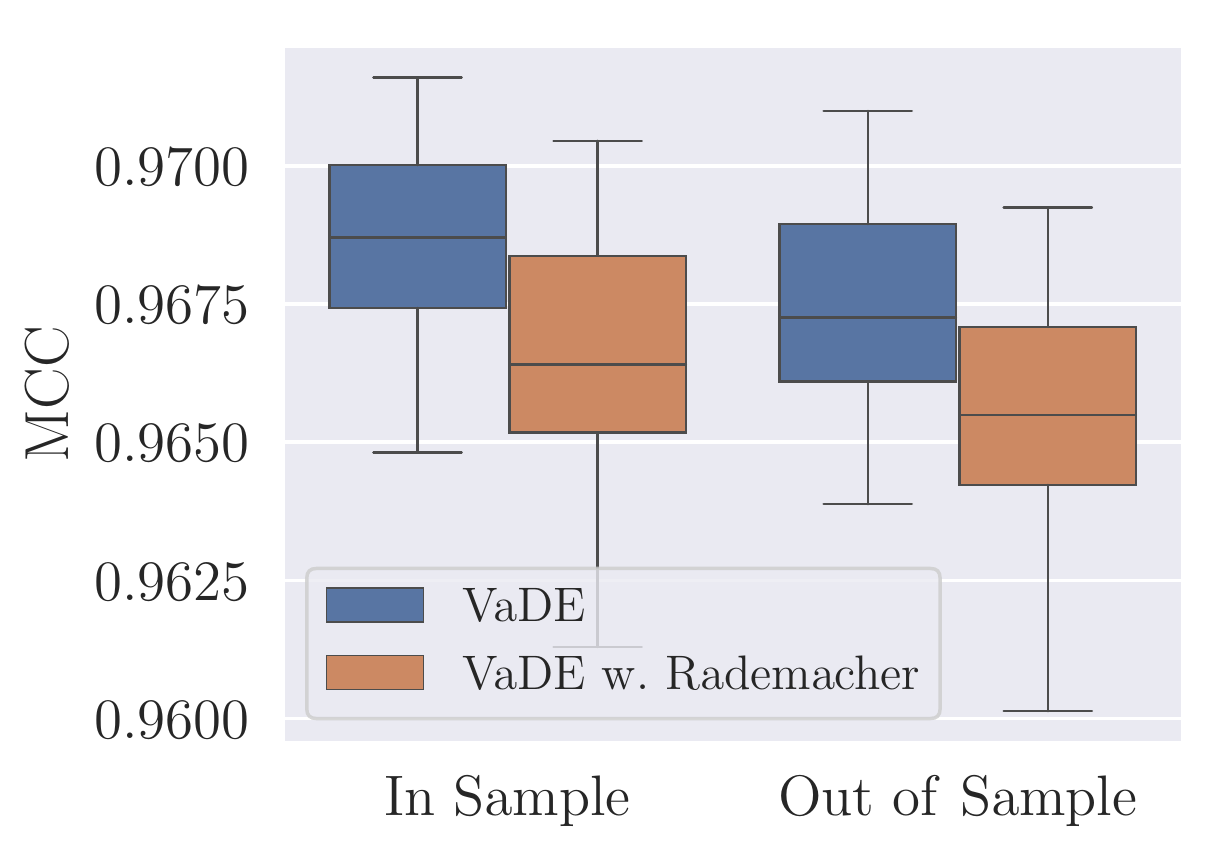}
\caption{MCC for Rademacher-$\v u$ and VaDE for $d_z=50$, 128-component mixtures, MLP encoder and decoder, trained on CIFAR10.}
    \label{fig:rademacher}
\end{figure*}

We train MLP mixture-model VAEs with $128$ mixture components, so $N=7$, on CIFAR10 with $d_z=50$ using this JL-based construction, so $q_\phi(\v u | \v x)=q^{\mathrm{Rad}}(\v u | \v x)$, and a matching VaDE model as baseline, training each 10 times and measuring the MCC between pairs of runs as in the main paper.
Each Rademacher-constructed run has differently-inited $\v A$ projections---they are inited using the same looped-over seed as all other neural network parameters in the model (though $\v A$ receives no updates, it is fixed).

We find that this method does lead to respectable MCC values, but it does not outperform directly learning to cluster in $\mathcal{Z}$, the approach that VaDE takes.
We expect this is because the $\v u$ in VaDE is an emergent property of the model, whereas this Rademacher-$\v u$ is an imposed and potentially unnatural partitioning of the data.

This constructive method for defining a $\v u$-task takes this method closer to the methods discussed in \S\ref{sec:back} and \S\ref{sec:relatedwork} based around learning a classifier on some defined $\v u$-task---and much like in the contrastive-learning methods for self-supervised learning, here we are defining a $\v u$ variable from our data in an algorithmic way and then learning representations that connect to that choice of task.
Recently JL methods have been used in non-Linear ICA, but in a different context: acting on the output of a flow to provide the means of a variational posterior~\citep{Camuto2020a}.

We are more interested in the naturally-emerging task of learning to cluster in $\mathcal{Z}$, as done in VaDE, but perhaps these more algorithmic approaches to obtain $\v u$ in an unsupervised setting could prove fruitful.

\section{GMM-Flows}
We also perform some experiments using flows~\citep{deepflows}.
We simply introduce a Gaussian Mixture Model $p_\theta(\v z)=\sum_{\v u \in \mathcal{U}}p_\theta(\v z | \v u)p_\theta(\v u)$ as the flow's base distribution and train under maximum likelihood,
\begin{equation}
    \log p_\theta(\v x) = \log p_\theta(\v z) \left\lvert\det \frac{\partial f_\theta^{-1}}{\partial \v x}\right\rvert
\end{equation}
where $\v x = f_\theta(\v z)$ is the bijective mapping between $\mathcal{Z}$ and $\mathcal{X}$.

We train an autoregressive flow of 10 layers on the same synthetic data as in the main paper, with batch size of 256, with a mixture model as the base distribution with 50 components.
We find that these flows generally perform as well or better than iVAE or VaDE at discovering latent structure.
Note that the iVAE and VaDE results are the same as in the main paper, Figure~\ref{fig:l4_mcc}.
\begin{figure*}[h]
    \centering
    \includegraphics[width=0.55\textwidth]{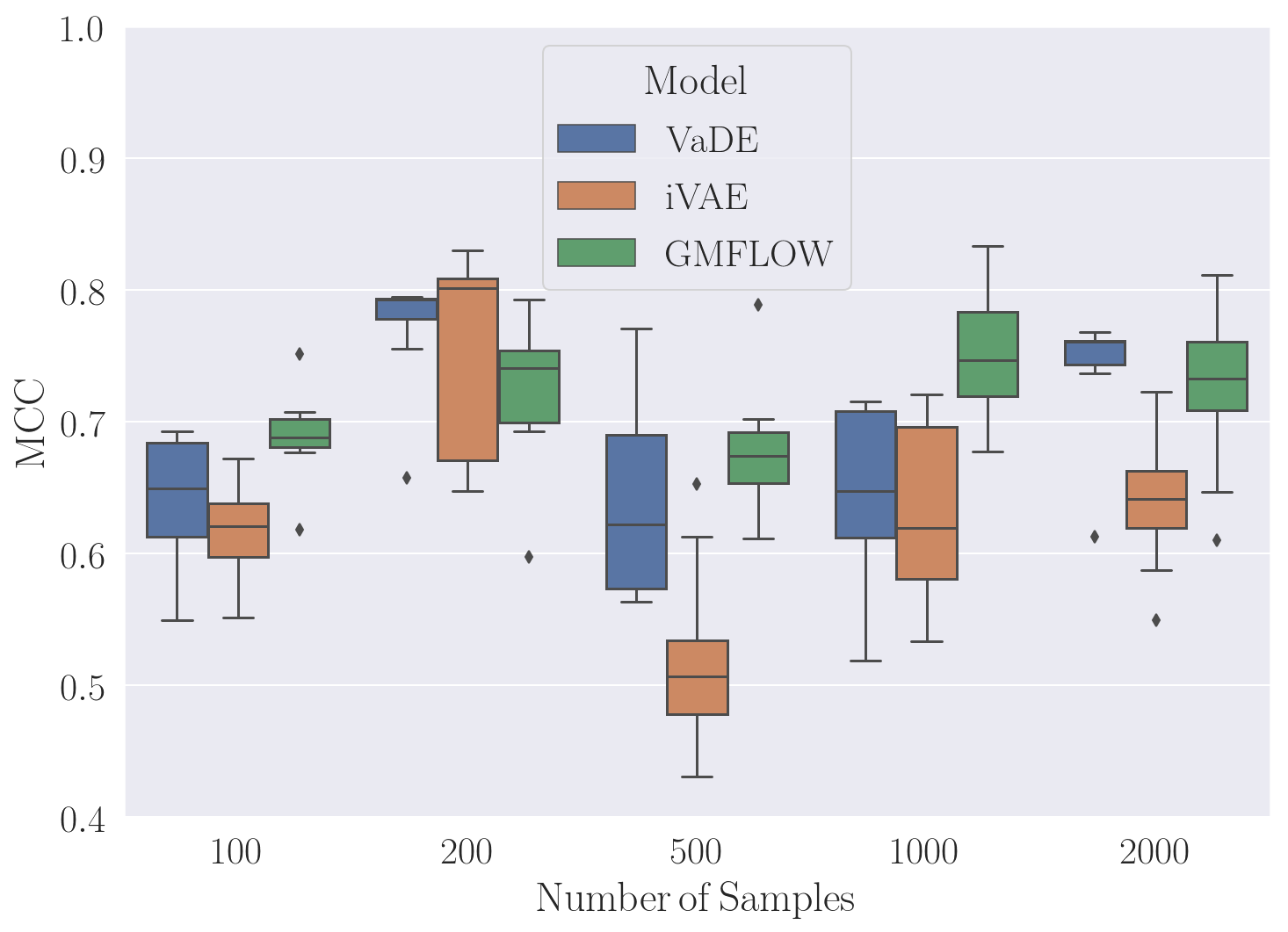}
\caption{MCC for VaDE, iVAE and GMM-Flow for synthetic data generated using $L=4$ mixing.}
    \label{fig:l4_mcc_with_flow}
    \vspace{-1cm}
\end{figure*}
\section{Compute Resources}
We used Azure Virtual Machines for the experiments on standard image datasets, NV series machines using M60 GPUs.
As we said in the main paper, each individual run, training over image data, takes about 8h on average, though of course MLPs take less time, only a few hours, and ResNet runs take longer, about a day.
Multiple runs can fit on a GPU at once.

The synthetic runs are much more lightweight---on a Dell XPS13 running Ubuntu these experiments run, over all datasets with their different numbers of datapoints, in about 6h for each VAE-derived model class, our GMM-flows taking about a day.

\section{Enforcing requirements for identifiability}
Identifiability in iVAEs is obtained when the matrix $\v L$, Eq~\eqref{eq:l_matrix}, made out of the natural parameters of our conditional priors $\{p_\theta(\v z | \v u)\}$, is invertible.
Optimising the ELBO of our clustering models while ensuring that this is the case means our problem is one of constrained optimisation.
We can thus use an approach reminiscent of projected gradient descent~\citep{beck2017first}.
We simple track the invertibility of $\v L$ during optimisation and if it becomes uninvertible we can add a small amount of noise to the natural parameters of the conditional priors.

Alternatively, one could view this as a soft constraint, and so add an additional term to the objective that captures the invertibility of $\v L$. The Condition Number, $\mathrm{CN}(\cdot)$, would be one way to do this: $\Tilde{\ELBO}=\ELBO - \alpha\mathrm{CN}(\v L)$.

However, we found empirically that this process was not invoked---the constraint on the mixture model's parameters is sufficiently easy to satisfy that, when simply optimising $\ELBO^{\mathrm{VaDE}}$, $\v L$ remains invertible.

Recall that in order to construct $\v L$ there have to be $(k\times d_z)+1$ different values of $\v u$ and for Gaussians $k=2$.
When using standard image datasets, the ground-truth class labels provide a natural $\v u$-task.
For many datasets (SVHN, CIFAR10, MNIST, etc), there are 10 classes.
This means that under the terms of this identifiability theory, an iVAE with Gaussian $p_\theta(\v z | \v u)$ distributions trained on these datasets can have at most $d_z=4$.
Yet already the ideas underpinning iVAEs have been extended to flows, where $d_z = d_x$ by construction, and have been found to work~\citep{gin}.
Empirically, it thus seems that the requirement of $\v L$ invertibility is a sufficient but not necessary condition.
To that end, in our empirical investigations we do not constrain the latent spaces of our models to be so small, and we find that iVAEs, and our clustering approach, produce identifiable models even with $d_z=200$---for which $\v L$ is not even possible to construct.

\newpage
\section{Statistical Tests}
\label{app:stats}
\subsection{iVAE--VaDE}
\begin{table}[h!]
\caption{MCC values, taken over all dimensions in each model, used in Wilcoxon statistical test, giving $p=0.422$.}
\centering
\begin{tabular}{ccc>{\centering}p{2.5cm}>{\centering}p{2.5cm}>{\centering\arraybackslash}p{5cm}}
\bf{ARCHITECTURE}  & \bf{DATASET}  & $d_z$ & \bf{iVAE MCC} & \bf{VaDE MCC} & \bf{iVAE MCC}$-$\bf{VaDE MCC}\\
\hline
\\
MLP    & MNIST    & 50    & 0.6270                       & 0.6360                       & -0.0090                                 \\
MLP    & MNIST    & 90    & 0.6000                       & 0.5828                       & 0.0172                                  \\
MLP    & MNIST    & 200   & 0.5830                       & 0.4899                       & 0.0931                                  \\
MLP    & CIFAR 10 & 50    & 0.7131                       & 0.7128                       & 0.0002                                  \\
MLP    & CIFAR 10 & 90    & 0.5584                       & 0.5703                       & -0.0118                                 \\
MLP    & CIFAR 10 & 200   & 0.3618                       & 0.5192                       & -0.1575                                 \\
MLP    & SVHN     & 50    & 0.7065                       & 0.7389                       & -0.0324                                 \\
MLP    & SVHN     & 90    & 0.5752                       & 0.6221                       & -0.0470                                 \\
MLP    & SVHN     & 200   & 0.4009                       & 0.5489                       & -0.1480                                 \\
ConvNet   & MNIST    & 50    & 0.6754                       & 0.6281                       & 0.0473                                  \\
ConvNet   & MNIST    & 90    & 0.7075                       & 0.6722                       & 0.0354                                  \\
ConvNet   & MNIST    & 200   & 0.8390                       & 0.7208                       & 0.1182                                  \\
ConvNet   & CIFAR 10 & 50    & 0.7301                       & 0.7165                       & 0.0136                                  \\
ConvNet   & CIFAR 10 & 90    & 0.5736                       & 0.5469                       & 0.0267                                  \\
ConvNet   & CIFAR 10 & 200   & 0.4351                       & 0.4656                       & -0.0305                                 \\
ConvNet   & SVHN     & 50    & 0.6731                       & 0.6604                       & 0.0127                                  \\
ConvNet   & SVHN     & 90    & 0.5634                       & 0.5425                       & 0.0209                                  \\
ConvNet   & SVHN     & 200   & 0.4467                       & 0.4384                       & 0.0083                                  \\
ResNet & CIFAR 10 & 50    & 0.8297                       & 0.8116                       & 0.0181                                  \\
ResNet & CIFAR 10 & 90    & 0.7158                       & 0.7014                       & 0.0144                                  \\
ResNet & CIFAR 10 & 200   & 0.5455                       & 0.5775                       & -0.0320                                 \\
ResNet & SVHN     & 50    & 0.7443                       & 0.7284                       & 0.0159                                  \\
ResNet & SVHN     & 90    & 0.6354                       & 0.6129                       & 0.0225                                  \\
ResNet & SVHN     & 200   & 0.5090                       & 0.4922                       & 0.0168                                 
\end{tabular}
\end{table}

\newpage
\subsection{iVAE--VAE}
\begin{table}[h!]
\caption{MCC values, taken over all dimensions in each model, used in Wilcoxon statistical test, giving $p=0.029$.}
\centering
\begin{tabular}{ccc>{\centering}p{2.5cm}>{\centering}p{2.5cm}>{\centering\arraybackslash}p{5cm}}
\bf{ARCHITECTURE}  & \bf{DATASET}  & $d_z$ & \bf{iVAE MCC} & \bf{VAE MCC} & \bf{iVAE MCC}$-$\bf{VAE MCC}\\
\hline
\\
MLP    & MNIST    & 50    & 0.6270                       & 0.6179                       & 0.0091                                 \\
MLP    & MNIST    & 90    & 0.6000                       & 0.5676                       & 0.0324                                  \\
MLP    & MNIST    & 200   & 0.5830                       & 0.5614                       & 0.0216                                  \\
MLP    & CIFAR 10 & 50    & 0.7131                       & 0.6904                       & 0.0226                                  \\
MLP    & CIFAR 10 & 90    & 0.5584                       & 0.5851                       & -0.0267                                 \\
MLP    & CIFAR 10 & 200   & 0.3618                       & 0.5047                       & -0.1429                                 \\
MLP    & SVHN     & 50    & 0.7065                       & 0.7034                       & 0.0031                                 \\
MLP    & SVHN     & 90    & 0.5752                       & 0.5737                       & 0.0014                                 \\
MLP    & SVHN     & 200   & 0.4009                       & 0.4838                       & -0.0829                                 \\
ConvNet   & MNIST    & 50    & 0.6754                       & 0.6098                       & 0.0656                                  \\
ConvNet   & MNIST    & 90    & 0.7075                       & 0.5326                       & 0.1749                                  \\
ConvNet   & MNIST    & 200   & 0.8390                       & 0.5958                       & 0.2432                                  \\
ConvNet   & CIFAR 10 & 50    & 0.7301                       & 0.7005                       & 0.0296                                  \\
ConvNet   & CIFAR 10 & 90    & 0.5736                       & 0.5581                       & 0.0155                                  \\
ConvNet   & CIFAR 10 & 200   & 0.4351                       & 0.4196                       & 0.0155                                 \\
ConvNet   & SVHN     & 50    & 0.6731                       & 0.6546                       & 0.0185                                  \\
ConvNet   & SVHN     & 90    & 0.5634                       & 0.5479                       & 0.0155                                  \\
ConvNet   & SVHN     & 200   & 0.4467                       & 0.4720                       & -0.0253                                  \\
ResNet & CIFAR 10 & 50    & 0.8297                       & 0.8273                       & 0.0024                                  \\
ResNet & CIFAR 10 & 90    & 0.7158                       & 0.6956                       & 0.0202                                  \\
ResNet & CIFAR 10 & 200   & 0.5455                       & 0.5127                       & 0.0328                                 \\
ResNet & SVHN     & 50    & 0.7443                       & 0.7431                       & 0.0012                                  \\
ResNet & SVHN     & 90    & 0.6354                       & 0.6237                       & 0.0118                                  \\
ResNet & SVHN     & 200   & 0.5090                       & 0.4866                       & 0.0223                                 
\end{tabular}
\end{table}
\newpage

\subsection{VAE--VaDE}
\begin{table}[h!]
\caption{MCC values, taken over all dimensions in each model, used in Wilcoxon statistical test, giving $p=0.039$.}
\centering
\begin{tabular}{ccc>{\centering}p{2.5cm}>{\centering}p{2.5cm}>{\centering\arraybackslash}p{5cm}}
\bf{ARCHITECTURE}  & \bf{DATASET}  & $d_z$ & \bf{VAE MCC} & \bf{VaDE MCC} & \bf{VAE MCC}$-$\bf{VaDE MCC}\\
\hline
\\
MLP    & MNIST    & 50    & 0.6179                       & 0.6360                       & -0.0181                                 \\
MLP    & MNIST    & 90    & 0.5676                       & 0.5828                       & -0.0152                                  \\
MLP    & MNIST    & 200   & 0.5614                       & 0.4899                       & 0.0715                                  \\
MLP    & CIFAR 10 & 50    & 0.6904                       & 0.7128                       & -0.0224                                  \\
MLP    & CIFAR 10 & 90    & 0.5851                       & 0.5703                       & 0.0148                                 \\
MLP    & CIFAR 10 & 200   & 0.5047                       & 0.5192                       & -0.0145                                 \\
MLP    & SVHN     & 50    & 0.7034                       & 0.7389                       & -0.0355                                 \\
MLP    & SVHN     & 90    & 0.5737                       & 0.6221                       & -0.0484                                 \\
MLP    & SVHN     & 200   & 0.4838                       & 0.5489                       & -0.0651                                 \\
ConvNet   & MNIST    & 50    & 0.6098                       & 0.6281                       & -0.018                                  \\
ConvNet   & MNIST    & 90    & 0.5326                       & 0.6722                       & -0.1396                                  \\
ConvNet   & MNIST    & 200   & 0.5958                       & 0.7208                       & -0.1250                                \\
ConvNet   & CIFAR 10 & 50    & 0.7005                       & 0.7165                       & -0.0160                                  \\
ConvNet   & CIFAR 10 & 90    & 0.5581                       & 0.5469                       & 0.0111                                  \\
ConvNet   & CIFAR 10 & 200   & 0.4196                       & 0.4656                       & -0.0460                                 \\
ConvNet   & SVHN     & 50    & 0.6546                       & 0.6604                       & -0.0057                                  \\
ConvNet   & SVHN     & 90    & 0.5479                       & 0.5425                       & 0.0051                                  \\
ConvNet   & SVHN     & 200   & 0.4720                       & 0.4384                       & 0.0336                                  \\
ResNet & CIFAR 10 & 50    & 0.8273                       & 0.8116                       & 0.0157                                  \\
ResNet & CIFAR 10 & 90    & 0.6956                       & 0.7014                       & -0.0058                                  \\
ResNet & CIFAR 10 & 200   & 0.5127                       & 0.5775                       & -0.0648                                 \\
ResNet & SVHN     & 50    & 0.7431                       & 0.7284                       & 0.0147                                  \\
ResNet & SVHN     & 90    & 0.6237                       & 0.6129                       & 0.0107                                  \\
ResNet & SVHN     & 200   & 0.4866                       & 0.4922                       & -0.0056                                 
\end{tabular}
\end{table}
\end{document}